%% file: cas-sc-template.tex
\documentclass[a4paper,fleqn]{cas-sc}

\usepackage[numbers,sort]{natbib}
\usepackage[utf8]{inputenc} 
\usepackage[T1]{fontenc}    
\usepackage{hyperref}       
\usepackage{url}            
\usepackage{booktabs}       
\usepackage{amsfonts}       
\usepackage{nicefrac}       
\usepackage{microtype}      
\usepackage{lipsum}
\usepackage{fancyhdr}       
\usepackage{graphicx}       
\graphicspath{{media/}}     
\usepackage{geometry}
\usepackage{graphicx}
\usepackage{booktabs}
\usepackage{caption}
\usepackage{subcaption}
\usepackage{enumitem}
\usepackage{multirow}
\usepackage{enumitem}
\usepackage{longtable}
\usepackage{array}
\usepackage{graphicx}  
\usepackage{booktabs} 
\usepackage{multicol}
\usepackage{booktabs}
\usepackage{multirow}
\usepackage[normalem]{ulem}
\useunder{\uline}{\ul}{}
\usepackage{lscape}
\usepackage{longtable}
\usepackage{rotating}
\usepackage{graphicx}
\usepackage{float}
\usepackage{longtable} 
\usepackage{array} 
\usepackage{booktabs} 
\usepackage{pifont} 
\usepackage{amssymb}
\usepackage{wasysym}
\def\tsc#1{\csdef{#1}{\textsc{\lowercase{#1}}\xspace}}
\tsc{WGM}
\tsc{QE}
\tsc{EP}
\tsc{PMS}
\tsc{BEC}
\tsc{DE}
\DeclareUnicodeCharacter {263}{\smile}
\usepackage{newunicodechar}
\newunicodechar{☺}{\smiley}
\begin{document}
\let\WriteBookmarks\relax
\def\floatpagepagefraction{1}
\def\textpagefraction{.001}

\shorttitle{Toxic Memes}

\shortauthors{DS Martinez Pandiani et~al.}

\title [mode = title]{Toxic Memes: A Survey of Computational Perspectives on the Detection and Explanation of Meme Toxicities}                      



%
\author[1]{Delfina S. Martinez Pandiani}[type=editor,
                        auid=000,bioid=1,
                        orcid=0000-0003-2392-6300]

\cormark[1]


\ead{delfina.martinez.pandiani@cwi.nl}


\credit{Conceptualization of this study, Methodology, Investigation, Data Curation, Writing - Original Draft, Writing - Review \& Editing, Visualization}

\affiliation[1]{organization={Centrum Wiskunde \& Informatica},
    addressline={Science Park 123}, 
    city={Amsterdam},
    postcode={1098 XG}, 
    country={The Netherlands}}

\author[2]{Erik Tjong Kim Sang}[orcid=0000-0002-8431-081X]
\credit{Conceptualization, Methodology, Data Curation, Writing - Review \& Editing}

\affiliation[2]{organization={Netherlands eScience center},
    addressline={Science Park 402}, 
    city={Amsterdam},
    postcode={1098 XH}, 
    country={The Netherlands}}

\author%
[1]
{Davide Ceolin}[orcid=0000-0002-3357-9130]
\ead{davide.ceolin@cwi.nl}
\credit{Conceptualization, Methodology, Validation, Writing - Review \& Editing, Supervision}

\cortext[cor1]{Corresponding author}



\begin{abstract}
Internet memes, channels for humor, social commentary, and cultural expression, are increasingly used to spread toxic messages. 
Studies on the computational analyses of toxic memes have significantly grown over the past five years, and the only three surveys on computational toxic meme analysis cover only work published until 2022, leading to 
inconsistent terminology and unexplored trends. Our work fills this gap by comprehensively surveying content-based computational perspectives on toxic memes, and reviewing key developments until early 2024. 
Employing the PRISMA methodology, we systematically extend the previously considered papers, achieving a threefold result. 

First, we include in the survey 119 new papers, analyzing 158 computational works focused on content-based toxic meme analysis. Also, we identify over 30 datasets used in toxic meme analysis and examine their labeling systems. Second, after observing the existence of unclear definitions of meme toxicity in computational works, we introduce a new taxonomy for categorizing meme toxicity types. We also note an expansion in computational tasks beyond the simple binary classification of memes as toxic or non-toxic, indicating a shift towards achieving a nuanced comprehension of toxicity.  Third, we identify three content-based dimensions of meme toxicity under automatic study: target, intent, and conveyance tactics. We develop a framework illustrating the relationships between these dimensions and various meme toxicities. 

The survey identifies and analyzes key challenges and recent trends, such as enhanced cross-modal reasoning, integrating expert and cultural knowledge, the demand for automatic toxicity explanations, and handling meme toxicity in low-resource languages. Also, it notes the rising utilization of Large Language Models (LLMs) and generative AI for detecting and generating toxic memes. Finally, the survey proposes pathways for advancing toxic meme detection and interpretation, addressing new opportunities and research gaps. 
\noindent \textcolor{red}{Caution: This work includes toxic memes that may cause psychological distress. Viewer discretion is strongly advised. The example memes do not represent the views or opinions of the authors.} 
\end{abstract}



\begin{highlights}
\item We provide a survey of over 150 papers that computationally analyze toxic memes.
\item We survey labels, task definitions, sources, and usage of 34 datasets of toxic memes.
\item We categorize meme toxicities, providing a meta-model to characterize toxicity dimensions.
\item  We identify key challenges and trends in toxic meme detection and explanation.
\end{highlights}



\maketitle

\section{Introduction}
\label{sec:introduction}
\input{src/intro}

\section{Background}
\label{sec:background}
\input{src/background}

\section{Related Work}
\label{sec:related}
\input{src/related}

\section{Methodology}
\label{sec:methodology}
\input{src/methodology}

\section{Coverage of this Survey}
\label{sec:coverage}
\input{src/coverage}

\section{Toxic Meme Datasets}
\label{sec:datasets}
\input{src/dataset_labels}

\section{Defining Meme Toxicities: A Taxonomy}
\label{sec:meme_toxicities}
\input{src/meme_toxicities}

\section{Dimensions of Meme Toxicities: Target, Intent, Tactic}
\label{sec:dimensions}
\input{src/dimensions}

\section{Recent Trends and Research Directions}
\label{sec:issues_trends}
\input{src/trends}

\section{Discussion and Future Directions}
\label{sec:towards}
\input{src/towards}

\section{Conclusion}
\label{sec:conclusion}
\input{src/conclusion}

\section{Acknowledgements}
 This publication has been supported by the Netherlands eScience Center project ``The Eye of the Beholder'' (project nr. 027.020.G15) and it is part of the AI, Media \& Democracy Lab (Dutch Research Council project number: NWA.1332.20.009). For more information about the lab and its further activities, visit \url{https://www.aim4dem.nl/}.


\appendix
\section{Appendix}
\label{sec:appendix}

\input{src/appendix}


\bibliographystyle{model1-num-names}  
\bibliography{references}

\end{document}

%% file: src/intro.tex

Memes, a term introduced by Richard Dawkins in 1976, serve as cultural replicators analogous to genes, rapidly transmitting ideas between human minds and shaping collective consciousness \cite{dawkins1989selfish}. However, memes also harbor dangers that can profoundly impact individuals and society. They can be likened to insidious ``flukes" that ``hijack and infect the brain" with hazardous ideas, particularly affecting those vulnerable to their influence \cite{ted2007dangerous}. In contemporary discourse, the term ``memes'' has become synonymous with ``internet memes'' or ``image memes'', referring to text-image pairs disseminating swiftly across digital platforms \cite{koutlis2023memetector}. These internet memes are powerful tools for communication and expression, rapidly sharing ideas, emotions, and cultural references across online communities. Though often humorous and lighthearted, internet memes significantly shape public and political discourse. For example, alt-right concepts on platforms like 4chan\footnote{\url{https://www.4chan.org}} and Encyclopedia Dramatica\footnote{\url{https://encyclopediadramatica.online}} create a subcultural language community linked to violent right-wing activism \cite{peeters2021vernacular}. Conversely, progressive leftist meme makers use memes for counternarrative techniques, dialectical seeding, and fostering solidarity within leftist communities \cite{arkenbout2022political}. The widespread presence of memes on social media has sparked interest and concern about their societal impact, playing a crucial role in digital culture and reflecting the collective consciousness of online societies.

The dangers commonly associated with memes are particularly pronounced with internet memes due to their accessibility and virality, which strongly contribute to the spread of toxic ideologies and narratives. The risks associated with toxic memes are extensive (see Table \ref{tab:toxic_memes}). Internet memes can convey toxic representations and messages, perpetuating harmful stereotypes, inciting divisive rhetoric, and contributing to a climate of violence in public discourse \cite{wagener2023semiotic}. Internet memes can serve as conduits of hate speech \cite{Kiela2020hateful_report, Kirk2021Memes, Kumar2022Hate-CLIPper}, harm \cite{sharma2022detecting, Grasso2024KERMIT, Yang2023Invariant}, abuse \cite{das2023banglaabusememe}, cyberbullying \cite{jain2023generative}, offensiveness \cite{shang2021aomd, sharma2020semeval} and various other forms of toxicity \cite{sharma2022detecting}. These memes often cross into illegal territory, exhibiting characteristics of hate speech, incitement to violence, or posing systemic risks to public discourse \cite{arora2023detecting}. Additionally, memes can serve as potent tools for opinion manipulation, including the dissemination of disinformation \cite{williams2020don}, propaganda \cite{Dimitrov2021detecting, abdullah2023combating, rodriguez2023paper}, and trolling \cite{shridara2023identification, suryawanshi2023trollswithopinion}. Political internet memes, often humorous, can oversimplify complex issues, potentially leading to misinterpretation \cite{bebic2018do}. 
Moreover, exposure to and engagement with toxic memes can have profound psychological impacts, leading to phenomena like groupthink or deindividuation \cite{mazambani2015impact} and promoting destructive thoughts and behaviors in individuals \cite{DangersMemes}. The normalization of dark humor, self-deprecating jokes, and derogatory slang within meme culture can blur social norms and exacerbate psychological distress \cite{thecurrentmsuMemeCulture}, as well as desensitizing individuals to tragic news, fostering apathy towards important issues \cite{studybreaksToxicityOnline}. By trivializing violence and desensitizing individuals, extremist behaviors are normalized with real-world manifestations, such as wearing meme-inspired clothing at extremist rallies or by perpetrators of violence, as seen in the Allen, Texas mass shooting \cite{progressiveFreeHelicopter}. Critically, the proliferation of toxic memes on online platforms extends their impact beyond the digital realm, influencing real-world events such as election results \cite{serna2024memes} and instances of physical violence.

\begin{table}[H]
\centering
\footnotesize
\caption{Potential risks associated with toxic memes, identified in literature across various disciplines including computer science, human-computer interaction, internet pragmatics, multimedia, cybernetics, visual and media studies, and more.}
\label{tab:toxic_memes}
\renewcommand{\arraystretch}{1.2}
\begin{tabular}{|p{0.23\linewidth}|p{0.5\linewidth}|p{0.17\linewidth}|}
\hline
\textbf{General Risk} & \textbf{Details} & \textbf{Citations} \\
\hline
Violence in Public Discourse &  Conveying toxic representations and messages, perpetuating harmful stereotypes via hate speech, harm, abuse, cyberbullying, offensiveness, and various other forms of toxicity. & \cite{wagener2023semiotic, Kiela2020hateful_report, Kirk2021Memes, Kumar2022Hate-CLIPper, sharma2022detecting, Grasso2024KERMIT, Yang2023Invariant, das2023banglaabusememe, jain2023generative, shang2021aomd, sharma2020semeval} \\
\hline
Opinion Manipulation & Serving as potent tools for the dissemination of disinformation, propaganda, and trolling, leading to polarization and misunderstanding. & \cite{williams2020don, Dimitrov2021detecting, abdullah2023combating, rodriguez2023paper, shridara2023identification, suryawanshi2023trollswithopinion, bebic2018do} \\
\hline
Psychological Impacts & Promoting groupthink and deindividuation,  destructive thoughts and behaviors, desensitizing individuals to tragic news, fostering apathy, and exacerbating psychological distress. & \cite{DangersMemes, studybreaksToxicityOnline, mazambani2015impact, thecurrentmsuMemeCulture} \\
\hline
Material-World Effects & Exerting tangible impacts of disinformation in election outcomes and instances of physical violence, contributing to the normalization of extremist behaviors. & \cite{serna2024memes, unimelbToxicWent, progressiveFreeHelicopter} \\
\hline
\end{tabular}
\end{table}

Addressing the proliferation of toxic internet memes is crucial for online safety, requiring effective detection and moderation strategies considering their nuanced features. However, moderating toxic internet content is incredibly complex due to fuzzy decision boundaries influenced by cultural considerations \cite{duchscherer2016when}, disagreements among annotators, unconscious biases, and the nuanced nature of harmful language \cite{duchscherer2016when}. Currently, moderation efforts often rely on human crowdworkers in regions like the Philippines, India, and Kenya, who face inadequate protection and significant mental health challenges due to their exposure to toxic content \cite{bennet2020moderates, nondo2023facing, perrigo2022inside, rowe2023destroyed, mbagathi2023africa}. These perils to the workers and the sheer volume of online content exacerbate moderation challenges, prompting a growing demand for automated solutions to detect toxicity and explain their assessment. In essence, there is an urgent need for explainable toxic meme detection, including the identification of specific toxicity types and the provision of criteria that the toxicity is based on, such as slur words, hate symbols, or intricate rhetorical strategies.

\begin{figure}[!h]
    \centering
    \includegraphics[width=.6\textwidth]{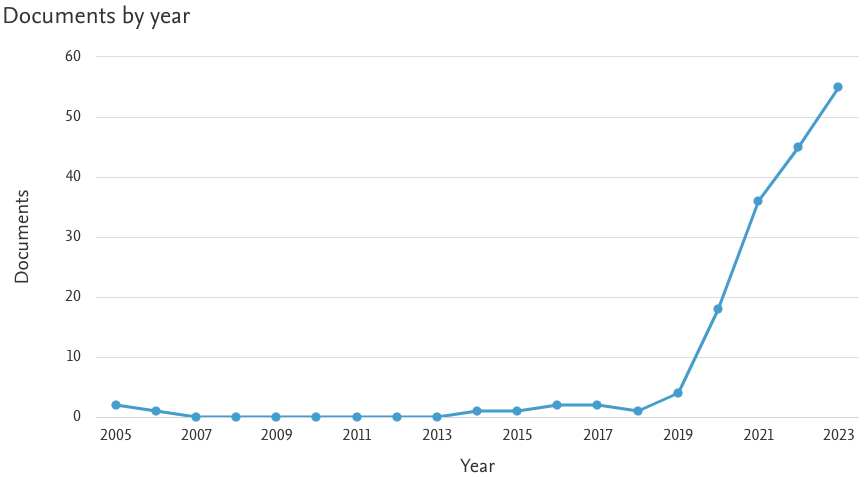}
    \caption{Graph depicting the exponential increase in publications within the field of computer science, as indexed by SCOPUS, focusing on research related to toxic memes. The data was gathered using a query targeting specific keywords associated with meme toxicities (see Section \ref{sec:methodology}).}
    \label{fig:scopus}
\end{figure}


In recent years, there has been a significant increase in computational analysis of internet memes, particularly in research on meme toxicity. This trend is evident from the rising number of Scopus-indexed computer science manuscripts addressing the harmfulness of memes (Figure \ref{fig:scopus}). Despite this growth, the existing literature lacks comprehensive systematic reviews. Only three previous surveys on toxic meme analysis \cite{sharma2022detecting, afridi2021multimodal, hermida2023detecting} exist, but they have limited scope and temporal coverage, ending in 2022.
Recognizing the rapidly evolving nature of this field, our survey aims to incorporate the latest advances and insights. Our goal is to provide researchers, practitioners, and policymakers with a comprehensive understanding of toxic memes from a computational, content-based perspective, covering key developments up to early 2024. Following the PRISMA methodology, our survey systematically reviews 158 computational works selected for their focus on content-based analysis of toxic memes. Content-based analysis refers to analyzing meme content independently of their propagation. While topics like meme propagation and contextual dynamics within social networks, such as measuring visual similarity between image memes across polarized Web communities \cite{zannettou2018origins}, and measuring the potential of harmful memes as precursors for spreading hate on platforms like Twitter \cite{chakraborty2022nipping}, are increasingly studied, they are beyond the scope of this survey.
The key contributions of this survey center around the following:

\begin{itemize}    
    \item \textbf{Analysis of Datasets and Tasks}: We provide an updated overview of available datasets containing toxic memes, including details such as labels, task definitions, sources, and their usage in research, highlighting the diversity in task definitions and labeling schemes.
    
    \item \textbf{Conceptual Overview and Standardized Taxonomy}: We comprehensively review concepts and definitions of computationally studied meme toxicities, addressing the lack of consensus in their definition. Additionally, we introduce a standardized taxonomy for categorizing meme toxicity types.
  
    \item \textbf{Dimensions of Meme Toxicities}: We identify specific dimensions of toxicity in memes (i.e., target, and conveyance tactics), outlining a framework illustrating the relationships between these dimensions and various meme toxicities.

    \item \textbf{Identification of Common Trends}: We identify prevalent challenges and recent trends, including enhanced multimodal reasoning and interpretability, integration of expert and cultural knowledge, and addressing meme toxicity in low-resource languages. We also note the increasing use of LLMs and generative AI in detecting and creating toxic memes.

    \item \textbf{Proposing Pathways for Advancement}: We suggest pathways to tackle challenges and advance the field of toxic meme detection and interpretation. 

\end{itemize}

This survey is structured as follows. In Section \ref{sec:background}, we provide an overview and definitions of (internet) memes and online toxicity. In Section \ref{sec:related}, we discuss previous surveys on toxic memes. Section \ref{sec:methodology} outlines our approach to selecting the papers surveyed. In Section \ref{sec:coverage}, we analyze and discuss the extent of coverage provided by this survey compared to three prior surveys. In Section \ref{sec:datasets}, we provide a catalog of toxic meme datasets and associated label and task definitions. Section \ref{sec:meme_toxicities} 
introduces a novel taxonomy for categorizing the types of meme toxicity. Section \ref{sec:dimensions} outlines a framework illustrating the relationships between dimensions of toxicity. 
Section \ref{sec:issues_trends} explores common challenges and recent trends in computational approaches to detect and interpret toxic memes. 
Lastly, in Section \ref{sec:towards}, we identify key future research directions and 
we conclude in Section \ref{sec:conclusion}.

%% file: src/background.tex
\subsection{Defining (Internet) Memes}





The concept of \textit{memes} draws parallels to biological \textit{genes} and evolution, describing ideas that self-replicate, evolve, and respond to selective Darwinian pressure, ultimately entering culture in ways similar to biological genes and potentially modifying human behavior \cite{dawkins1989selfish}. However, the advent of the digital era has led to a resemantization of the term \cite{polli2023multimodal}, with internet memes being intentional modifications of Dawkins' original concept, characterized by creative alterations rather than random mutation \cite{solon2013richard}. In the digital realm, memes are often defined as digital artifacts sharing common traits of content, form, or perspective, created by users and disseminated, imitated, or transformed via the Internet \cite{shifman2013memes}. While a `memetic construct' is the foundational structure comprising form, content, and perspective, a meme represents a specific instance or manifestation of a memetic construct, embodying a singular multimodal expression of a widespread cultural idea \cite{dynel2022life}. Thus, in this survey, we use the term `meme' to specifically denote what others commonly refer to as `internet' memes \cite{thakur2023explainable}, `visual memes' \cite{xie2011visual}, or `Image With Text' (IWT) memes \cite{du2020understanding}. 

Multimodality stands as a defining feature of memes, as memes rely on a combination of text and images to convey complex messages \cite{dynel2022life}. Typically, memes consist of an image paired with short text, allowing for easy sharing on social media \cite{pramanick2021detecting}. They blend visual and verbal elements to convey humor, irony, or sarcasm, often referencing cultural symbols or events \cite{polli2023multimodal,knobel2007online}. Studies computationally operationalizing the term `meme' have primarily utilized such visuo-linguistic association analysis to distinguish memes from non-meme images and relied on implicit judgement or crowdsourced annotations for differentiation \cite{sharma2020meme}. Recent research \cite{sherratt2023multi} employs multi-channel convolutional neural networks to distinguish memes from not only photographs but also other image-with-text (IWT) formats, such as advertisements, movie posters, online news articles, or screenshots of posts, commonly circulated online. Table \ref{tab:features} summarizes key features of internet memes and associated citations.


\begin{table}[h!]
    \centering
    \caption{Features of (internet) memes, along with their corresponding descriptions and relevant citations.}
        \renewcommand{\arraystretch}{1.5} 
    \begin{tabular}{|p{3.5cm}|p{11.2cm}|}
        \hline
        \textbf{Feature} & \textbf{Description} \\
        \hline
        Multimodal & Combine visual and language information creatively \cite{polli2023multimodal, thakur2023explainable} \\
        \hline
        Succinct & Spread complex messages with a minimal information unit that connects virtual circumstances to real ones \cite{thakur2023explainable} \\
        \hline
        Fluid & Subject to variations and alterations \cite{thakur2023explainable} \\
        \hline
        Anomalous Juxtaposition/ Incongruity & Leverage lack of relevance in the arrangement of textual and/or visual constituents to produce unexpected outcomes \cite{knobel2007online} \\
        \hline
        Intertextual & Reference popular culture, symbols, artifacts, or events that hold meaning within the community of reference \cite{knobel2007online, thakur2023explainable} \\
        \hline
        Relatable/Tacit Background & Rely on viewer’s familiarity with certain contextualized aspects of the world \cite{Kiela2020}, shared knowledge, and implicit cultural references \cite{kostadinovska2018internet} \\
        \hline
    \end{tabular}
    \label{tab:features}
\end{table}

\subsection{Defining Online Toxicity}
\label{sec:toxic_section}

Online toxicity typically refers to negative behaviors on the internet that damage others' or even one's own self-image, hindering personal growth \cite{lapidot2012effects}. This definition is adopted by researchers studying text-image combinations \cite{gordeev2020toxicity} and those exploring `toxic memes' \cite{peeters2021vernacular}. Understanding online toxicity is challenging due to its multifaceted nature, including fine-grained categories and overlapping terms, requiring models capable of recognizing various aspects of toxic behavior \cite{sheth2022defining}. Defining and detecting online toxicity, particularly from a computer science perspective, is further complicated by challenges in annotating data and developing machine learning models to identify toxicity types and relationships \cite{powerof0ToxicityMemes}. For instance, sharing misinformation on social media is associated with harmful language, highlighting the importance of integrating research on two types of toxicity (misinformation and harmful language) usually studied separately \cite{mosleh2024misinformation}.

\subsubsection{Textual Toxicity}
Much of the research on online toxicity has focused on textual data, with extensive studies on detecting toxicity in texts \cite{ghosh2021detoxy} and tasks such as toxic comment classification, hate speech detection, and identification of offensive language. Different taxonomies have been proposed to categorize abusive language, distinguishing between abusive content directed at individuals or groups, and between explicit and implicit abusive content \cite{waseem2017understanding, wiegand2018overview, zampieri2019semeval}.
Even within `hate' speech, there are various definitions and fine-grained labels, and thus a need for comprehensive datasets to train robust models for combating hate speech effectively \cite{piot2024metahate}. 
Some works follow a three-level taxonomy considering the type and target of offense \cite{zampieri2019semeval},
while other toxic comment detection systems follow a multi-label classification framework, with comments labeled as `toxic,' `severe toxic,' `insult,' `threat,' `obscene,' and `identity hate' \cite{jigsaw}. Other datasets use labels like `hate speech,' `offensive but not hate speech,' and `neither offensive nor hate speech,' highlighting biases and challenges in classifying offensive language in short-form content like tweets \cite{davidson2017automated}.


The differentiation of concepts associated with toxic speech presents complexities, often characterized by contentious definitions \cite{ayo2020machine}. Efforts to delineate the hierarchy of hate speech concepts within computer science literature reveal fuzzy boundaries between toxicity-related concepts \cite{alkomah2022literature}, and instances of interchangeability and even conflicting hierarchical relations. For instance, some perspectives consider toxicity as a subset of hate speech \cite{chhabra2023literature}. In response to these challenges, some initiatives aim to harmonize toxicity labels for textual content. For instance, in \cite{fortuna2020toxic}, researchers analyzed six publicly available datasets related to hate speech, aggression, and toxicity in text. Their objective was to standardize categories to ensure consistency and comparability across datasets. This involved merging related categories, clarifying ambiguous labels, and aligning similar concepts under common headings. The resulting taxonomy identifies various types of toxicity and recommends considering unique definitions and contexts before merging terms.

\subsubsection{Multimodal Toxicity}

Exploring multimodal toxicity from a computational perspective has not received as much attention as textual (unimodal) toxicity, yet understanding its dynamics is increasingly crucial \cite{yankoski2021meme}. Recent studies have begun to bridge this gap, offering varied perspectives on toxic multimodal content found online. This section provides an overview of these works, each presenting diverse taxonomies summarized in Table \ref{tab:taxonomies}, with varying levels of granularity and emphasis on different facets of harmful content. For instance, Banko et al. \cite{banko2020unified} provide a taxonomy covering hate, harassment, self-inflicted harm, ideological harm, and exploitation, while Nakov et al. \cite{nakov2021detecting} offer a simpler list of harmful categories. Halevy et al. \cite{halevy2022preserving} focus on violations for malicious purposes, including misinformation and community standard violations such as hate speech and crimes. Pramanick et al. \cite{pramanick2021detecting} distinguish between hateful, offensive, and generally harmful memes, while Sharma et al. \cite{sharma2022detecting} present a taxonomy for internet memes with categories like hateful, offensive, propaganda, harassment/cyberbullying, violence, self-inflicted harm, and exploitation. As seen in Table \ref{tab:taxonomies}, there are commonalities and differences across these taxonomies. For instance, both \cite{banko2020unified} and \cite{nakov2021detecting} cover hate speech, harassment, and violence. However, \cite{banko2020unified} offers more detailed subcategories like doxing and identity attack, whereas \cite{nakov2021detecting} includes a broader range, such as dangerous organizations/people and glorifying crime. 

To elucidate potential taxonomical relationships among the toxicities presented in Table \ref{tab:taxonomies}, we constructed a Venn diagram, available in the appendix (refer to Appendix section \ref{sec:appendix:venn}), to illustrate their intersections and semantic relationships. This process highlighted the complex nuances in defining toxic or harmful multimodal content. For example, we found that many behaviors are categorized by platforms as \textit{misbehavior}, which includes actions that may not necessarily be toxic or harmful but are nonetheless restricted on many platforms, such as nudity. We identified diverse forms of harmful content, including ideological harm, hatefulness targeting protected groups, harassment, and more, emphasizing the multifaceted nature of online toxicity. Echoing the findings of \cite{alkomah2022literature} regarding textual toxicities, our analysis revealed the inherent challenges in delineating boundaries between different types of toxicities.  This exploration also paves the way for identifying novel forms of toxic content that emerge uniquely in multimodal contexts and on previously unrecognized forms of harmful behavior that may not be contingent upon multimodality.

\begin{table}
    \centering
    \caption{Comparison of Categories and Subcategories in Different Taxonomies of Toxic Content}
    \centering
    \footnotesize
    \begin{tabular}{|p{3.5cm}|p{1.cm}|p{1.2cm}|p{1.6cm}|p{4.1cm}|p{2.6cm}|}
        \hline
        \textbf{Taxonomy} & \textbf{Source} & \textbf{Focus} & \textbf{Top-Level \newline Distinctions} & \textbf{Second Level \newline Distinctions} & \textbf{Granular\newline Distinctions} \\
        \hline
        \multirow{12}{*}{\parbox{3cm}{Banko et al 2020 \cite{banko2020unified}}} & \multirow{12}{*}{\footnotesize Research} & \multirow{12}{*}{\footnotesize \shortstack{Harmful\\content}} & \multirow{2}{*}{\scriptsize \parbox{3cm}{Hate/ \\ Harassment}}  & \scriptsize Doxing & \\
        \cline{5-6}
        & & & & \scriptsize Identity Attack &  \\
        \cline{5-6}
        & & & & \scriptsize Identity Misrepresentation & \\
        \cline{5-6}
        & & & & \scriptsize Insult & \\
        \cline{5-6}
        & & & & \scriptsize Sexual Aggression & \\
        \cline{5-6}
        & & & & \scriptsize Threat of Violence & \\
        \cline{4-6}
        & & & \multirow{2}{*}{\scriptsize \parbox{3cm}{Self-Inflicted \\ Harm}}  & \scriptsize Eating Disorder Promotion & \\
        \cline{5-6}
        & & & & \scriptsize Self-Harm & \\
        \cline{4-6}
        & & & \multirow{2}{*}{\scriptsize \parbox{3cm}{Ideological \\ Harm}}  & \scriptsize Misinformation & \\
        \cline{5-6}
        & & & & \scriptsize Extremism, Terrorism \& Organized Crime & {\tiny White Supremacist Extremism}\\
        \cline{4-6}
        & & & \multirow{3}{*}{\scriptsize Exploitation} & \scriptsize Adult Sexual Services & \\
        \cline{5-6}
        & & & & \scriptsize Child Sexual Abuse Materials &  \\
        \cline{5-6}
        & & & & \scriptsize Scams & \\
        \hline

     \multirow{19}{*}{\parbox{3cm}{Nakov et al 2021 \cite{nakov2021detecting} \\ / Arora et al 2023 \cite{arora2023detecting} }} & \multirow{19}{*}{\footnotesize {\parbox{3cm}{Social \\ Media}}} & \multirow{19}{*}{\footnotesize \shortstack{Policy\\clauses}} & \multicolumn{3}{|p{6cm}|}{\scriptsize Violence} \\
        \cline{4-6}
        & & & \multicolumn{3}{|p{6cm}|}{\scriptsize Dangerous Orgs/people} \\
        \cline{4-6}
        & & & \multicolumn{3}{|p{6cm}|}{\scriptsize Glorifying Crime} \\
        \cline{4-6}
        & & & \multicolumn{3}{|p{6cm}|}{\scriptsize Illegal Goods} \\
        \cline{4-6}
        & & & \multicolumn{3}{|p{6cm}|}{\scriptsize Self-Harm} \\
        \cline{4-6}
        & & & \multicolumn{3}{|p{6cm}|}{\scriptsize Child Sexual Abuse} \\
        \cline{4-6}
        & & & \multicolumn{3}{|p{6cm}|}{\scriptsize Sexual Abuse (Adults)} \\
        \cline{4-6}
        & & & \multicolumn{3}{|p{6cm}|}{\scriptsize Animal Abuse} \\
        \cline{4-6}
        & & & \multicolumn{3}{|p{6cm}|}{\scriptsize Human Trafficking} \\
        \cline{4-6}
        & & & \multicolumn{3}{|p{6cm}|}{\scriptsize Bullying and Harassment} \\
        \cline{4-6}
        & & & \multicolumn{3}{|p{6cm}|}{\scriptsize Revenge Porn} \\
        \cline{4-6}
        & & & \multicolumn{3}{|p{6cm}|}{\scriptsize Hate Speech} \\
        \cline{4-6}
        & & & \multicolumn{3}{|p{6cm}|}{\scriptsize Graphic Content} \\
        \cline{4-6}
        & & & \multicolumn{3}{|p{6cm}|}{\scriptsize Nudity and Pornography} \\
        \cline{4-6}
        & & & \multicolumn{3}{|p{6cm}|}{\scriptsize Sexual Solicitation} \\
        \cline{4-6}
        & & & \multicolumn{3}{|p{6cm}|}{\scriptsize Spam} \\
        \cline{4-6}
        & & & \multicolumn{3}{|p{6cm}|}{\scriptsize Impersonation} \\
        \cline{4-6}
        & & & \multicolumn{3}{|p{6cm}|}{\scriptsize Misinformation} \\
        \cline{4-6}
        & & & \multicolumn{3}{|p{6cm}|}{\scriptsize Medical Advice} \\
        \hline

    \multirow{3}{*}{\parbox{3cm}{Pramanick et al 2021 \cite{pramanick2021detecting}}} & \multirow{3}{*}{\footnotesize Research} & \multirow{3}{*}{\footnotesize \parbox{3cm}{Harmful \\ memes}} & 
        \multicolumn{3}{|p{3cm}|}{\scriptsize Hateful} \\
        \cline{4-6}
        & & & \multicolumn{3}{|p{3cm}|}{\scriptsize Offensive} \\
        \cline{4-6}
        & & & \multicolumn{3}{|p{8cm}|}{\scriptsize Other (Generally) Harmful} \\
    \hline

    \multirow{7}{*}{\parbox{3cm}{Halevy et al 2022 \cite{halevy2022preserving}}} & \multirow{7}{*}{\footnotesize {\parbox{3cm}{Social \\ Media}}}  & \multirow{7}{*}{\footnotesize \parbox{1.8cm}{Violations \\ /Malicious \\ Purposes}} & \multicolumn{3}{|p{3cm}|}{\scriptsize Misinformation} \\
        \cline{4-6}
        & & & \multirow{4}{*}{\scriptsize \parbox{3cm}{Community \\ Standards \\ Violations}}  & \multicolumn{2}{l|}{\scriptsize Hate Speech} \\
        \cline{5-6}
        & & & & \scriptsize \multirow{3}{*}{Crimes} & {\tiny Selling Illegal Drugs}\\
        \cline{6-6}
        & & & & & {\tiny Coordinating Sex Trafficking}\\
        \cline{6-6}
        & & & & & {\tiny Child Exploitation}\\
        \cline{6-6}
    \hline

    \multirow{13}{*}{\parbox{3cm}{Sharma et al 2022 \cite{sharma2022detecting}}} & \multirow{13}{*}{\footnotesize Research} & \multirow{13}{*}{\footnotesize \parbox{3cm}{Harmful \\ memes}} & \multirow{8}{*}{\scriptsize Hateful} & \multicolumn{2}{|p{3cm}|}{\scriptsize Doxxing}  \\
            \cline{5-6}
            & & & & \multicolumn{2}{|p{3cm}|}{\scriptsize Identity Attack}  \\
            \cline{5-6}
            & & & & \multicolumn{2}{|p{3cm}|}{\scriptsize Identity Misrepresentation}  \\
            \cline{5-6}
            & & & & \multicolumn{2}{|p{3cm}|}{\scriptsize Insult}  \\
            \cline{5-6}
            & & & & \multicolumn{2}{|p{3cm}|}{\scriptsize Racist}  \\
            \cline{5-6}
            & & & & \multicolumn{2}{|p{3cm}|}{\scriptsize Misogynistic/Sexist}  \\
            \cline{5-6}
            & & & & \multicolumn{2}{|p{3cm}|}{\scriptsize Sexual Aggression}  \\
            \cline{5-6}
            & & & & \multicolumn{2}{|p{7cm}|}{\scriptsize Extremism, Terrorism \& Organized Crime}  \\
            \cline{4-6}
        & & & \multicolumn{3}{|p{3cm}|}{\scriptsize Offensive} \\
        \cline{4-6}
        & & & \multicolumn{3}{|p{8cm}|}{\scriptsize Propaganda} \\
        \cline{4-6}
        & & & \multicolumn{3}{|p{8cm}|}{\scriptsize Harassment/Cyberbullying} \\
        \cline{4-6}
        & & & \multicolumn{3}{|p{8cm}|}{\scriptsize Violence} \\
        \cline{4-6}
        & & & \multirow{2}{*}{\scriptsize \parbox{3cm}{Self-inflicted \\ Harm}} & \multicolumn{2}{|p{3cm}|}{\scriptsize Eating Disorder Promotion}  \\
            \cline{5-6}
            & & & & \multicolumn{2}{|p{3cm}|}{\scriptsize Self-harm}  \\
            \cline{4-6}
        & & & \multirow{3}{*}{\scriptsize Exploitation} & \multicolumn{2}{|p{3cm}|}{\scriptsize Adult Sexual Service}  \\
            \cline{5-6}
            & & & & \multicolumn{2}{|p{3cm}|}{\scriptsize Child Sexual Abuse Material}  \\
            \cline{5-6}
            & & & & \multicolumn{2}{|p{3cm}|}{\scriptsize Scams}  \\
    \hline
    \end{tabular}
    \label{tab:taxonomies}
\end{table}

\clearpage

%% file: src/related.tex
\subsection{Surveys on Toxic Memes from Computational Perspective}
\label{sec:related_surveys}

Only three published works have surveyed the emerging field of computational toxic meme analysis \cite{sharma2022detecting, afridi2021multimodal, hermida2023detecting}. Afridi et al. (2020) \cite{afridi2021multimodal} 
were the first to survey efforts in automatic meme understanding, highlighting key challenges for future research. These challenges include defining hate in memes, distinguishing between humor and hate, and categorizing memes into subcategories targeting relevant issues. They stressed the need for detailed analysis within toxic memes, covering subcategories like hateful/non-hateful, rumor, fake news, and extremism. Additionally, they emphasized advancing techniques for cross-modal entailment in meme interpretation, crucial for tasks like automatic detection. Based on their literature review, they proposed a generic multimodal architecture for meme classification.

Building upon this foundation, Sharma et al. (2022) \cite{sharma2022detecting} conducted a thorough survey to categorize harmful meme types, proposing a new typology including hate, offensive, propaganda, harassment/cyberbullying, violence, and self-inflicted harm. This framework offers a structured approach for understanding and classifying harmful memes based on their characteristics and potential impact. The survey identified significant gaps in current research, highlighting challenges and future directions. Authors stress the need for detailed analysis in detection and interpretation, noting limited exploration of certain meme types due to dataset shortages, like those depicting self-harm and extremism. They also emphasize the global impact of memes, requiring research on cross-cultural implications. Furthermore, the study underscores the semiotic complexity of interpreting multimodal meme content, revealing a need for sophisticated analysis techniques to understand nuanced meanings conveyed through visual and linguistic elements.

Most recently, Hermida and dos Santos (2023) \cite{hermida2023detecting} surveyed methodologies for detecting hateful memes and introduced a taxonomy of machine learning architectures specifically for this purpose. This taxonomy operates on three levels: Level 1 distinguishes between non-attention mechanism-based and attention mechanism-based methods, with the latter generating multimodal representations for memes. Level 2 categorizes methodologies based on how they handle text and image components, with ``restricted'' methods directly using meme text and image information, while ``extended'' methods utilize indirectly extracted data like object tags and sentiment analysis. Both approaches are identified within attention mechanism-based methods, while non-attention mechanism-based methods fall under the restricted category. Level 3 considers the feature extraction process, distinguishing between auto-feature extraction techniques and hand-crafted techniques. Key insights from the survey include the importance of dataset quality and annotation guidelines in training models to detect toxic memes, highlighting limitations such as biases in datasets, small sample sizes, and the ongoing need for human moderation despite advancements in deep learning feature extractors.

\subsection{Other Relevant Surveys On Hate Speech and Disinformation}

Other surveys exploring areas such as multimodal disinformation and hate, while not specifically focusing on memes, provide valuable insights into broader issues surrounding multimodal toxicity. For instance, \cite{alam2021survey} survey computational approaches to multimodal disinformation and harm across different content types: text, speech, images, videos, and network data. Key findings underscore the importance of explainability in model interpretation, the necessity of considering personal preferences and cultural aspects beyond content and network signals, and the promise of knowledge-based approaches for factuality checking. \cite{chhabra2023literature} examine hate speech detection across multimodal and multilingual contexts, exploring various content types, including text, images, and videos, across platforms like private messages, stories, authorized account posts, comments, tweets, ads, user profiles, and sensitive content such as graphic violence and adult material. The authors advocate for proactive measures like blocking or reporting trolls on social media platforms, promoting data analysis before sharing posts, and stressing the importance of robust policy frameworks to counter abusive behavior by social media entities. A recent survey \cite{anjum2024hate} primarily analyzing hate speech in text but also emphasizing the importance of multimodal features, identifies the prevalent reliance on surface-level features such as word frequency, punctuation usage, capitalization, specific keywords or phrases, and basic syntactic structures in hate speech detection approaches. The authors note that this reliance may limit the ability to fully capture the context and meaning of the text, highlight the lack of comparative studies, and stress the importance of open-source code and dataset links for evaluations. Although not addressing multimodality, two other surveys on textual toxicity are worth mentioning: \cite{lewandowska2023integrated} categorizes offensive language types into a hierarchical taxonomy, differentiating between explicit and implicit language, while \cite{garg2023handling} provides insights into the subjective nature of toxicity detection, biases in existing datasets, the influence of content source and topic on dataset characteristics, and challenges in collecting toxic comments. 


\subsection{Need for this Survey}

While the surveys discussed in section \ref{sec:related_surveys} provide valuable insights, they leave a critical gap in the comprehensive analysis of what exactly constitutes ``toxicity'' or ``harm' in memes from a computational perspective. While advocating for refinement of categorization schemes and clearer taxonomies and definitions, none delve deeply into the nuances of terminology interchangeability (e.g., toxic, harmful, malicious), their definitions, or the hierarchical relationships among different categories of harmful memes. Moreover, there remains a dearth of systematic examination regarding the alignment of dataset labels with these taxonomies. The surveys discussed have temporal limitations, covering research up to early 2022. However, in the subsequent two years, computational research on toxic memes has proliferated exponentially, and, impressively, with only three months into 2024, Scopus has indexed over 100 publications and preprints for the year. As such, several emerging trends and areas of research have not been thoroughly explored in existing surveys. These include the utilization of background knowledge and the emphasis on explainability in computational approaches \cite{lin2024towards, jha2024meme}, the increasing use of LLMs for various tasks such as detecting hatefulness, misogyny, offensiveness, sarcasm, harmfulness, and specific harmful memes \cite{lin2023beneath, lin2024towards}, shifts towards more sophisticated evaluation methodologies \cite{chen2023minigpt, dai2023instructblip, hu2023visual, wang2023gpt}, novel approaches for generating toxic memes from benign prompts \cite{wu2023proactive}, and the emergence of new datasets, including GOAT-Bench and datasets in multiple languages beyond English \cite{das2023banglaabusememe, lin2024goat, hossain2022identification}.


%% file: src/methodology.tex
\begin{figure}[!h]
    \centering
    \includegraphics[width=.7\textwidth]{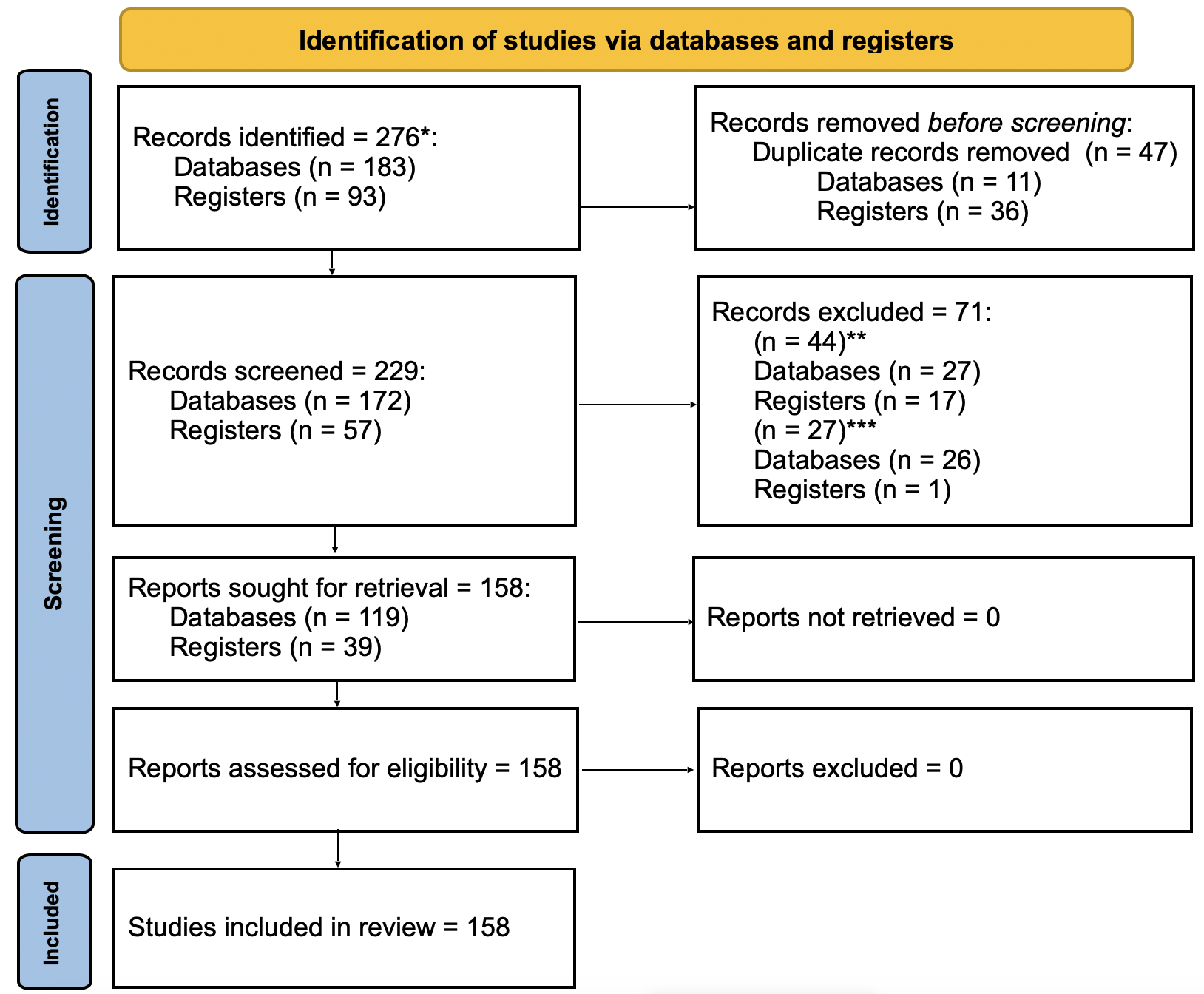}
    \caption{PRISMA 2020 flow diagram for systematic reviews on *SCOPUS and Web of Science (WOS) databases. Registers refers to SCOPUS preprints. Records excluded due to: ** Topic Non-Relevance. *** Computational Non-Relevance.}
    \label{fig:prisma}
\end{figure}

\paragraph{Selection of Database}
We used Scopus\footnote{\scriptsize \url{https://www.scopus.com/}} and Web of Science (WOS)\footnote{\url{https://www.webofscience.com/}} for our literature search because they are two of the largest and most reputable databases, indexing a wide range of high-quality journals across various disciplines and now including preprints.

\paragraph{Inclusion of Preprints}
Acknowledging the novelty of this research field and the rapid pace of publication, we included preprints in our methodology. This decision aligns with the approach taken by the three previous surveys on this topic, emphasizing the importance of capturing the latest developments in this rapidly evolving field.

\paragraph{Identification of Studies to Review}

Inspired by PRISMA (Preferred Reporting Items for Systematic Reviews and Meta-Analyses),\footnote{\scriptsize \url{https://www.prisma-statement.org/}} we used a structured approach to select the manuscripts. Figure \ref{fig:prisma} presents a diagram of our selection process following the PRISMA 2020 guidelines, which outlines the process of manuscript selection.

\paragraph{Database Query} On March 6th, 2024, we queried the databases with a manually crafted key.\footnote{\scriptsize TITLE-ABS-KEY ((toxic OR harmful OR hateful OR unethical OR malicious OR malevolent OR offensive OR propaganda) AND meme)} In SCOPUS, results were confined to “Computer Science”, while in WOS, results were confined to “Computer Science Artificial Intelligence” or “Computer Science Information Systems,” yielding 276 records (183 from databases and 93 from registers).

\paragraph{Record Screening:} We removed 47 duplicate preprints, resulting in 229 manuscripts. 
We manually screened these based on abstracts and titles, excluding 71 irrelevant to our computational focus on toxic memes. We flagged 27 of those for discussion in the related work section, leaving 158 for further evaluation.

\paragraph{Retrieval Assessment for Eligibility:} We retrieved all 158 records (119 peer-reviewed articles and 39 preprints). All 158 manuscripts were manually assessed and were considered eligible.

%% file: src/coverage.tex
\begin{figure}
    \centering
    \includegraphics[width=\linewidth]{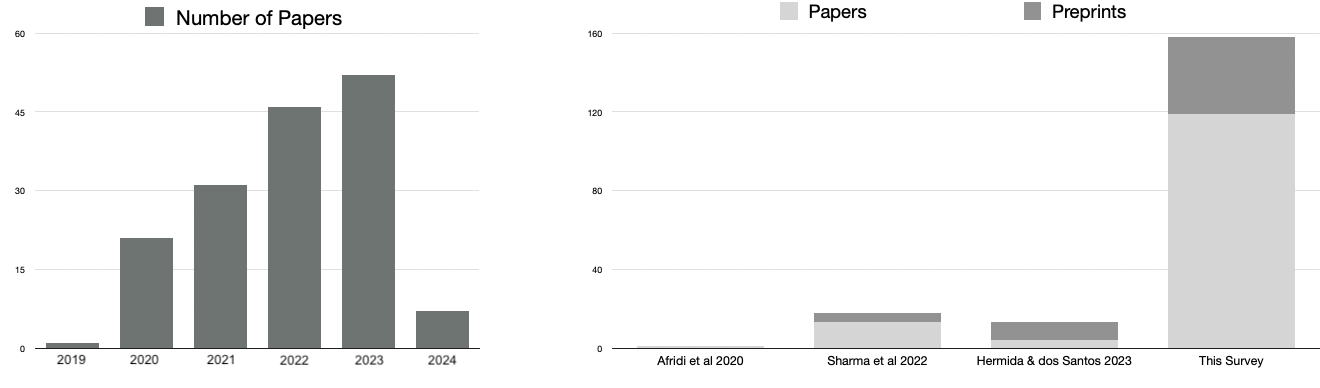}
    \caption{Left: Distribution of surveyed papers by publication year. The figure illustrates a steady increase in the number of publications from year to year. Right: Comparison of coverage across previous surveys based on the papers surveyed here.}
    \label{fig:distribution}
\end{figure}

Our survey includes 158 papers from 2019 to 2024, as shown in Subfigure \ref{fig:distribution}. The number of publications on toxic meme detection has steadily increased over time, peaking in 2023 with 52 papers. This rise indicates growing interest and research in the field. The apparent decline in 2024 is due to our survey covering only the first three months of the year. To evaluate the comprehensiveness of our systematic review on toxic memes, we conducted a systematic analysis to determine the extent to which the papers identified in our review were covered by three previous surveys conducted by Afridi et al.\ (2020) \cite{afridi2021multimodal}, Sharma et al.\ (2022) \cite{sharma2022detecting}, and Hermida and dos Santos (2023) \cite{hermida2023detecting}.  We manually cross-referenced each of the 158 papers in our survey with those covered in the previous works, noting whether each peer-reviewed paper or preprint was included. Detailed tracking information is provided in
the project's Github page.
This analysis quantified the number of our 158 papers covered by each prior survey, considering peer-reviewed papers, preprints, and the total number of papers, as well as the number of additional papers each survey reviewed compared to all previous surveys.
Our results are shown in Table \ref{tab:survey_comparison} and visually represented in Figure \ref{fig:distribution}. We found that although Afridi et al.\ \cite{afridi2021multimodal} call their work a survey on multimodal meme classification, they mostly discuss generic multimodal and visual understanding architectures due to a lack of specific research on memes at the time. Their surveyed papers focus mainly on hate speech detection from text and social media, with only 3 papers computationally addressing memes. 
We note that by the time \cite{sharma2022detecting} was published, the number of papers and preprints on computational meme analysis had increased. This survey extends the previous survey with 16 additional papers. Like Afridi et al.\, Sharma broadens their scope by including related research on textual and multimodal classification to supplement the limited work on memes. Finally, \cite{hermida2023detecting} survey 7 additional papers beyond those covered by \cite{sharma2022detecting}. Our survey reveals that most of the existing computational literature on toxic memes has not undergone a comprehensive review. Only 26 out of the 158 papers identified had been previously surveyed, meaning 84\% (132 out of 158) of the papers we identified using the PRISMA methodology have not, to the best of our knowledge, been reviewed in the context of toxic meme detection and interpretation. These results are unsurprising considering the time gap and rapid rate of new publications; most papers were published after the completion of prior surveys. 
The most recent survey published is the work of Hermida and dos Santos \cite{hermida2023detecting} (2023), but the most comprehensive survey is from Sharma et al.\ \cite{sharma2022detecting}, although it was published in 2022.

\begin{table}[!htp]
    \centering
    \footnotesize
    \caption{Number of papers out of the 158 identified manuscripts included in surveys about automatic detection of (toxic) memes. 
    }
    \label{tab:survey_comparison}
    \begin{tabular}{lcccc}
        \toprule
        \textbf{Survey} & \textbf{Papers} & \textbf{Preprints} & \textbf{Total} & \textbf{Novel Papers Considered} \\
        \midrule
        Afridi et al.\ (2021) \cite{afridi2021multimodal} & 1 & 2 & 3 & 3 \\
        Sharma et al.\ (2022) \cite{sharma2022detecting} & 13 & 6 & 19 & 16 \\
        Hermida \& dos Santos (2023) \cite{hermida2023detecting} & 3 & 10 & 13 & 7 \\
        This survey (2024) & 119 & 39 & 158 & 132 \\
        \bottomrule
    \end{tabular}
\end{table}

%% file: src/dataset_labels.tex
We thoroughly surveyed the selected works for this study, documenting the datasets used in their computational analyses. For each dataset, we recorded details such as its size (number of memes), language, sources (e.g., specific social media platforms, Google Images, etc.), the computational task(s) addressed, annotated features and classification labels, task definitions (e.g., binary, single label multi-class, multi-label multi-class, etc.), and baseline macro F1 scores. This exhaustive examination yielded over 30 datasets (see Table \ref{tab:overview}). More details are available on GitHub.\footnote{\url{https://github.com/delfimpandiani/toxic_memes}}

\begin{table}[!h]
\caption{Overview of the 34 datasets containing toxic memes identified in the literature. For each dataset, we specify the year of introduction, the corresponding manuscript, the languages included, the number of memes, the main focus/task, and the sources of the memes. Abbreviations used: SE - search engines; IP - image hosting platforms (e.g., Pinterest, Imgur); SM - social media platforms (e.g., Facebook, Reddit); MM - meme-specific resources (e.g., Know Your Meme, Memedroid).}
\label{tab:overview}

\scriptsize
\begin{tabular}{p{3.4cm}|p{0.5cm}p{2.5cm}p{0.5cm}p{3.8cm}p{0.3cm}p{0.3cm}p{0.3cm}p{0.3cm}}
        \toprule
        \multirow{2}{*}{\textbf{Dataset}}&
        \multirow{2}{*}{\textbf{Year}}&
        \multirow{2}{*}{\textbf{Language}} &
        \multirow{2}{*}{\textbf{Size}} &
        \multirow{2}{*}{\textbf{Main focus}} &
        \multicolumn{4}{c}{\textbf{Sources}} \\ 
        & & & & & \textbf{SE} & \textbf{IP} & \textbf{SM} & \textbf{MM} \\ \midrule
AOMD Gab \cite{shang2021aomd} & 2021& English & 1965 & offensive meme detection &  &  & \ding{51} &  \\ \midrule
AOMD Reddit \cite{shang2021aomd} & 2021& English & 1094 & offensive meme detection &  &  & \ding{51} &  \\ \midrule
BanglaAbuseMeme \cite{das2023banglaabusememe} & 2021& Bengali, English & 4043 & abusive meme detection & \ding{51} &  & \ding{51} &  \\ \midrule
CrisisHateMM \cite{bhandari2023crisishatemm} & 2022& English & 4700 & hateful meme detection &  &  & \ding{51} &  \\ \midrule
Derogatory Fb-Meme \cite{bhowmick2022multimodal} & 2023& Hindi, English & 650 & derogatory meme detection &  &  & \ding{51} &  \\ \midrule
DisinfoMeme \cite{qu2022disinfomeme} & 2022& English & 1170 & disinformation meme detection &  &  & \ding{51} &  \\ \midrule
ELEMENT \cite{zhang2023mvlp} & 2022& English & 7912 & unethical meme detection & \ding{51} &  & \ding{51} & \ding{51} \\ \midrule
Emoffmeme \cite{kumari2023emoffmeme} & 2023& Hindi & 7500 & offensive meme detection & \ding{51} &  &  &  \\ \midrule
Ext-Harm-P \cite{sharma2022disarm} & 2023& English & 4446 & harmful reference detection & \ding{51} &  & \ding{51} &  \\ \midrule
Facebook Hateful Memes \cite{Kiela2020} & 2022& English & 9540 & hateful meme detection &  &  & \ding{51} &  \\ \midrule
FAME dataset \cite{jabiyev2021game} & 2020& English & 1000 & fake meme detection & \ding{51} &  &  &  \\ \midrule
Fine grained HM \cite{woah2021shared} & 2020& English & 9540 & fine-grained hateful meme detection &  &  & \ding{51} & \ding{51} \\ \midrule
GOAT-Bench \cite{lin2024goat} & 2021& English & 6626 & abusive/toxic meme detection & \ding{51} & \ding{51} & \ding{51} & \ding{51} \\ \midrule
Harm-C (aka HarMeme) \cite{pramanick2021detecting} & 2024& English & 3544 & harmful meme detection & \ding{51} &  & \ding{51} &  \\ \midrule
Harm-P \cite{pramanick2021momenta} & 2021& English & 3552 & harmful meme detection & \ding{51} &  & \ding{51} &  \\ \midrule
Hate Speech in Pixels \cite{sabat2019hate} & 2021& English & 5030 & hateful meme detection & \ding{51} &  & \ding{51} &  \\ \midrule
HatReD \cite{hee2023decoding} & 2019& English & 3228 & hateful meme explanation &  &  & \ding{51} & \ding{51} \\ \midrule
HVVMemes \cite{sharma2022findings} & 2023& English & 7000 & entity roles in harmful memes & \ding{51} &  & \ding{51} &  \\ \midrule
Indian Political Memes \cite{rajput2022hate} & 2022& Hindi, English & 1218 & hateful meme detection & \ding{51} &  &  &  \\ \midrule
Innopolis Hateful Memes \cite{Badour2021Hateful} & 2022& English & 23000 & hateful meme detection & \ding{51} &  & \ding{51} & \ding{51} \\ \midrule
KAU-Memes \cite{bacha2023deep} & 2022& English & 2582 & offensive meme detection &  &  & \ding{51} &  \\ \midrule
Meme-Merge \cite{alzu2023multimodal} & 2023& English & 10000 & offensive meme detection & \ding{51} &  & \ding{51} &  \\ \midrule
Memotion 1 \cite{sharma2020semeval} & 2023& English & 10000 & offensive meme detection & \ding{51} &  &  &  \\ \midrule
Memotion 2 \cite{ramamoorthy2022memotion} & 2020& English & 10000 & offensive meme detection &  & \ding{51} & \ding{51} &  \\ \midrule
MET-Meme \cite{xu2022met} & 2022& English, Chinese & 10045 & offensive meme detection & \ding{51} &  & \ding{51} &  \\ \midrule
Misogynistic-MEME \cite{gasparini2022benchmark} & 2022& English & 800 & misogynous meme detection &  &  & \ding{51} &  \\ \midrule
MultiBully \cite{maity2022multitask} & 2022& Hindi, English & 5854 & cyberbullying meme detection &  &  & \ding{51} &  \\ \midrule
MultiBully-Ex \cite{jha2024meme} & 2022& Hindi, English & 3222 & cyberbullying meme explanation &  &  & \ding{51} &  \\ \midrule
MAMI \cite{fersini2022semeval} & 2024& English & 10000 & misogynous meme detection &  & \ding{51} & \ding{51} & \ding{51} \\ \midrule
MultiOFF \cite{suryawanshi2020multimodal} & 2022& English & 743 & offensive meme detection &  &  & \ding{51} &  \\ \midrule
Pol\_Off\_Meme \cite{kumari2024enhancing} & 2020& Hindi, English & 7500 & offensive meme detection & \ding{51} &  &  &  \\ \midrule
SemEval-2021 Task 6 \cite{Dimitrov2021detecting} & 2024& English & 950 & propagandistic technique detection &  &  & \ding{51} &  \\ \midrule
TamilMemes \cite{suryawanshi2020dataset} & 2021& Tamil & 2969 & troll meme detection &  & \ding{51} & \ding{51} &  \\ \midrule
TrollsWithOpinion \cite{suryawanshi2023trollswithopinion} & 2020& English & 8881 & troll meme detection & \ding{51} &  &  &  \\ \bottomrule
\end{tabular}
\end{table}

\begin{figure}[!ht]
    \centering
    \includegraphics[width=.7\linewidth]{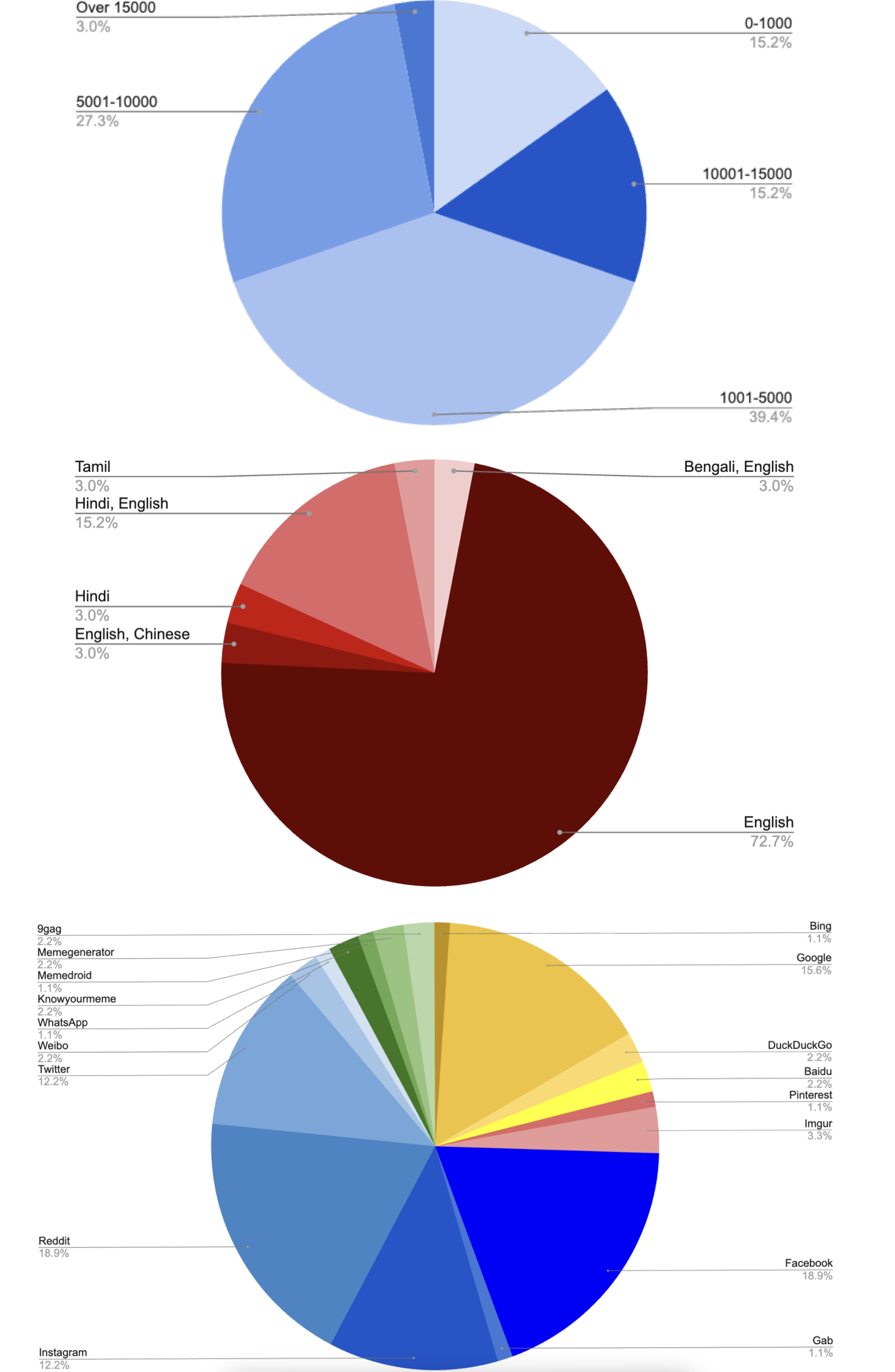}
    \caption{Top: Pie chart showing the distribution of datasets based on the number of memes they contain. Approximately 40\% of datasets contain between 1000 to 5000 memes, followed by nearly 30\% with 5000 to 10,000 memes. Less than 3\% of the datasets have over 15,000 memes. Middle: Pie chart illustrating the distribution of dataset languages. English dominates with nearly 75\% of the datasets exclusively in English, while the remaining datasets include Hindi, Bengali, Tamil, and code-mixed memes. Bottom: Pie chart displaying the sources of memes in the datasets. Social media platforms contribute the most (blue), followed by meme-specific sources (green), search engines (yellow), and image-hosting platforms (red). Over half of the datasets include memes from social media.}
    \label{fig:piecharts}
\end{figure}



\subsection{Dataset Characteristics}

\paragraph{Dataset Size} Our survey reveals a diverse spectrum of dataset magnitudes, spanning from a few hundred to tens of thousands of memes. Among the smaller datasets are Derogatory Facebook-Meme \cite{bhowmick2022multimodal} with 650 memes and MultiOFF \cite{suryawanshi2020dataset} with 743 memes. Conversely, larger datasets include Facebook Hateful Memes (HM) \cite{Kiela2020} with 9,540 memes, alongside MAMI \cite{fersini2022misogynous}, Memotion 1 \cite{sharma2020semeval}, Memotion 2 \cite{ramamoorthy2022memotion}, and MET-Meme \cite{xu2022met}, all hovering around the 10,000 mark. Notably, Innopolis Hateful Memes \cite{Badour2021Hateful} boasts a substantial 23,000 memes. These findings underscore a considerable variance in dataset sizes within the domain. The distribution of dataset magnitudes demonstrates that most datasets lie within the range of 2500 to 10,000 memes 
(see Figure \ref{fig:piecharts} (top)).
    
\paragraph{Language} The majority of datasets primarily contain memes in English (see Figure \ref{fig:piecharts} (middle)), with nearly 75\% exclusively featuring English-language memes \cite{shang2021aomd,sharma2020semeval,Dimitrov2021detecting,suryawanshi2023trollswithopinion,pramanick2021detecting,Kiela2020,lin2024goat,bhandari2023crisishatemm,bhowmick2022multimodal,qu2022disinfomeme,zhang2023mvlp,sharma2022disarm,jabiyev2021game,woah2021shared,pramanick2021momenta,sabat2019hate,hee2023decoding,sharma2022findings,Badour2021Hateful,bacha2023deep,alzu2023multimodal,ramamoorthy2022memotion,gasparini2022benchmark,fersini2022semeval,suryawanshi2020multimodal,suryawanshi2020dataset}. However, we also observe a recent surge in datasets incorporating Asian languages such as Hindi \cite{jha2024meme, kumari2023emoffmeme, maity2022multitask}, Bengali \cite{das2023banglaabusememe}, and Chinese \cite{xu2022met}. Notably, many of these datasets feature ``code-mixed" memes, which blend these languages with English. Interestingly, beyond English and select Asian languages, no datasets have been utilized in the studies we surveyed that incorporate other widely spoken languages, such as Spanish. 

\paragraph{Data Origin} Examining the sources of meme collection revealed a diverse range of platforms, which can be categorized into four macro-categories: social media platforms (e.g., Facebook, WhatsApp), search engines (e.g., Google, Bing), image hosting platforms (e.g., Imgur, Pinterest), and dedicated meme creation and sharing resources (e.g., KnowYourMeme, 9gag) (see Figure \ref{fig:piecharts} (bottom)). Social media platforms emerged as the most prominent sources, with Facebook and Reddit being the most used sources, followed by search engines, with Google being the most utilized search engine. Additionally, platforms such as Pinterest and Imgur were used for meme acquisition. Of particular interest was the utilization of meme-specific platforms like Memegenerator, KnowYourMeme, and 9gag, highlighting the importance of dedicated meme communities in the proliferation of memes. We provide detailed source information for each of the datasets in Table \ref{sec:appendix:dataset_sources} in the Appendix. We also observed instances where new datasets were derived by augmenting previously introduced datasets: Ext-Harm-P \cite{sharma2022disarm} is derived from Harm-P  \cite{pramanick2021momenta}, while Fine grained HM \cite{woah2021shared} and HatReD (Hateful meme with Reasons Dataset) \cite{hee2023decoding} are derived from Facebook Hateful Memes (HM) \cite{Kiela2020hateful_challenge}. Moreover, other datasets were formed by merging existing datasets: Meme-Merge \cite{alzu2023multimodal} is a merge of MET-Meme \cite{xu2022met}, Memotion 1 \cite{sharma2020semeval}, and Memotion 2 \cite{ramamoorthy2022memotion},  while GOAT-Bench \cite{lin2024goat} was created by amalgamating data from several sources including Facebook HM \cite{Kiela2020hateful_challenge}, MAMI \cite{fersini2022semeval}, MultiOFF \cite{suryawanshi2020dataset}, Harm-C \cite{pramanick2021detecting}, and Harm-P \cite{pramanick2021momenta}. This analysis highlights the dynamic and varied meme distribution and consumption landscape across numerous online platforms. It also reveals a growing trend toward reusing existing datasets. Also, it emphasizes the evolution of dataset creation in meme research, showcasing how existing resources are being adapted to address new research questions.

\clearpage

\subsection{Dataset Annotations}
\label{subsec:annotations}

We carefully examined the annotation guidelines of each dataset to identify the types of labels assigned to toxic memes and the computational aspects addressed. Table \ref{tab:annotations} provides a comprehensive summary of the annotated features of meme images across multiple datasets. Each dataset is represented by a column, and each feature is represented by a row. A check mark indicates the presence of labels for a specific feature in each dataset, serving as a checklist to show which aspects of meme images are covered by each dataset. In this process, we found that different datasets employ various annotation methods for the same aspects of memes, including binary, single-label multi-class, or multi-label multi-class schemes, as described below.

\textbf{Abusiveness} is indicated in the BanglaAbuseMeme dataset \cite{das2023banglaabusememe} with each meme labeled as either abusive or not abusive in a binary (yes/no) format.

\textbf{Aggressiveness} is evaluated in the Misogynistic-MEME (MM) dataset \cite{gasparini2022benchmark} using a binary approach.

\textbf{Attack types} are annotated in various datasets. Fine-grained HM \cite{woah2021shared} adopts a multi-label multi-class approach, delineating different attack types within memes, including dehumanizing, inferiority, inciting violence, mocking, contempt, slurs, and exclusion. Similarly, the Multimedia Misogyny Dataset (MAMI) \cite{fersini2022semeval} provides attack type labels specific to misogyny in a multi-label multi-class format, encompassing general misogyny, shaming, stereotype, objectification, and violence.

\textbf{Bullying} categorization is found in both the MultiBully dataset \cite{maity2022multitask} and its extension, MultiBully-Ex \cite{jha2024meme}, with memes labeled as either bully or non-bully using binary labels.

\textbf{Disinformation} presence is annotated in the DisinfoMeme dataset \cite{qu2022disinfomeme} through a binary approach, with memes labeled as either containing disinformation or not (yes/no).

\textbf{Emotion} is a feature of memes that is labeled across datasets in various ways.  Some datasets such as MultiBully \cite{maity2022multitask}, and MultiBully-Ex \cite{jha2024meme} utilize a single-label multi-class approach, tagging memes with emotions like joy, sadness, fear, surprise, anger, disgust, anticipation, trust, or ridicule. Similarly, MET-Meme \cite{xu2022met} employs a single-label multi-class system, categorizing memes with emotions such as happiness, love, anger, sorrow, fear, hate, or surprise. Conversely, Emoffmeme \cite{kumari2023emoffmeme} and Pol\_Off\_Meme \cite{kumari2024enhancing} utilize a multi-label multi-class approach, encompassing annotations for fear, neglect, irritation, rage, disgust, nervousness, shame, disappointment, envy, suffering, sadness, joy, pride, and surprise. In contrast, Memotion 1 \cite{sharma2020semeval} and Memotion 2 \cite{ramamoorthy2022memotion} offer annotations for sarcastic, humorous, motivational, and offensive emotions within a multi-label multi-class framework. The term \textbf{Sentiment} is used to describe this same aspect of memes in Emoffmeme \cite{kumari2023emoffmeme} which uses a multi-label multi-class paradigm for including annotations for fear, neglect, irritation, rage, disgust, nervousness, shame, disappointment, envy, suffering, sadness, joy, pride, and surprise. In contrast, most other datasets utilize sentiment labels in a Likert scale, single-label multi-class approach. For example, Memotion 1 \cite{sharma2020semeval} and Memotion 2 \cite{ramamoorthy2022memotion} categorize sentiments as very positive, positive, neutral, negative, and very negative. Similarly, BanglaAbuseMeme \cite{das2023banglaabusememe}, MultiBully \cite{maity2022multitask}, and MultiBully-Ex \cite{jha2024meme} follow a single-label multi-class scheme with simpler labels: positive, neutral, and negative. Furthermore, the Derogatory Facebook-Meme dataset \cite{bhowmick2022multimodal} employs a binary annotation to indicate negative sentiment (yes/no).

\textbf{Explanation} annotations are present in a couple of datasets: HatReD (Hateful meme with Reasons Dataset) \cite{hee2023decoding} includes human-provided textual explanations. MultiBully-Ex \cite{jha2024meme} provides both textual explanations as textual rationales (highlighted words or phrases) and visual explanations via visual masks (image segmentation) for its categorization of memes as harmful.

\textbf{Fake/misattribution} is annotated in the FAME dataset \cite{jabiyev2021game}, categorizing memes as either fake or real (binary).

\textbf{Harmfulness} is annotated in varied ways across the datasets. The GOAT-Bench dataset \cite{lin2024goat} uses a binary system to indicate harmfulness, labeling each instance as either harmful or not (yes/no). Several other datasets, including Harm-P \cite{pramanick2021momenta}, Harm-C (also known as HarMeme) \cite{pramanick2021detecting}, MultiBully \cite{maity2022multitask}, and MultiBully-Ex \cite{jha2024meme}, categorize the degree of harmfulness using a single-label multi-class system. These datasets label content as very harmful, partially harmful, or harmless, providing a more nuanced understanding of harmfulness. Additionally, the Ext-Harm-P dataset \cite{sharma2022disarm} focuses on the harmfulness of the reference to a social entity. It employs a binary system, labeling references as either harmful or harmless, thus specifically targeting the impact on social entities.

\textbf{Hate speech} is annotated in a varied way across different datasets. Most datasets, including Hate Speech in Pixels \cite{sabat2019hate}, Facebook Hateful Memes (HM) \cite{Kiela2020}, Fine grained HM, \cite{woah2021shared}, CrisisHateMM \cite{bhandari2023crisishatemm}, Innopolis Hateful Memes \cite{Badour2021Hateful}, GOAT-Bench \cite{lin2024goat}, and Derogatory Facebook-Meme \cite{bhowmick2022multimodal}, use a binary labeling system to identify hate speech, categorizing content simply as either hateful or not (yes/no). However, the Indian Political Memes (IPM) dataset \cite{rajput2022hate} employs a more nuanced approach with single-label multi-class annotations among non-offensive, hate-inducing, and satirical. Additionally, CrisisHateMM \cite{bhandari2023crisishatemm} includes an extra dimension by labeling the direction of hate. This dataset differentiates between directed hate and undirected hate, using a binary system to specify whether the hate is aimed at a specific target or more general in nature.

\textbf{Humour} is assessed in Memotion 1 \cite{sharma2020semeval} and Memotion 2 \cite{ramamoorthy2022memotion} datasets on a continuum scale, spanning from not funny to hilarious, within a single-label multi-class framework.

\textbf{Intention} is addressed in a single-label multi-class format in the MET-Meme dataset \cite{xu2022met}, offering labels such as interactive, expressive, purely entertaining, offensive, or other.

\textbf{Irony} is annotated as a binary label (yes/no) in the Misogynistic-MEME dataset \cite{gasparini2022benchmark}, while \textbf{Sarcasm} is similarly represented in binary format (yes/no) in BanglaAbuseMeme \cite{das2023banglaabusememe}, MultiBully \cite{maity2022multitask}, MultiBully-Ex \cite{jha2024meme}, and GOAT-Bench \cite{lin2024goat}. In contrast, sarcasm in Memotion 1 \cite{sharma2020semeval} and Memotion 2 \cite{ramamoorthy2022memotion} is classified along a continuum, with labels ranging from "not sarcastic" to "very twisted," within a single-label multi-class framework.

\textbf{Metaphors} in toxic memes are addressed in the MET-Meme dataset \cite{xu2022met} annotates metaphor, which includes binary labels for \textbf{metaphorical expression} (metaphorical/literal), and also employs a single-label multi-class approach to classify metaphor types into text dominant, image dominant, or complementary.

\textbf{Misogyny} is labeled in a binary manner (yes/no) across multiple datasets, including the Misogynistic-MEME (MM) dataset \cite{gasparini2022benchmark}, Multimedia Misogyny Dataset (MAMI) \cite{fersini2022semeval}, and GOAT-Bench \cite{lin2024goat}.

\textbf{Modality Class} in Memotion 2 \cite{ramamoorthy2022memotion} employs a single-label multi-class approach to label memes based on their modality, allowing them to be categorized as either image and text, image only, or text only.

\textbf{Motivation} is annotated in Memotion 2 \cite{ramamoorthy2022memotion} and Memotion 1 \cite{sharma2020semeval} datasets, distinguishing memes as either motivational or not motivational in a binary manner.

\textbf{Offensiveness} is assessed differently across datasets. Some datasets, such as MultiOFF \cite{suryawanshi2020multimodal}, AOMD Gab \cite{shang2021aomd}, AOMD Reddit \cite{shang2021aomd}, Pol\_Off\_Meme \cite{kumari2024enhancing}, Emoffmeme \cite{kumari2023emoffmeme}, KAU-Memes \cite{bacha2023deep}, TrollsWithOpinion \cite{suryawanshi2023trollswithopinion}, and GOAT-Bench \cite{lin2024goat}, use a binary labeling system (yes/no) to indicate offensiveness. However, in other datasets, such as Memotion 1 \cite{sharma2020semeval}, Memotion 2 \cite{ramamoorthy2022memotion}, and MET-Meme \cite{xu2022met}, offensiveness is assessed on a single-label multi-class scale, ranging from not offensive to hateful offensive. Meme-Merge \cite{alzu2023multimodal} also adopts a degree-based labeling system, with offensiveness categorized from non-offensive to very offensive. Additionally, Pol\_Off\_Meme \cite{kumari2024enhancing}, and Emoffmeme \cite{kumari2023emoffmeme} include binary labels for the explicitness of offensiveness, distinguishing between implicit and explicit offensiveness.

\textbf{Opinion manipulation} is addressed in TrollsWithOpinion \cite{suryawanshi2023trollswithopinion}, which employs a binary labeling approach, categorizing memes as either involving opinion manipulation or not. Additionally, it provides labels for different opinion manipulation types such as political, product, or other, utilizing a single-label multi-class manner to capture these distinctions.

\textbf{Political attributes} are identified in Pol\_Off\_Meme \cite{kumari2024enhancing} through binary classification, distinguishing memes as either political or not political.

\textbf{Profanity} labels are included in the Derogatory Facebook-Meme dataset \cite{bhowmick2022multimodal}, while \textbf{vulgarity} labels are present in BanglaAbuseMeme \cite{das2023banglaabusememe}. Both are annotated in a binary framework of yes/no.

\textbf{Propagandistic techniques} are annotated in SemEval-2021 Task 6 \cite{Dimitrov2021detecting} using a multi-label multi-class approach. This encompasses a wide range of techniques, including Loaded Language, Name Calling/Labeling, Smears, Doubt, Exaggeration/Minimization, Slogans, Appeal to Fear/Prejudice, and more. Each meme can be assigned multiple labels corresponding to the propagandistic techniques it employs.

\textbf{Target} identification in toxic memes is a common aspect annotated across datasets. While Derogatory Facebook-Meme \cite{bhowmick2022multimodal} employs a binary label to indicate whether a meme is targeted or not, most datasets annotate the target in a single-label multi-class manner. Common categories include individual, community, organization, and society, as seen in Harm-P \cite{pramanick2021momenta}, Harm-C (also known as HarMeme) \cite{pramanick2021detecting}, and CrisisHateMM \cite{bhandari2023crisishatemm}. Some datasets provide more specific community targets: BanglaAbuseMeme \cite{das2023banglaabusememe} uses single-label multi-class labeling with options like gender, religion, national origin, individual, political, social sub-groups, and others. Additionally, Fine-grained HM \cite{woah2021shared} employs a multi-label multi-class approach, including labels for protected categories such as religion, race, sex, nationality, and disability. Relatedly, HVVMemes \cite{sharma2022findings} utilizes a single-label multi-class framework to annotate each entity's role, with options including villain, victim, hero, or other.

\textbf{Troll} is annotated as a dimension in both TamilMemes \cite{suryawanshi2020dataset} and TrollsWithOpinion \cite{suryawanshi2023trollswithopinion} using a binary labeling.

\textbf{Unethical} is an aspect of memes annotated in the ELEMENT dataset \cite{zhang2023mvlp} using a binary (yes/no) labeling system.

\begin{table}[H]
\caption{Overview of the various aspects addressed in the identified datasets. Each row represents a dataset, and each column represents an aspect. Tick marks indicate that the dataset addresses the corresponding aspect.}
\scriptsize
\begin{tabular}{p{3.1cm}|p{0.1cm}|p{0.1cm}|p{0.1cm}|p{0.1cm}|p{0.1cm}|p{0.1cm}|p{0.1cm}|p{0.1cm}|p{0.1cm}|p{0.1cm}|p{0.1cm}|p{0.1cm}|p{0.1cm}|p{0.1cm}|p{0.1cm}|p{0.1cm}|p{0.1cm}|p{0.1cm}|p{0.1cm}|p{0.1cm}|p{0.1cm}|p{0.1cm}|p{0.1cm}|p{0.1cm}|p{0.1cm}|}    
&  
        \rotatebox{90}{abusiveness} & \rotatebox{90}{aggressiveness} & \rotatebox{90}{attack type} & \rotatebox{90}{bullying} & \rotatebox{90}{disinformation} & \rotatebox{90}{emotion/sentiment} & \rotatebox{90}{explanation} & \rotatebox{90}{fake/misattribution} & \rotatebox{90}{harmfulness} & \rotatebox{90}{hate speech} & \rotatebox{90}{humour} & \rotatebox{90}{intention} & \rotatebox{90}{irony/sarcasm} & \rotatebox{90}{metaphor}& \rotatebox{90}{misogyny}& \rotatebox{90}{modality-class}& \rotatebox{90}{motivation}& \rotatebox{90}{offensiveness} & \rotatebox{90}{opinion manipulation} &  \rotatebox{90}{political attributes}&  \rotatebox{90}{profanity/vulgarity}&  \rotatebox{90}{propagandistic technique}& \rotatebox{90}{target} & \rotatebox{90}{troll} & \rotatebox{90}{unethical} 
        \\ \midrule
 AOMD Gab \cite{shang2021aomd} & & & &  &  & &   & & & & & & & & & & & \ding{51}& & & & & & &\\ \midrule
 AOMD Reddit \cite{shang2021aomd} & & & &  &  & &   & & & & & & & & & & & \ding{51}& & & & & & &\\ \midrule
 BanglaAbuseMeme \cite{das2023banglaabusememe} & \ding{51}& & & &  & \ding{51}&   & & & & & & \ding{51}& & & & & & & & \ding{51}& & \ding{51}& &\\ \midrule
 CrisisHateMM \cite{bhandari2023crisishatemm} & & & &  &  & &   & & & \ding{51}& \ding{51}& & & & & & & & & & & & \ding{51}& &\\ \midrule
 Derogatory Fb-Meme \cite{bhowmick2022multimodal} & & & &  &  & \ding{51}&   & & & \ding{51}& & & & & & & & & & & \ding{51}& & \ding{51}& &\\ \midrule
 DisinfoMeme \cite{qu2022disinfomeme} & & & &  &  \ding{51}& &   & & & & & & & & & & & & & & & & & &\\ \midrule
 ELEMENT \cite{zhang2023mvlp} & & & & &  & &  & & & & & & & & & & & & & & & & & & \ding{51}\\ \midrule
 Emoffmeme \cite{kumari2023emoffmeme} & & & & &  &  \ding{51} &   & & & & & & & & & & & \ding{51}& & & & & & &\\ \midrule
 Ext-Harm-P \cite{sharma2022disarm} & & & & &  & &   & & \ding{51}& & & & & & & & & & & & & & & &\\ \midrule
 Facebook Hateful Memes \cite{Kiela2020} & & & &  &  & &   & & & \ding{51}& & & & & & & & & & & & & & &\\ \midrule
 FAME dataset \cite{jabiyev2021game} & & & & &  &  &   & \ding{51}& & & & & & & & & & & & & & & & &\\ \midrule
 Fine grained HM \cite{woah2021shared} & & & \ding{51}&  &  & &  & & & \ding{51}& & & & & & & & & & & & & \ding{51}& &\\ \midrule
 GOAT-Bench \cite{lin2024goat} & & & & & & &  & & \ding{51}& \ding{51}& & & \ding{51}& & \ding{51}& & & \ding{51}& & & & & & &\\ \midrule
 Harm-C (aka HarMeme) \cite{pramanick2021detecting} & & & & &  & &   & & \ding{51}& & & & & & & & & & & & & & \ding{51}& &\\ \midrule
 Harm-P \cite{pramanick2021momenta} & & & & &  & &   & & \ding{51}& & & & & & & & & & & & & & \ding{51}& &\\ \midrule
 Hate Speech in Pixels \cite{sabat2019hate} & & & & &  & &   & & & \ding{51}& & & & & & & & & & & & & & &\\ \midrule
 HatReD \cite{hee2023decoding} & & & &  &  & &  \ding{51}& & \ding{51}& & & & & & & & & & & & & & & &\\ \midrule
 HVVMemes \cite{sharma2022findings} & & & & &  & &   & & & & & & & & & & & & & & & & \ding{51}& &\\ \midrule
 Indian Political Memes \cite{rajput2022hate} & & & & &  &  &   & & & \ding{51}& & & & & & & & & & & & & & &\\ \midrule
 Innopolis Hateful  \cite{Badour2021Hateful} & & & & &  & &  & & & \ding{51}& & & & & & & & & & & & & & &\\ \midrule
 KAU-Memes \cite{bacha2023deep} & & & &  &  & &   & & & & & & & & & & & \ding{51}& & & & & & &\\ \midrule
 Meme-Merge \cite{alzu2023multimodal} & & & & &  & &   & & & & & & & & & & & \ding{51}& & & & & & &\\ \midrule
 Memotion 1 \cite{sharma2020semeval} & & & & &  &  \ding{51}&   & & & & \ding{51}& & \ding{51}& & & & \ding{51}& \ding{51}& & & & & & &\\ \midrule
 Memotion 2 \cite{ramamoorthy2022memotion} & & & &  & & \ding{51}&   & & & & \ding{51}& & \ding{51}& & & \ding{51}& \ding{51}& \ding{51}& & & & & & &\\ \midrule
 MET-Meme \cite{xu2022met} & & & & &  & \ding{51}&   & & & & & \ding{51}& & \ding{51}& & & & \ding{51}& & & & & & &\\ \midrule
 Misogynistic-MEME \cite{gasparini2022benchmark} & & \ding{51}& &  &  & &   & & & & & & \ding{51}& & \ding{51}& & & & & & & & & &\\ \midrule
 MultiBully \cite{maity2022multitask} & & & &  \ding{51}&  & \ding{51}&   & & \ding{51}& & & & \ding{51}& & & & & & & & & & & &\\ \midrule
 MultiBully-Ex \cite{jha2024meme} & & & &  \ding{51}&  & \ding{51}&   \ding{51}& & \ding{51}& & & & \ding{51}& & & & & & & & & & & &\\ \midrule
MAMI \cite{fersini2022semeval} & & & \ding{51}&  & & &  & & & & & & & & \ding{51}& & & & & & & & & &\\ \midrule
 MultiOFF \cite{suryawanshi2020multimodal} & & & &  &  & &   & & & & & & & & & & & \ding{51}& & & & & & &\\ \midrule
 Pol\_Off\_Meme \cite{kumari2024enhancing} & & & & &  &  \ding{51}&   & & & & & & & & & & & \ding{51}& & \ding{51}& & & & &\\ \midrule
 SemEval-2021 Task 6 \cite{Dimitrov2021detecting} & & & &  &  & &   & & & & & & & & & & & & & & & \ding{51}& & &\\ \midrule
 TamilMemes \cite{suryawanshi2020dataset} & & & &  & & &   & & & & & & & & & & & & & & & & & \ding{51}&\\ \midrule
 TrollsWithOpinion \cite{suryawanshi2023trollswithopinion} & & & & &  &  &   & & & & & & & & & & & \ding{51}& \ding{51}& & & & & \ding{51}&\\ \bottomrule
\end{tabular}

\label{tab:annotations}
\end{table}

\clearpage

\begin{table}[H]
\caption{Meme toxicity-related terms as utilized in the surveyed literature focusing on computationally addressing toxic memes. It outlines each term alongside its corresponding definition provided by the papers, with the last column indicating the surveyed papers and preprints primarily focused on that specific term.}
\label{tab:toxicity_types}
\scriptsize
\begin{longtable}{|c|p{10cm}|p{3cm}|}
\hline
    \textbf{Toxicity Type} & \textbf{Definition} & \textbf{Focus of} \\
    \hline
    Abusive & Used interchangeably with ``harmful" by \cite{das2023banglaabusememe, das2023transfer}, who describe memes used by bad actors to threaten and abuse individuals or specific target communities. These memes contain words that target individuals or different protected communities, implicitly containing hateful, harmful, and antisemitic content. \cite{titli2023automated} uses it interchangeably with ``offensiveness." & \cite{das2023banglaabusememe, das2023transfer, titli2023automated} \\
    \hline
    Cyberbullying & Defined as any communication that disparages an individual based on characteristics such as color, gender, race, sexual orientation, ethnicity, nationality, or other features \cite{smith2008cyberbullying}. This definition, used by \cite{jha2024meme}, builds upon the work by \cite{jain2023generative}, who consider cyberbullying as an antisocial activity where victims are targeted with malicious behavior, including posting cruel comments and messages without fear of being identified. It appears that both \cite{jain2023generative} and \cite{das2023transfer} consider cyberbullying as a subset of harmful behavior. & \cite{jain2023generative, jha2024meme, kumar2023efficient} \\
    \hline
    Derogatory & Term used by \cite{bhowmick2022multimodal} with no explicit definition. Vaguely defined as posts that convey a derogatory notion about a recognized individual (or person) of the country; malicious content about a political, spiritual, and cultural entity; and/or posts that include hateful and negative sentiments of sentences. & \cite{bhowmick2022multimodal} \\
    \hline
    Disinformation & \cite{qu2022disinfomeme} defines disinformation memes as those designed to actively spread inaccurate information. They differentiate instances that criticize misinformation from those actively spreading inaccurate information. & \cite{williams2020don} \\
       \hline
    Fake/Misattribution & Defined by \cite{jabiyev2021game} as a type of disinformation meme, these contain messages, fabricated or otherwise, falsely attributed to specific individuals. Such memes could be deployed against political opponents during smear campaigns. & \cite{jabiyev2021game} \\
    \hline
    Harmful & \cite{pramanick2021detecting} see harmful memes as those that ``have the potential to cause harm to an individual, an organization, a community, or the society more generally. Here, harm includes mental abuse, defamation, psycho-physiological injury, proprietary damage, emotional disturbance, and compensated public image. [...] Harmful is a more general term than offensive and hateful: offensive and hateful memes are harmful, but not all harmful memes are offensive or hateful'' (page 4). As defined in \cite{sharma2022disarm}, harmful memes encompass various forms of harm expressed overtly or subtly towards target entities, socio-cultural or political ideologies, beliefs, principles, or doctrines associated with them. The harm may manifest as abuse, offense, disrespect, insult, demeaning, or disregard towards the target entity or its affiliations. It can also include more subtle attacks such as mockery or ridicule of a person or an idea. & \cite{Grasso2024KERMIT, Yang2023Invariant, lin2024towards, pramanick2021detecting, lin2023beneath, sharma2022findings, sharma2022disarm, pramanick2021momenta, Ji2023Identifying, fharook2022are, constraint2022, nandi2022detecting, sharma2023characterizing, singh2022combining, zhou2022ddtig, sharma2022domain, koutlis2023memefier} \\
    \hline
    Hateful & In line with \cite{Kiela2020hateful_challenge}, hateful memes are characterized as direct or indirect attacks on individuals based on protected characteristics such as ethnicity, race, nationality, immigration status, religion, caste, sex, gender identity, sexual orientation, disability, or disease. An attack is defined as containing violent or dehumanizing speech, statements of inferiority, or calls for exclusion or segregation. Additionally, mocking hate crimes is classified as hate speech. Exceptions to this definition include attacks on individuals/famous people not based on protected characteristics and attacks on groups perpetrating hate, such as terrorist groups. Detection of hate speech in memes often requires nuanced understanding of societal norms and context. More recently, fine-grained hateful types \cite{zia2021racist} have been studied, focusing on Protected Categories (PC) such as race, disability, religion, nationality, and sex, along with different Attack Types (AT) including contempt, mocking, statements of inferiority, slurs, exclusion, dehumanizing content, and incitement to violence. & \cite{woah2021shared,Badour2021Hateful,Armenta-Segura2023Ometeotl,Aggarwal2021Two,arya2024multimodal,Aggarwal2023HateProof,Aggarwal2021VL,zia2021racist,Ahmed2021Hateful,Bhat2022Detection,Bhat2023Hate,Bi2023You,Blaier2021Caption,Cao2022Prompting,Cao2023Pro-Cap,Chhabra2023Multimodal,Constantin2021Hateful,Deshpande2021interpretable,Fang2022Multimodal,Gaikwad2022Can,Goswami2023Detection,Hee2022On,hee2023decoding,Kiela2020hateful_challenge,Kiela2020hateful_report,Kiran2023Getting,Kirk2021Memes,Kougia2021Multimodal,kougia2023memegraphs,Kumar2022Hate-CLIPper,Lee2021Disentangling,Liang2022TRICAN,Ma2022Hateful,macrayo2023please,Mookdarsanit2021Combating,Nayak2022Detection,Qu2023On,qu2023unsafe,Sethi2021Study,Wanbo2021Research,Wu2022MARMOT,Zhang2023TOT,Zhou2021Multimodal,Zhu2022Multimodal,aggarwal2024text,chen2022multimodal,da2020detecting,evtimov2020adversarial,Gao2021hateful,jennifer2022feels,jin2022hateful,li2021enhance,Lippe2020multimodal,mei2023improving,miyanishi2023causal,Muennighoff2020vilio,polli2023multimodal,rajput2022hate,sandulescu2020detecting,van2023detecting,Velioglu2020detecting,wang2023gpt,wu2023proactive,Yuan2023rethinking,zhang2020hateful,zhao2023review,zhong2020classification,zhou2020multimodal,zhu2020enhance,burbi2023mapping,sabat2019hate}\\
    \hline
    Misogynous &  Misogyny is a form of hate against women, and misogynous memes manifest different expressions of hate directed towards women, encompassing a broad spectrum \cite{fersini2019detecting}. These misogynous manifestations may be categorized into four main types. & \cite{attanasio2022milanlp,behzadi2022mitra,chen2022rit,fersini2022misogynous,gu2022mmvae,kalkenings2022university,obeidat2023just_one,paraschiv2022upb,ravagli2022jrlv,rizzi2023recognizing,singh2023female,thakur2023explainable,tung2023semimemes,singh2023devi} \\
    \hline
    Offensive &  In some of the earliest meme analyses, ``offensive" was conceived as a type of ``emotion" alongside humor, sarcasm, and motivation, simply defined as something that aims to torment or disturb people \cite{sharma2020semeval}. The work introducing one of the most widely used datasets, multiOFF \cite{suryawanshi2020multimodal}, follows \cite{drakett2018old}'s definition of offensive content as intending to upset or embarrass people by being rude or insulting. Offensive or abusive content on social media can be explicit or implicit \cite{waseem2017understanding}, and could be classified as explicitly offensive or abusive if it is unambiguously identified as such. For example, it might contain racial, homophobic, or other offending slurs. In the case of implicit offensive or abusive content, the actual meaning is often obscured by the use of ambiguous terms, sarcasm, lack of profanity, hateful terms, or other means. & \cite{bacha2023deep,alzubi2023multimodal,kumari2024enhancing,aman2021identification,baruah2020iiitg,bejan2020memosys,boinepelli2020sis,bucur2022blue,cui2022meme,dela2020prhlt,giri2021approach,gupta2020bennettnlp,hakimov2022tib,hossain2022identification,hossain2023inter,kumari2023emoffmeme,myilvahanan2023study,nguyen2022hcilab,phan2022little,shang2021aomd,shang2021knowmeme,sharma2020semeval,walinska2020urszula,yu2022wentaorub,zhong2022combining,pramanick2021exercise,gaurav2020machine} \\
    \hline
    
    Propaganda & Defined by \cite{Dimitrov2021detecting} as a form of communication to influence the opinions or actions of people towards a specific goal, achieved through well-defined rhetorical and psychological devices. & \cite{Gundapu2022detection, abdullah2023combating, abujaber2021lecun, alam2022overview, cui2023multimodal, Dimitrov2021detecting, hossain2021csecudsg, li20211213li, liu2022figmemes, rodriguez2023paper, wu2022propaganda, zhu2021ynuhpcc, chen2023multimodal, cui2023beornot} \\
    \hline
    Troll & Defined by \cite{suryawanshi2023trollswithopinion} as a meme that is often provocative, distractive, digressive, or off-topic with the intent to demean or offend particular people, groups, or races, containing either I) Offensive text and non-offensive images; II) Offensive images with non-offensive text; or III) Sarcastically offensive text with non-offensive images, or sarcastic image with offensive text.  & \cite{shridara2023identification, das2022hate, Hedge2021do, nandi2022team} \\
    \hline
    Unethical &As defined by \cite{zhang2023mvlp}, unethical memes are those deemed "inconsistent with human values," failing to comply with ethical norms. They specify ethical memes as multi-modal units consisting of an image and embedded text aiming to promote fairness, justice, harmony, security, avoidance of bias, discrimination, privacy, and prevention of information leakage. While harmful and hateful memes are considered unethical, not all unethical memes are necessarily harmful or hateful. Detection of unethical content in memes is particularly challenging due to its often deeply implicit nature. Ethical memes focus not only on interpersonal principles but also on societal, human, and environmental relationships. & \cite{zhang2023mvlp}
\end{longtable}
\end{table}

\clearpage

%% file: src/meme_toxicities.tex
Based on our investigation of the datasets being developed and utilized in the field, we discovered a diverse array of terms used to describe toxic memes, indicating that multiple types of toxicity are being explored computationally. To identify and classify various types of meme toxicity for computational analysis, we meticulously individually scrutinized each of the 158 research papers to extract explicit terms denoting the types of toxicity they addressed.

\subsection{Meme Toxicities Identification and Definitions}

\begin{figure}[h!]
    \centering
    \includegraphics[width=.6\linewidth]{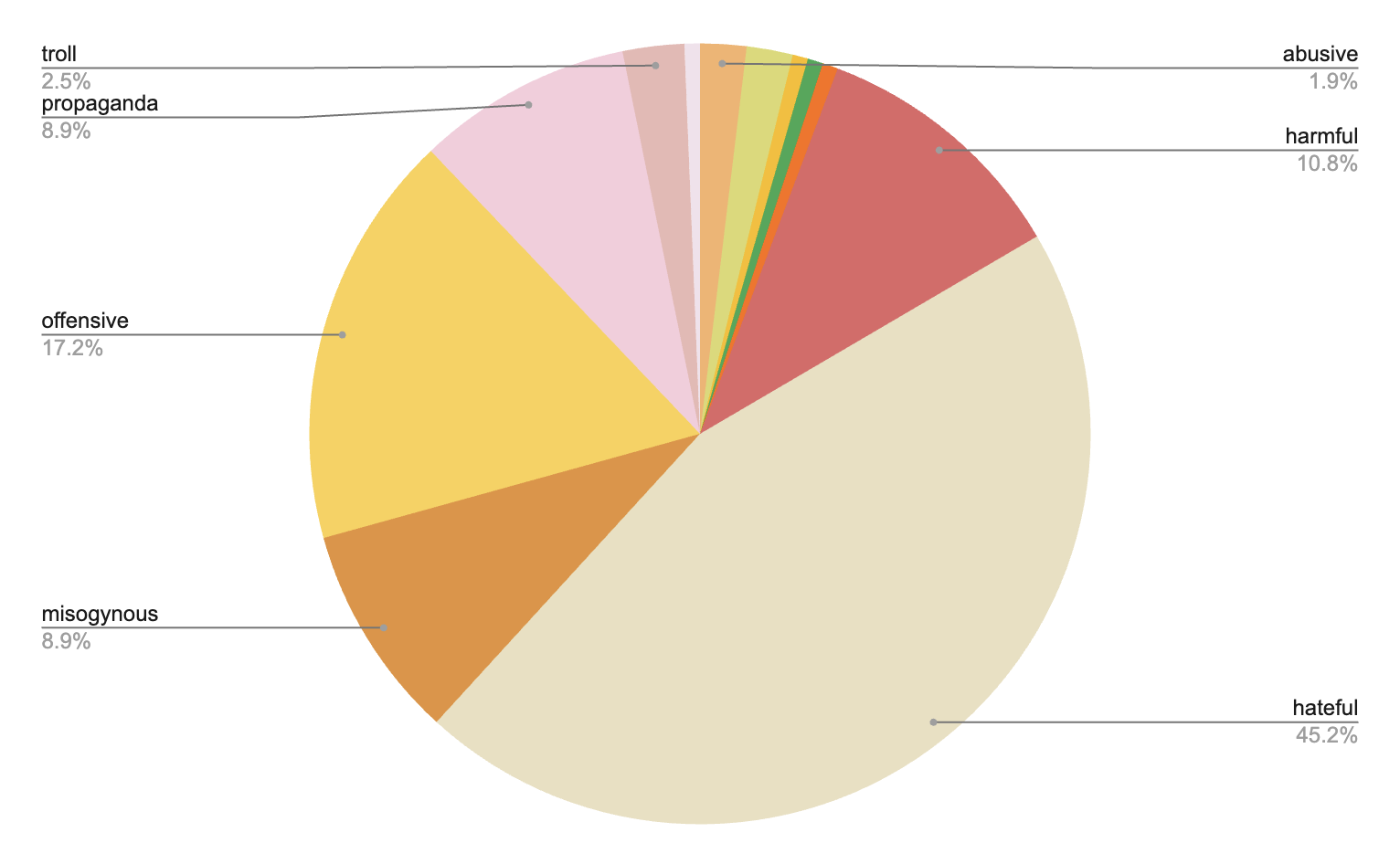}
    \caption{Proportion of papers focused on different meme toxicity categories, based on authors' explicit labels.}
    \label{fig:pie_chart}
\end{figure}

We identified 12 meme toxicity terms: \textit{abusive}, \textit{cyberbullying}, \textit{derogatory}, \textit{disinformation}, \textit{fake}, \textit{harmful}, \textit{hateful}, \textit{misogynous}, \textit{offensive}, \textit{propaganda}, \textit{troll}, and \textit{unethical}. We then compiled the definitions associated with each term and harmonized them into Table \ref{tab:toxicity_types}. Our analysis of toxicity-related terms showed that some terms were used interchangeably in the literature. For example, \textit{abusive} was used synonymously with \textit{offensive} by \cite{titli2023automated} and with \textit{harmful} by \cite{das2023banglaabusememe, das2023transfer}. Additionally, terms like \textit{derogatory} lacked clear definitions and were vaguely described using other toxicity descriptors. Figure \ref{fig:pie_chart} provides an overview of research attention across various meme toxicity categories. Notably, there is a clear imbalance, with a significant focus on hateful memes. Nearly half of the research papers concentrate on this aspect, highlighting the prevalence of hate speech in online discourse. This trend indicates the significant impact of the hateful memes challenge proposed by Kiela et al. (2020) \cite{Kiela2020}, as evidenced by the widespread use of its dataset and methodologies in subsequent research. Our analysis also reveals a lack of research on less prominent toxicity categories like troll, derogatory, and disinformation memes, suggesting gaps in understanding and addressing these forms of toxicity. No surveyed literature dealt with categories flagged in previous work \cite{sharma2022detecting}, e.g., self-harm and exploitation.

\begin{figure}[h!]
    \begin{minipage}{\textwidth}
        \centering
        \includegraphics[width=.6\linewidth]{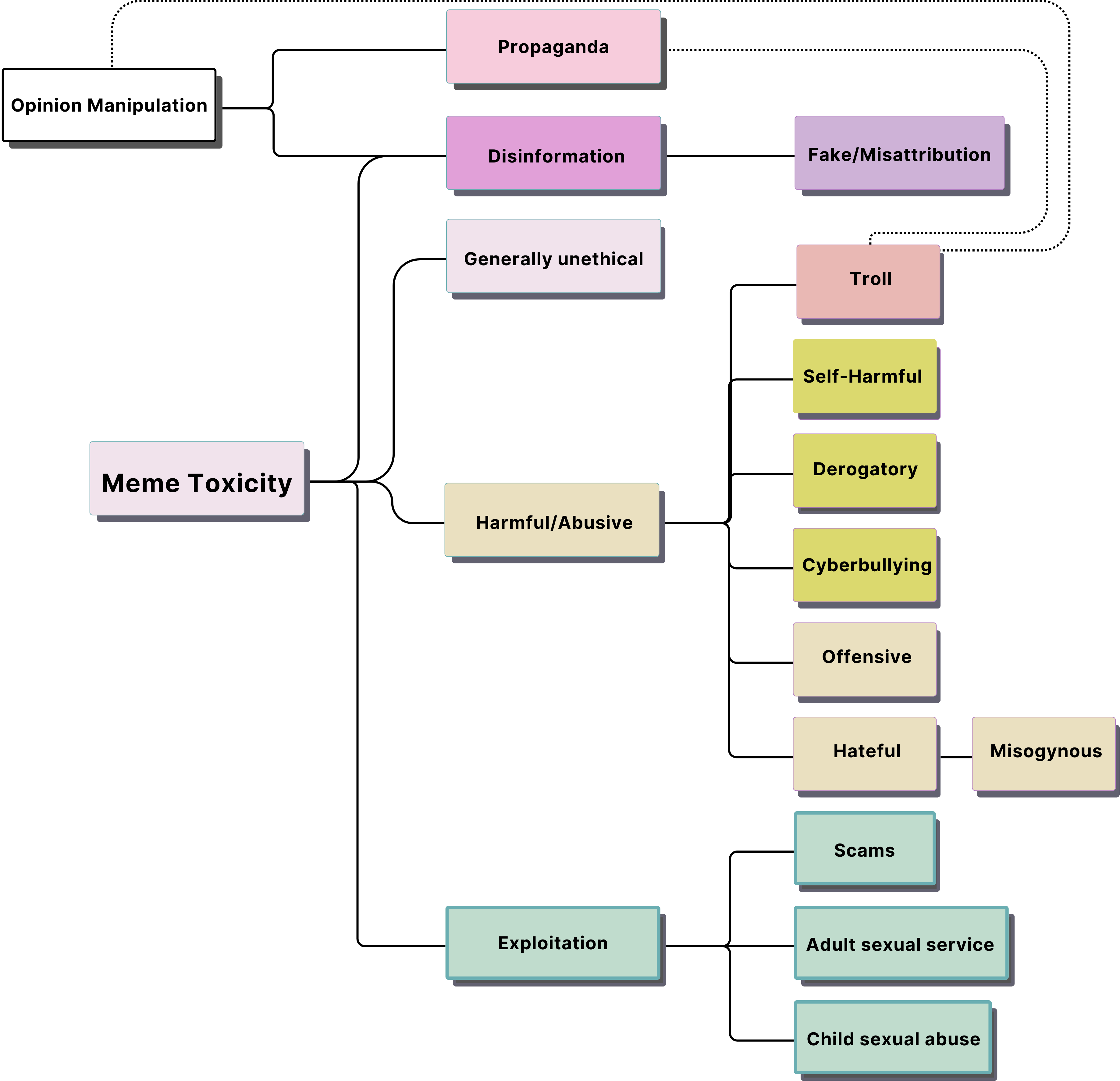}
        \caption{The taxonomy for meme toxicities that we propose is inspired by the taxonomy presented in \cite{sharma2022detecting}, while addressing discrepancies, enhancing taxonomical clarity, and including the most recent types of toxicities being computationally studied.}
        \label{fig:tax_comparison}
    \end{minipage}
    \hfill
    \begin{minipage}{\textwidth}
        \centering
        \includegraphics[width=.8\linewidth]{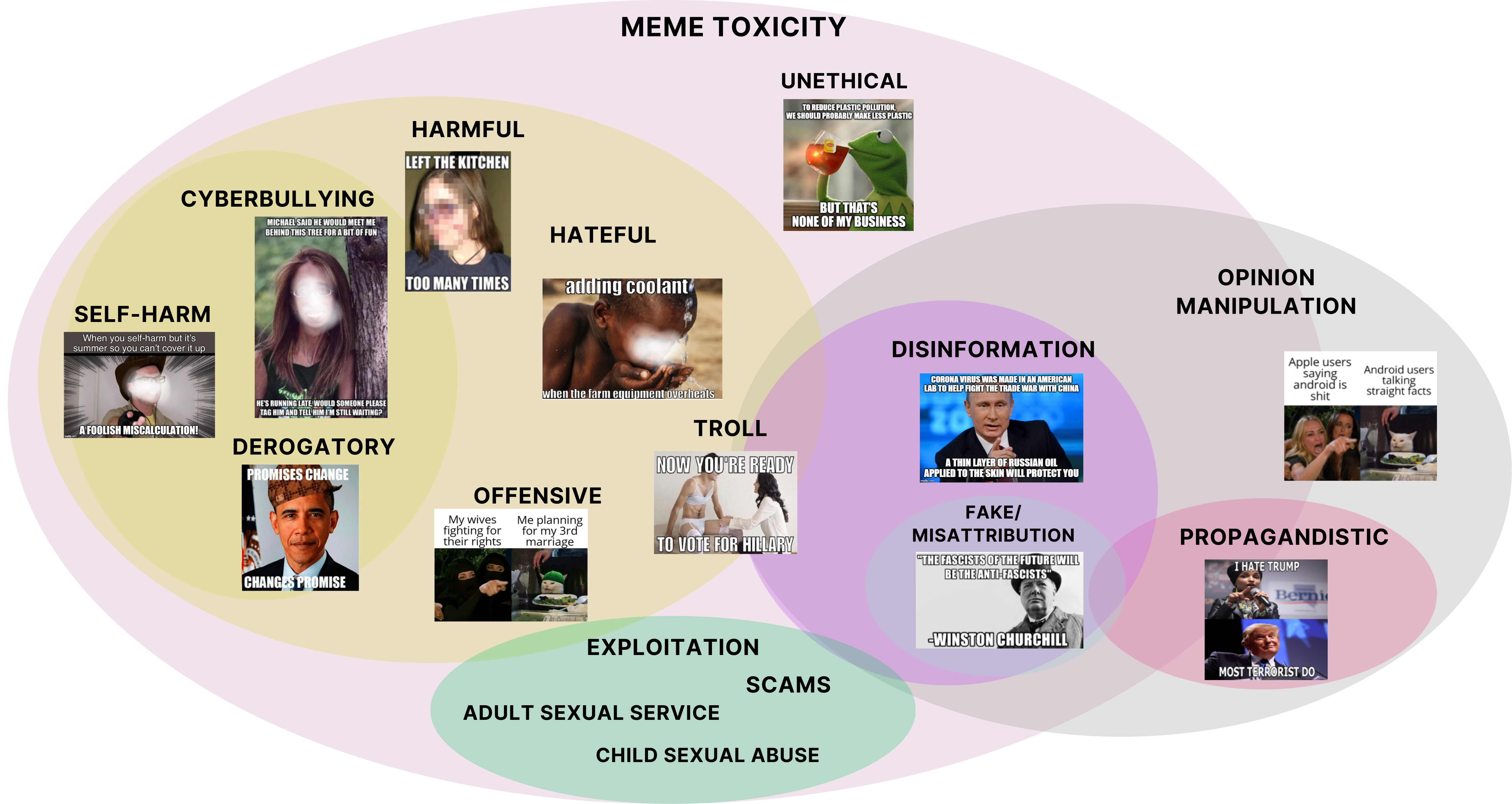}
        \caption{Venn diagram illustrating taxonomical relations and fuzzy boundaries between meme toxicities, along with examples. Memes in the exploitation category are not depicted due to the absence of corresponding datasets. The diagram is color-coded to match our taxonomy. \textcolor{red}{Caution: Depicted memes contain toxic content, potentially inducing psychological distress. Viewer discretion is advised. These images do not reflect the views or endorsements of the authors.}}
        \label{fig:meme_tax_venn}
    \end{minipage}
\end{figure}

\subsection{A Taxonomy of Meme Toxicities}

In examining toxicity term definitions, we identified explicit and implicit taxonomical relationships among certain terms. For example, offensive, hateful, troll, and cyberbullying memes are typically defined as inherently harmful. Fake memes are seen as a subset of disinformation memes, while misogynous memes are considered a subtype of hateful memes targeting protected categories, such as sex. Recognizing the importance of clarifying these relationships, we aimed to establish a coherent taxonomy. We referred to the existing meme toxicity taxonomy by Sharma et al. (2022) \cite{sharma2022detecting}, which included overlooked meme toxicities like self-harm and exploitation (e.g., adult sexual services, child sexual abuse, and scams). However, significant discrepancies emerged. Their taxonomy overlooked certain toxicities like disinformation, fake, and derogatory memes and lacked the concept of unethical memes, a superclass encompassing harmful memes and more.  Additionally, the rationale for placing certain terms under specific macrocategories was unclear. For example, `doxing'—defined as publicly publishing someone's private information as punishment or revenge—was categorized under hate rather than cyberbullying, even though it targets individuals rather than protected communities. Most critically, the taxonomy conflated two dimensions of meme toxicities: the type of toxicity (e.g., hateful, disinformation) and the techniques or attacks used (e.g., identity attack, identity misrepresentation). We believe separating these dimensions is vital, as certain techniques can be used across different toxicities (see Section \ref{sec:dimensions}).

Building on the taxonomy by Sharma et al. (2022) \cite{sharma2022detecting}, we refined it to address identified discrepancies and nuances, enhancing its adaptability for future meme toxicity studies. Our revised taxonomy, shown in Figure \ref{fig:tax_comparison}, is based on definitions from the literature, with each category as a subcategory of another. We found that certain toxicities, such as disinformation and propaganda, stem more from opinion manipulation, while many computationally studied toxicities fall under harmful/abusive, characterized by intent to abuse, demean, offend, or exploit. Our review of the state of the art revealed complex relationships between categories and subcategories of meme toxicities, showing a level of nuance beyond what a taxonomy alone can convey. To visualize these relationships, we developed a Venn diagram (see Figure \ref{fig:meme_tax_venn}) that uses color coding to enhance clarity and illustrate the interconnectedness of various toxicity labels. This diagram provides insight into the complexity of meme toxicity relationships, highlighting the nuanced and sometimes fuzzy boundaries between different toxicities. While the taxonomical approach remains crucial for computational studies, the Venn diagram reflects the real-world complexity of these boundaries.

%% file: src/dimensions.tex
\begin{figure}[!ht]
    \centering
    \includegraphics[
    width=0.4\linewidth]{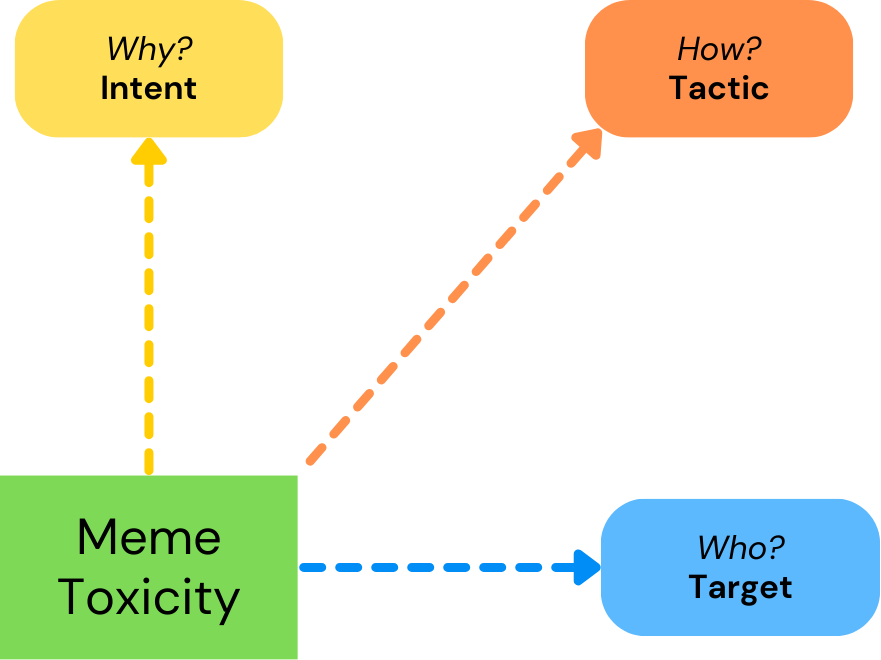}
\caption{The toxicity of memes is a complex phenomenon with multiple dimensions, represented by dotted edges. In this survey, we have identified three content-based dimensions of meme toxicity: target, intent, and tactic. These dimensions address the fundamental questions: \textit{who} is the toxicity directed towards, \textit{why} the toxic meme is shared (i.e., the underlying goal), and \textit{how} the toxicity is manifested and conveyed. Other dimensions also exist and should be studied, potentially answering questions like \textit{where} the meme was posted and \textit{when} it was shared.}    \label{fig:dimensions}
\end{figure}

While refining the taxonomy, we noticed that meme toxicity definitions often focused on three main aspects: the entity to whom the toxicity is directed, the intention behind a meme's creation or dissemination, and the rhetorical strategies employed to convey the toxic narratives. While terminology varied and their interrelation wasn't explicitly addressed, we identified these as three dimensions of meme toxicities: intent, target, and tactic(s) (see Figure \ref{fig:dimensions}). These dimensions could provide a structured framework for understanding and addressing meme toxicity.

\subsection{Target}

The most explicitly researched dimension of toxic memes is the \textit{intended target} of the meme. In memes, targeting involves the harmful referencing of social entities \cite{pramanick2021detecting, sharma2022disarm}. This harm can manifest in various forms such as mental abuse, psycho-physiological injury, proprietary damage, emotional disturbance, or damage to public image, based on the background of the entity (bias, social background, educational background, etc.) by the meme author. It is important to note that memes can reference social entities without causing harm; simply being referenced does not constitute being the target. Non-harmful referencing in memes includes benign mentions (or depictions) via humor, limericks, harmless puns, or any content that doesn't cause distress \cite{sharma2022disarm}. Research has focused on detecting the targets of, particularly, harmful memes \cite{pramanick2021detecting, bhandari2023crisishatemm, sharma2022disarm, pramanick2021momenta}, based on the following target types:

\paragraph{Individual}Toxic memes often target single individuals, typically celebrities like politicians, actors, artists, scientists, or environmentalists, as noted by Sharma et al. \cite{sharma2022disarm}. These individuals include figures such as Donald Trump, Joe Biden, Vladimir Putin, Hillary Clinton, Barack Obama, Chuck Norris, Greta Thunberg, and Michelle Obama. Other studies have noted specific subtypes of individuals as targets of certain meme categories: \textbf{famous or known individuals} are targeted in misattribution memes, with 20 well-known figures commonly featured in fake quote memes \cite{jabiyev2021game}. Additionally, \textbf{meme creators/posters} are implicitly recognized as targets of self-harmful memes, while\textbf{ meme viewers/receivers} are implicitly targeted in scam memes. Individual \textbf{children} may be targeted in memes related to child sexual abuse, and \textbf{adults} are targeted in memes related to adult sexual service \cite{sharma2022detecting}.

\paragraph{Organization}Groups with specific purposes, such as businesses, government departments, companies, institutions, or associations are targeted, e.g., Facebook, the WTO, and the Democratic Party.

\paragraph{Community}These targets are social units defined by shared personal, professional, social, cultural, or political attributes, including religious views, country of origin, or gender identity. These communities may manifest in geographical areas or virtual spaces facilitated by online communication platforms. Within this context, \textbf{Protected Category (PC) Communities} represent a focal point, as identified in Fine-grained version of Hateful Menes \cite{woah2021shared}. This dataset introduced fine-grained labels for protected categories, including \textbf{race}, \textbf{ethnicity}, \textbf{national origin}, \textbf{disability}, \textbf{religious affiliation}, \textbf{caste}, \textbf{sexual orientation}, \textbf{sex},\textbf{ gender identity}, and \textbf{serious disease}. However, the dataset provided primarily covers race, disability, religion, nationality, and sex \cite{zia2021racist}. A similar target community labeling scheme was done in the BanglaAbuseMeme dataset\cite{das2023banglaabusememe}, in which communities are seen as related to gender, religion, national origin, individual, political, social sub-groups, or others. 

\paragraph{Society} The target is the entire societal fabric, e.g., when memes promote conspiracies, harming the general public.

\subsection{Intent}

Multimodal intent recognition plays a crucial role in understanding human language within real-world multimodal scenes. However, existing intent recognition methods face limitations in fully utilizing multimodal information, largely due to the constraints of benchmark datasets containing only text information \cite{zhang2022mintrec}. In the case of memes, intent refers to the underlying purpose or motivation behind their creation or dissemination, and it has been sometimes annotated \cite{xu2022met}. However, the labels offered to choose between are interactive, expressive, purely entertaining, offensive, or other. This implies that the only intent connected to toxicity is the \textit{offensive} intent, defined as the intent for discrimination, satire, and abuse directed towards others' occupation, gender, appearance, or personality \cite{xu2022met}. While it's widely assumed in the literature we surveyed that the intent of all toxic memes is unethical, involving the conveyance of information contrary to ethical societal values like fairness, justice, and privacy \cite{zhang2023mvlp}, specific types of toxic intents unique to toxic memes have not, to the best of our knowledge, been studied. However, as based on the collected definitions of meme toxicities, there is also an acknowledgment of specific intents, such as the intent to harm, misinform, or exploit, that partly characterize the type of toxicity that the meme carries. comprehensive research explicitly identifying and comparing these different intents of toxic memes to each other is lacking. Taking a step in this research direction, based on our survey of meme toxicities, we have identified three subtypes of intents in toxic memes:

\paragraph{Harm/Abuse}The intent is to harm, offend, demean, abuse, or insult social entities \cite{pramanick2021momenta}.

\paragraph{Disinform}The intent is to manipulate opinions by actively spreading false or inaccurate information \cite{williams2020don, jabiyev2021game}.
    
\paragraph{Exploit}This intent is mentioned, but not explicitly defined, by Sharma et al (2022) \cite{sharma2022detecting}, who identified memes attempting to exploit people, including children for sexual abuse, adults for sexual services, or viewers for scams. 


\subsection{Tactic(s)}

We also noticed a growing emphasis on automatically identifying the rhetorical techniques or tactics used by memes to convey toxicity. This focus is justified as memes employ a wide range of rhetorical strategies, persuasion techniques, and tactics to convey messages within their narratives \cite{Dimitrov2021detecting}. This aspect, explored through attack types \cite{woah2021shared, fersini2022misogynous, zia2021racist}, persuasion/propagandistic techniques \cite{Dimitrov2021detecting}, or entities' roles in harmful memes \cite{constraint2022}, represents a third dimension implicitly or explicitly mentioned in many papers, but these terms have not, to our knowledge, been collectively examined.

\paragraph{\textit{Attack types}} delineate how memes, particularly those with hateful connotations, target individuals, groups, or ideas, and serve to categorize how memes express their message \cite{ woah2021shared, zia2021racist}.  
Attack types were defined within the context of the WOAH 5 shared task on fine-grained hateful memes detection \cite{woah2021shared}. Attack types have also been specialized to address specific forms of hateful messaging in the realm of misogynous memes \cite{fersini2022misogynous}:
    
    \begin{itemize}
     
        \item \textbf{Dehumanization}: Explicitly or implicitly describing or presenting a group as subhuman \cite{woah2021shared}. It includes \textbf{misogynous objectification}, i.e., treating or regarding women as objects, devoid of agency/humanity \cite{fersini2022misogynous}.
            
        \item \textbf{Statements of Inferiority}: Claiming that a group is inferior, less worthy or less important than either society in general or another group \cite{woah2021shared}.
        
        \item \textbf{Violence}: Explicitly or implicitly calling for harm to be inflicted on a group, including physical attacks \cite{woah2021shared}. It includes \textbf{misogynous violence}, i.e., the incitation or direct expression of acts of violence against women \cite{fersini2022misogynous}.
        
        \item \textbf{Mocking/Shaming}: Making jokes about, undermining, belittling, or disparaging a group \cite{woah2021shared}. It includes \textbf{misogynous shaming}, defined as the practice of criticising women who violate expectations of behaviour and appearance regarding issues related to gender typology (such as “slut shaming”) or related to physical appearance (such as “body shaming”). This category focuses on content that seeks to insult and offend women because of some characteristics of their body or personality \cite{fersini2022semeval}.
        
        \item \textbf{Expression of Contempt}: Expressing intensely negative feelings or emotions about a group \cite{woah2021shared}.
        
        \item \textbf{Slurs}: Using prejudicial terms to refer to, describe or characterise a group \cite{woah2021shared}.
        
        \item \textbf{Calls for Exclusion}: Advocating, planning or justifying the exclusion or segregation of a group from all of society or certain parts \cite{woah2021shared}.
    
        \item \textbf{Misogynous Stereotyping}: Propagating generalized beliefs about women across various contexts, including societal roles, personalities, and behaviors \cite{fersini2022misogynous}. Stereotypes are fixed, conventional ideas or set of characteristics assigned to a woman, a meme can use an image of a woman according to her role in the society (role stereotyping), or according to her personality traits and domestic behaviours (gender stereotyping) \cite{fersini2022semeval}.
    
    \end{itemize}

\paragraph{\textit{Persuasion or propagandistic techniques}} encompass various methods employed to influence the perception of a meme's audience. These methods include selective editing of images or text, framing narratives to elicit specific emotional responses, and employing symbols and motifs associated with particular ideologies or agendas. One significant study \cite{Dimitrov2021detecting} identifies key propaganda techniques that serve as shortcuts in the argumentation process of memes specifically, leveraging audience emotions or logical fallacies to influence perception. Notably, the presence of these techniques does not inherently categorize a meme as propagandistic; rather, memes are only annotated based on the propaganda techniques they contain.
    
    \begin{itemize}
        \item \textbf{Loaded language}: Using emotionally charged words and phrases to influence the audience.
        \item \textbf{Name calling or labeling}: Assigning labels to entities that evoke strong reactions from the target audience.
        \item \textbf{Doubt}: Questioning the credibility of someone or something.
        \item \textbf{Exaggeration / Minimization}: Representing something in an excessive or diminished manner.
        \item \textbf{Appeal to fear / prejudices}: Instilling anxiety or panic to build support for an idea.
        \item \textbf{Slogans}: Brief and striking phrases acting as emotional appeals.
        \item \textbf{Whataboutism}: Discrediting an opponent's argument by charging them with hypocrisy.
        \item \textbf{Flag-waving}: Appealing to strong national or group sentiments to justify or promote an action or idea.
        \item \textbf{Straw man}: Substituting an opponent's proposition with a similar one, which is then refuted.
        \item \textbf{Causal oversimplification}: Assuming a single cause or reason when multiple causes exist for an issue.
        \item \textbf{Appeal to authority}: Stating that a claim is true simply because an authority on the issue said it was true.
        \item \textbf{Thought-terminating cliché}: Phrases discouraging critical thought and meaningful discussion on a given topic.
        \item \textbf{Black-and-white fallacy or dictatorship}: Presenting two alternative options as the only possibilities.
        \item \textbf{Reductio ad Hitlerum}: Disapproving an action or idea by suggesting it's popular with hated groups.
        \item \textbf{Repetition}: Repeating the same message to influence acceptance.
        \item \textbf{Obfuscation}, Intentional vagueness, Confusion: Using deliberately unclear words.
        \item \textbf{Presenting irrelevant data (Red Herring)}: Introducing irrelevant material to divert attention.
        \item \textbf{Bandwagon}: Persuading the target audience to join a course of action because ``everyone else is doing it.''
        \item \textbf{Smears}: Attempting to damage someone's reputation through negative propaganda.
        \item \textbf{Glittering generalities}: Using words or symbols that produce a positive image when attached to a person/issue.
        \item \textbf{Appeal to (strong) emotions}: Using emotionally charged images to influence the audience.
        \item \textbf{Transfer}: Evoking an emotional response by projecting qualities of a person, entity, or value onto another.
    \end{itemize}

\paragraph{\textit{Entity roles}} are another feature investigated in the rhetorical dimension of meme toxicity, particularly within harmful memes. This task revolves around identifying which entities are glorified, vilified, or victimized within a meme. Framed from the perspective of the meme's author, the objective is to classify, for a given pair of a meme and an entity, whether the entity is depicted as a Hero, Villain, Victim, or falls into another category within that meme \cite{constraint2022}:
    \begin{itemize}
        \item \textbf{Hero}: Entities portrayed in a positive light, often glorified for their actions or inferred from context.
        \item \textbf{Villain}: Entities depicted negatively, associated with adverse traits like wickedness, cruelty, or hypocrisy.
        \item \textbf{Victim}: Entities shown suffering from the negative consequences of someone else’s actions or conveyed implicitly.
        \item \textbf{Other}: Entities that do not fit the categorization of hero, villain, or victim within the context of the meme.
    \end{itemize}

\begin{figure}[!ht]
    \centering
    \includegraphics[width=0.52\linewidth]{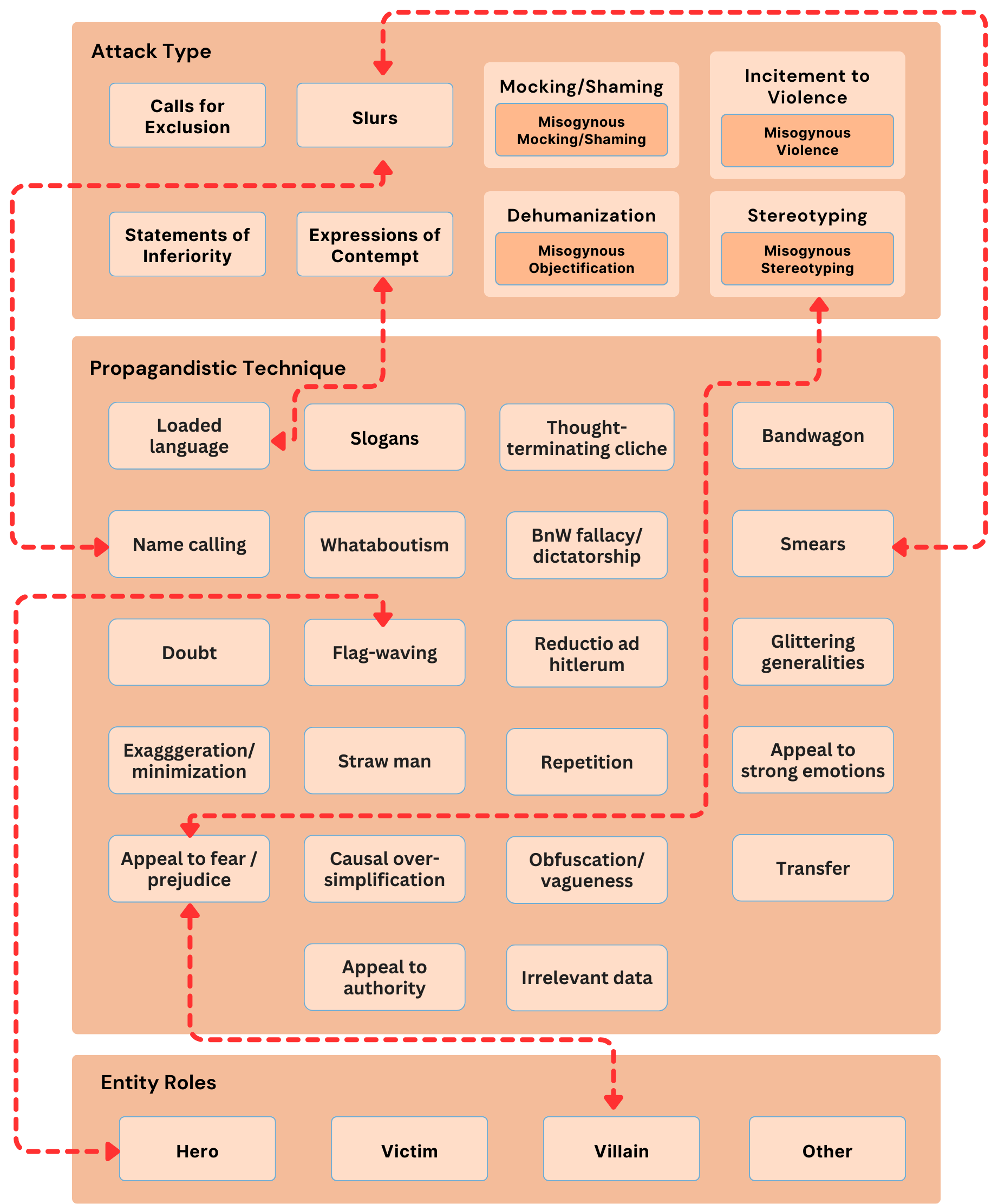}
\caption{While relationships between attack types, persuasion/propagandistic techniques, and entity roles have not been thoroughly explored, insights can be gleaned from their definitions, suggesting potential connections, equivalences, and other relationships (illustrated by red dotted lines).}
    \label{fig:tactics}
\end{figure}

\begin{figure}[!ht]
    \centering
    \includegraphics[width=1\linewidth]{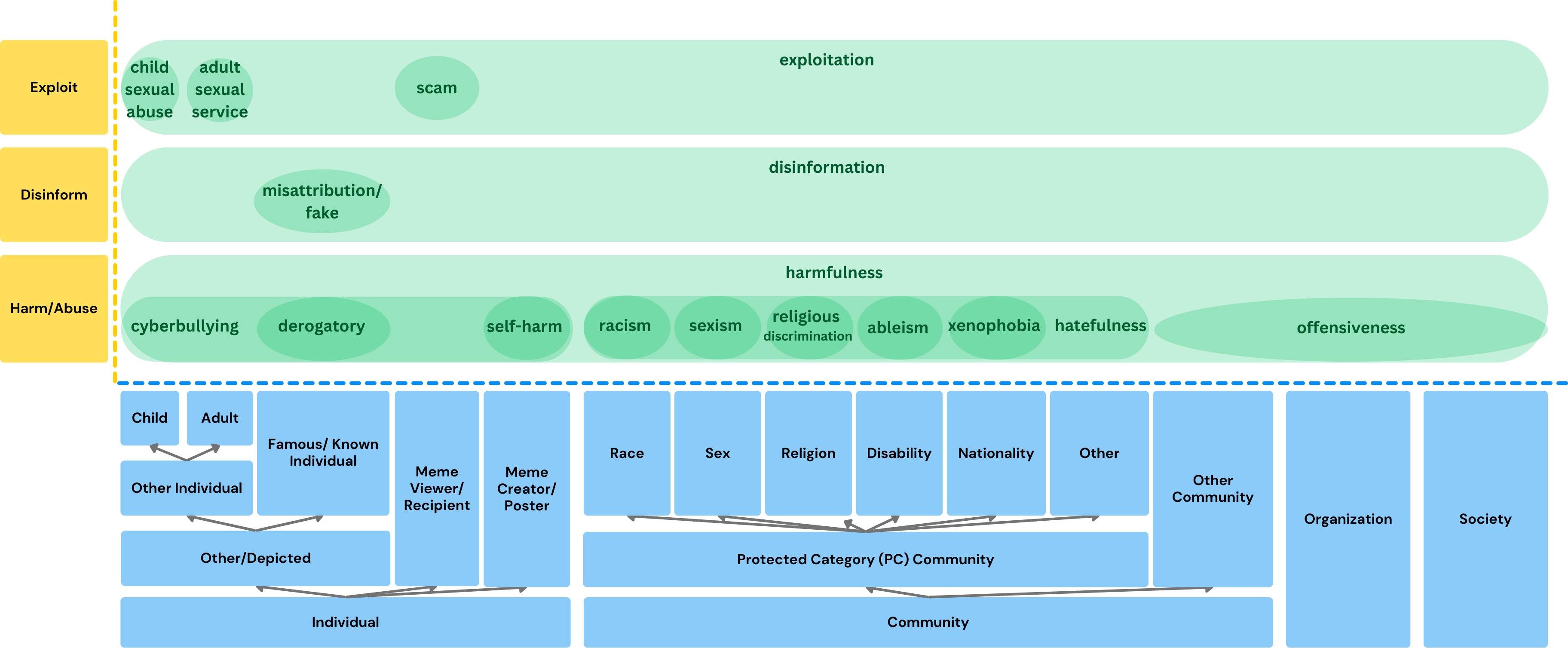}
\caption{Intersections between intent and target dimensions hint at specific types of meme toxicity. The y-axis, highlighted in yellow, categorically presents the three identified intents. Meanwhile, the x-axis, depicted in blue, categorizes target types with varying levels of specificity (from less specific at the bottom to more specific at the top). The green plane illustrates the toxicity landscape observed through the lens of intent and target dimensions, with greener areas indicating more specific toxicity types determined by particular intersections of intent and target.}
    \label{fig:intent_x_target}
\end{figure}

Overall, these three aspects of memes—attack types, persuasion techniques, and entity roles—are all related to the rhetorical strategies used to convey meme toxicity, yet they have been studied separately thus far. However, more relationships may exist between specific attack types, persuasion techniques, and entity roles than previously recognized (see Figure \ref{fig:tactics}). For instance, the attack type \textit{slurs} may be equivalent or strongly related to the persuasion techniques of \textit{name calling} and \textit{smears}, as they all involve the use of prejudicial terms to influence perception. Similarly, the attack type of \textit{contempt} may be equivalent or related to the persuasion technique of \textit{loaded language}, as they both entail the use of strongly charged emotional language. Additionally, there may be specific attack types or persuasion techniques employed to portray certain entities as heroes or villains. Furthermore, as noted in Section \ref{subsec:annotations}, some toxic meme datasets include labels for additional features that may be part of the rhetorical strategy for conveying toxicity in memes: emotion \cite{maity2022multitask, jha2024meme, xu2022met, kumari2023emoffmeme, sharma2020semeval, ramamoorthy2022memotion}, sentiment \cite{sharma2020semeval, ramamoorthy2022memotion, das2023banglaabusememe, maity2022multitask}, humour \cite{sharma2020meme, ramamoorthy2022memotion}, irony/sarcsam \cite{gasparini2022benchmark, das2023banglaabusememe, maity2022multitask, lin2024goat, sharma2020semeval, ramamoorthy2022memotion}, metaphor \cite{xu2022met}, profanity/vulgarity \cite{bhowmick2022multimodal, das2023banglaabusememe}, and more.  These connections remain largely unexplored, underscoring the need for a comprehensive examination of these aspects within the rhetorical dimension.



\subsection{Dimensions' Intersections}

\paragraph{Intent x Target} While examining individual dimensions of meme toxicity is valuable, we propose that simultaneous analysis of multiple dimensions holds untapped potential, especially for tasks like automatic detection and explanation of toxicity. We particularly emphasize the value of identifying intent and target dimensions concurrently, as this can offer insights into the specific types of toxicity in memes (see Figure \ref{fig:intent_x_target}). This visual representation highlights the types of toxicity typically arising from specific combinations of targets and intents. For example, when the intent is to harm and the target is people with disabilities, this may result in ableism, a type of toxicity that is both hateful and harmful. It is worth noting that while we have included toxicities explicitly addressed or mentioned in the surveyed literature, this depiction may not be comprehensive. There may be manifestations of toxicities that combine all types of intent with all types of targets, suggesting that all parts of the two-dimensional plane can potentially be populated. This implies that while these types of memes may exist, they have not yet been categorized under a specific type of toxicity.

%% file: src/trends.tex
\subsection{Trend 1: Tackling Cross-Modal Entailments and Reasoning}

Until recently, AI models for identifying toxic or hateful speech mainly relied on single-mode classifiers or text-limited datasets (e.g., \cite{davidson2017automated, waseem2017understanding}). However, there's been advocacy in social sciences for a multi-modal critical discourse approach, stressing the importance of understanding meanings constructed through various sign systems, including language and visuals \cite{mirzoeff2023white, sekimoto2023race, polli2023multimodal}. Memes, characterized by their co-creation of meaning through text and image, present significant hurdles for machine interpretation because of the complexity of multimodal understanding \cite{yus2019multimodality, knobel2007online, shifman2013memes}. The Hateful Memes Challenge at NeurIPS 2020 \cite{Kiela2020hateful_challenge} played a key role in revealing deficiencies in recent AI models lacking holistic, multi-modally informed reasoning. The challenge spurred significant efforts to deepen the multi-modal understanding of memes, leading to increased use of multi-modal deep learning models that integrate information from specialized neural networks analyzing specific modalities. This integration, often achieved through fusion techniques, enables a more comprehensive analysis and decision-making process \cite{hermida2023detecting}. Despite advancements, the performance of toxic meme detection algorithms remains suboptimal
and many memes evade filters due to their complex, multi-modal nature \cite{chakraborty2022nipping}.

\begin{figure}[!ht]
    \centering
    \includegraphics[width=0.65\linewidth]{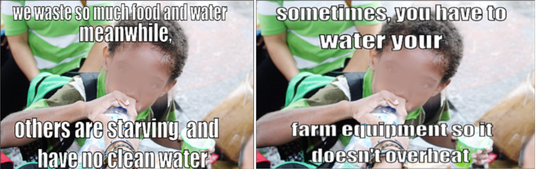}
    \caption{Illustration of how altering text modifies the interpretation of an image. The left meme conveys a message about water crisis awareness. The right meme alters the text, resulting in a hateful interpretation of the child as ``farm equipment'', perpetuating dehumanization and racism. We blurred the face of the child for privacy concerns. \textcolor{red}{Caution: The depicted memes contain toxic content. Viewer discretion is advised. These images do not reflect the views of the authors.}}
        \label{fig:polli_farm}
\end{figure}

In their recent study, Polli et al. (2023) \cite{polli2023multimodal} emphasize the complex interplay between textual and visual elements in toxic meme interpretation, revealing persistent challenges in automated detection beyond current computational models' capabilities. Drawing from sociosemiotics and critical multimodal studies, they demonstrate that meaning-making in hateful memes defies unimodal determination or basic multimodal fusion used in most computational approaches. Analyzing examples, they illustrate how seemingly harmless elements can combine to create toxic outcomes. For instance, in Figure \ref{fig:polli_farm}, while both examples feature a non-hateful image, the second example becomes hateful and toxic due to text implicitly objectifying and dehumanizing the child, and thus perpetuating racist stereotypes. This example highlights the complex cross-modal reasoning needed to detect the toxicity of the meme: given the absence of explicit racial references in the text, AI-driven language-based classification systems struggle to identify it as hateful. These findings underscore the shortcomings of classification systems relying solely on text keywords or image features. Consequently, Polli et al. advocate for rejecting the simplistic view of single modalities, and instead advocate for computational models informed by semiotic and multimodal approaches that prioritize multiplicative meanings \cite{lemke1998metamedia}.

In this survey, we observe a rise in the exploration of sophisticated multimodal approaches to tackle the complex interplays between different modes in memes. Researchers are increasingly focusing on understanding the significance of each modality for specific meme types, experimenting with combining and exploring unimodal and multimodal learning and representations. For instance, in \cite{Ma2022Hateful}, hateful meme detection is approached via a multi-task learning method for hateful memes detection, comprising a primary multimodal task and two unimodal auxiliary tasks, while the authors in \cite{aggarwal2024text} delve into how textual and visual components contribute differently to hateful meme detection, shedding light on the nuanced interactions between modalities. Meanwhile, the authors in \cite{Yang2023Invariant} propose a representation framework that facilitates inter-modal interaction and dynamically balances inter-modal and intra-modal relationships, providing a systematic way to disentangle memes into modality-invariant and modality-specific spaces. 

Attention frameworks are increasingly prevalent in cross-modal research for meme analysis. For instance, \cite{hossain2023inter} employs an inter-modal attention framework to detect offensive memes by synergistically fusing visual and textual information, while the authors in \cite{Liang2022TRICAN} use a cross-attention network to explore connections between visual modalities (using object detection and image caption models) and text features (using OCR) for hateful memes detection. Similarly, \cite{chen2023multimodal} proposes a Multimodal Visual-Textual Object Graph Attention Network (MViTO-GAT) for detecting propaganda techniques in memes. Their model learns semantic and positional relationships between visual and textual objects through attention-based sequential intra-modality and cross-modality graph reasoning, outperforming state-of-the-art baselines in performance. Many of the works we reviewed adopt CLIP embeddings, which are representations learned from diverse image-text pairs, providing meaningful embeddings for both images and text \cite{radford2021learning}. For example, \cite{jha2024meme} utilizes CLIP representations and modality-specific gating mechanisms to manage the interaction between textual and visual data. Similarly, the HateCLIPper architecture \cite{Kumar2022Hate-CLIPper} explicitly models cross-modal interactions between image and text representations obtained using CLIP encoders and a feature interaction matrix to learn meaningful concepts. Also, \cite{Qu2023On} systematically analyzes semantic regularities in CLIP-generated embeddings, allowing for the study of how hateful memes evolve by fusing visual elements from multiple images or combining text with a hateful image.

Additionally, we notice that an increasing number of works explicitly focus on capturing more complex or nuanced connections between different modalities and discourse-intensive modeling of complex linguistic phenomena. For example, \cite{koutlis2023memefier} propose the MemeFier approach, which employs a dual-stage modality fusion strategy to capture nuanced connections between text and image features. Leveraging Transformer encoders, MemeFier effectively learns inter-modality correlations at the token level, contributing to a deeper understanding of multimodal data. \cite{Zhang2023TOT} introduces Topology-Aware Optimal Transport (TOT), a framework designed to handle complex cross-modal interactions by formulating optimal transportation plans and leveraging topology information for representation alignment. \cite{shang2021aomd} develops the Analogy-aware Offensive Meme Detection (AOMD) framework, which focuses on accurately detecting offensive analogy memes. By capturing implicit analogy and aligning complex analogies across different modalities, AOMD achieves significant performance improvements over existing methods. Also,  \cite{liu2022figmemes} present FigMemes, a dataset tailored for figurative language classification in politically-opinionated memes, providing benchmark results for both unimodal and multimodal models, while \cite{wu2024fuser} utilize brain-like perceptual integration to reason about the subtle metaphors behind memes, exemplifying the trend toward more discourse-intensive modeling of cross-modal interplays.

\subsection{Trend 2: Tackling Contextuality, Cultural and Background Knowledge}

Memes are characterized by their intertextuality, often referencing elements from popular culture, symbols, events, or artifacts that hold significance within specific communities or affinity spaces \cite{knobel2007online}. This intertextuality renders memes highly contextual, necessitating background knowledge on various topics, including politics, current events, and cultural references, for accurate interpretation. This background knowledge, often referred to as ``meme literacy'' or ``prior knowledge'', is fundamental for comprehending the nuanced meanings embedded within memes \cite{Bi2023You}. In the realm of \textit{toxic} memes, contextualization is even more critical: a seemingly innocuous image can take on a malicious connotation simply with a shift in context, which can completely change the interpretation and impact of the
content.\footnote{For instance, the Nesquik bunny, originally featured in public information and awareness campaigns, was repurposed and recontextualized into a disparaging humor meme mocking the African water crisis \cite{polli2023multimodal}.} Therefore, the direct incorporation of external knowledge into the classification process is recently emerging as a promising strategy to enhance harmful meme detection effectiveness and improve real-world applicability \cite{Grasso2024KERMIT}. 

A prominent trend involves combining named entity recognition (NER) with background knowledge linking. The authors in \cite{zhong2022combining} incorporate conceptual information from the external knowledge base Probase \cite{wu2012probase}, a large-scale knowledge base that provides isA semantic relations, to enhance semantic representation and allows their model, MeBERT, to retrieve relevant concepts from meme text. 
This conceptual information is used as an extra modality to bridge the meme text to the image, using a concept-image attention module to align the concepts with corresponding image regions. This approach results in a concept-sensitive visual representation, where important image regions receive more attention, and significantly improves meme classification performance. However, challenges remain with text-only memes and long-tail entities with sparse information in the knowledge base. The authors in \cite{kougia2023memegraphs} enhance meme classification with their MemeGraphs method, which builds upon but goes beyond \cite{zhong2022combining} by using automated NER and knowledge base (KB) augmentation not only on meme text but also on text extracted by first transforming memes into scene graphs. This process retrieves background knowledge for each identified entity from Wikidata, concatenates these augmentations to the meme text, and then feeds it into a Transformer for text classification. By integrating structured representations and interactions between internal entities and external knowledge, the MemeGraphs method consistently improves classification performance compared to models that rely solely on learned representations.

Other works utilize ConceptNet as the background knowledge base. \cite{shang2021knowmeme} introduce KnowMeme, which constructs a graph representing meme content and related knowledge retrieved from ConceptNet, and then conducts graph classification to determine whether the meme is harmful. By capturing implicit meanings and cross-modal relations in memes, KnowMeme achieves significant performance enhancements compared to baseline methods. Similarly, \cite{Grasso2024KERMIT} introduce KERMIT (Knowledge-Empowered Model), which builds a meme's knowledge-enriched network by merging internal meme entities with relevant external knowledge from ConceptNet. It starts by extracting the meme's text and caption using OCR and BLIP. Graph extraction involves identifying nodes (primarily nouns) from the text and caption through POS tagging, establishing relationships between them via dependency parsing, simplifying these relationships, and merging dependency trees to connect related nodes from the text and caption to form the final meme graph. For knowledge enrichment, KERMIT leverages ConceptNet iteratively. Subsequently, KERMIT employs a dynamic learning mechanism that leverages memory-augmented neural networks and attention mechanisms to discern the most informative segment of the knowledge-enriched information network to accurately classify harmful memes. 

A key trend is the development of meme-specific external knowledge bases.The Image Meme Knowledge Graph (IMKG), recently published by \cite{tommasini2023imkg}, is a groundbreaking tool for studying Internet memes, providing a comprehensive and structured repository of meme-related semantics in text, vision, and metadata. It integrates data from diverse sources, enriches information through entity extraction and semantic links, and follows Semantic Web principles for effective knowledge representation. The authors used Wikidata \cite{vrandevcic2014wikidata} to extract background knowledge about meme seeds and entities, and KnowYourMeme (KYM) to collect and catalog meme-related data such as lore, interpretations, historical context, origins, and popularity. For textual enrichment, they used DBpedia Spotlight \cite{auer2007dbpedia} to extract entities from KYM paragraphs (including sections about, origin, and spread) and from ImgFlip captions, linking these DBpedia entries to Wikidata entities. Additionally, they employed the Google Vision API for visual enrichment to detect objects and link them to Freebase \cite{bollacker2008freebase}. While IMKG lacks image data and does not utilize the graph for downstream tasks, the authors explicitly suggest incorporating IMKG into methods to enhance the accuracy and explainability of neuro-symbolic methods for internet memes, such as hate speech detection and classification. The recent work by \cite{bates2023template} introduces KYMKB, a knowledge base with 54,000 meme-related images and detailed information, notably focusing on meme templates. It distinguishes between templatic and non-templatic memes, a first in AI literature. KYMKB provides extensive data about templatic memes, including title, meaning, and origin, making it valuable for various tasks. The authors employ KYMKB for offensive meme detection, achieving a promising accuracy.

\subsection{Trend 3: Interpretability, Explanations and Meme Literacy}

We have observed a growing emphasis on interpretability in toxic meme detection, with many of the works we surveyed acknowledging it as a crucial aspect \cite{Grasso2024KERMIT, sharma2022findings, Aggarwal2023HateProof, kougia2023memegraphs, alzubi2023multimodal}. This trend seems to be part of a broader movement towards developing complementary measures alongside algorithms for automatic detection and removal of harmful content \cite{scheuerman2021framework}. Specifically, there's a critical trend emerging where works focus on providing comprehensive rationales or explanations to aid users and content moderators in understanding the nuances of toxic memes \cite{hee2023decoding}. This shift reflects a growing recognition of the importance of enhancing users' understanding of harmful content, particularly in the context of memes, to improve media literacy \cite{Bi2023You}. While straightforward indicators like nudges, labels, and red flags have been suggested for identifying hateful memes, their efficacy remains largely untested (\cite{jahanbakhsh2021exploring}).

To enhance the interpretability of meme toxicity, recent trends include explicit identification of targeted entities, exploration of underlying themes and similarities, and utilization of pre-trained language models for enhanced analysis and explanation. The explicit identification of entities targeted by toxicity increasingly approached automatically, is exemplified in DisMultiHate (\cite{Lee2021Disentangling}), which disentangles target entities within multimodal memes to improve the classification and explainability of hateful content, as well as in DISARM (\cite{sharma2022disarm}), which employs named entity recognition and person identification for this purpose. Recent research efforts also aim to provide insights into underlying themes and similarities, such as \cite{thakur2023explainable}, which proposes modular and explainable architectures for meme understanding using example- and prototype-based reasoning and the Hate-CLIPper approach by \cite{Kumar2022Hate-CLIPper}, which illuminates underlying themes and similarities across modalities.  Furthermore, a recent trend in works addressing toxic memes involves enhancing interpretability by leveraging pre-trained language models, such as BERT and RoBERTa, to analyze linguistic patterns and contextual cues and generate insights into why certain content is classified as sexist \cite{obeidat2023just_one} and to perform abductive reasoning to explore the interplay of multimodal information in memes (\cite{lin2023beneath}).


Critically, recent datasets include ground truth explanation annotations for memes, as noted in Section \ref{subsec:annotations}. For instance, the authors in \cite{Bi2023You} propose quality-controlled crowdsourcing as an effective strategy for offering explanations and background knowledge for hateful memes through a Generate-Annotate-Revise workflow. The MultiBully-Ex dataset \cite{jha2024meme} provides multimodal explanations, combining visual cues like image segmentation with textual cues such as words relevant to or explaining the cyberbullying. Their experimental results demonstrate that training with multimodal explanations improves performance in generating textual justifications and accurately identifying visual evidence. Additionally, in the HatReD dataset \cite{hee2023decoding}, hateful memes are annotated with textual explanations, and models are trained to decode the meaning of multimodal hateful memes and provide explicit explanations for the classifications.

\subsection{Trend 4: Low-Resource Languages}

As social media continues to expand globally, the spread of harmful content, including toxic memes, transcends linguistic and cultural boundaries. While considerable attention has been given to detecting such content in English, there is a growing realization of the urgency to address this issue in low-resource languages and contexts. Researchers have proposed integrating state-of-the-art deep learning models, such as BERT and Electra, for multilingual text analysis, alongside face recognition and optical character recognition models for comprehending meme images \cite{bhowmick2022multimodal}. Additionally, Bengali BERT models have been deployed for automated Bengali abusive text classification, aiming to streamline the hate speech filtering process in resource-constrained languages, achieving notable accuracy and performance \cite{titli2023automated}. In low-resource settings, multi-modal prompt tuning has emerged as an effective approach for detecting propaganda techniques in memes. This method incorporates visual cues into language models through prompt-based multi-modal fine-tuning, showcasing efficacy in resource-limited scenarios \cite{wu2022propaganda}. Furthermore, transfer learning techniques have been explored to extend abusive meme detection to multiple languages. By leveraging model transfer techniques, researchers aim to bridge the language gap and establish baseline models for detecting abusive memes \cite{das2023transfer}. Dataset creation is also a significant effort, particularly in Asian languages such as Hindi, Bengali, Tamil, and Chinese. Efforts have been made to address the lack of benchmark datasets for specific languages by creating resources like the BanglaAbuseMeme dataset for Bengali abusive meme classification. These datasets facilitate the development and evaluation of models for detecting abusive content in low-resource languages (\cite{das2023banglaabusememe}).

\subsection{Trend 5: Generative AI, Large Language Models (LLMs)}

The emergence of Large Language Models (LLMs) and generative artificial intelligence (AI) has opened up new avenues for detecting and interpreting toxic memes. These advanced models offer capabilities for understanding implicit meanings and context in multimodal content, thereby addressing the challenges posed by harmful memes. In \cite{lin2023beneath}, LLMs are utilized for abductive reasoning to detect harmful memes. By distilling multimodal reasoning knowledge from LLMs and conducting lightweight fine-tuning, the model demonstrates effectiveness in identifying toxic content. Similarly, \cite{lin2024towards} explores the use of multimodal debate between LLMs for explainable harmful meme detection. By facilitating a debate between LLMs, the model gains insights into the context and implicit meanings of memes, leading to more interpretable results and a deeper understanding of meme content. Another notable application is in the correction of hate speech within multimodal memes, as proposed by \cite{van2023detecting}. Leveraging a large visual language model, the method effectively detects and corrects hate speech, contributing to the mitigation of online toxicity. Furthermore, \cite{wang2023gpt} investigates the capability of GPT models to analyze the emotions conveyed through memes. By analyzing meme content using GPT models, the study provides insights into the emotional content and context of memes, showcasing the potential of LLMs for emotion analysis tasks. Additionally, \cite{zhao2023review} offers a comprehensive review of vision-language models and their performance on the Hateful Memes Challenge. Through an analysis of different models and techniques, valuable insights are provided for future research in the domain of detecting hateful memes. \cite{Cao2022Prompting} propose PromptHate, a prompt-based model that leverages pre-trained language models for hateful meme classification. By constructing prompts and providing context examples, PromptHate exploits implicit knowledge in pre-trained models to achieve high classification accuracy, showcasing the potential of leveraging external knowledge for improved classification. Finally, \cite{jain2023generative} introduces a unified Multimodal Generative framework (MGex) for detecting cyberbullying in memes. This framework reframes the problem of meme detection as a multimodal text-to-text generation task, achieving competitive performance against baselines and state-of-the-art models. We also note that the growing number of guardrails in LLMs to prevent unsafe use will likely limit their ability to generate and analyze unsafe content in the future. Many examples in toxic datasets trigger these guardrails, designed to mitigate the processing of hateful or offensive content (\cite{kumarage2024harnessing}).

A related trend is the investigation of text-to-image models for the generation of unsafe images and hateful memes. In \cite{qu2023unsafe}, the focus is on demystifying the generation of unsafe images and hateful memes from Text-to-Image models. The study assesses the proportion of unsafe images generated by advanced Text-to-Image models and evaluates the potential of generating hateful meme variants, highlighting risks associated with model misuse. Furthermore, \cite{wu2023proactive} explores the proactive generation of unsafe images from Text-To-Image models using benign prompts. By studying the generation process, the research highlights potential risks associated with model misuse and emphasizes the importance of implementing safety measures to mitigate the generation of harmful content.


%% file: src/towards.tex
This survey serves as a roadmap for researchers seeking to understand the landscape of computational perspectives on toxic internet memes. It focuses on multimodal toxic meme analysis and sheds light on the nuanced, complex taxonomical relationships within harmful online content (see Section \ref{sec:background}). Our examination of existing surveys on computational toxic meme analysis emphasized the need for an up-to-date survey on toxic memes to address critical gaps in the literature while highlighting emerging trends and areas of research in Section \ref{sec:related}. Employing the PRISMA 2020 guidelines (see Section \ref{sec:methodology}), we surveyed 158 papers from 2019 to 2024, filling a crucial gap in the literature.

\subsection{Contributions and Implications}

This survey provides insights to support the advancement of computational models and tools for detecting, analyzing, and mitigating the proliferation of toxic memes in online environments. 
The theoretical implications of our study are described in Sections~\ref{sec:meme_toxicities} and ~\ref{sec:dimensions}, where we harmonize the definitions of meme toxicities and provide a framework to identify their constituting elements. 
We identified and harmonized 12 meme toxicity terms, noting a significant focus on hateful memes and gaps in research on less prominent categories like troll, derogatory, and disinformation memes. We developed a taxonomy to clarify explicit and implicit taxonomical relationships among these terms, addressing discrepancies and enhancing adaptability for future studies. This offers a structured framework for comprehensively understanding and classifying various types of toxic content found in memes. This taxonomy serves as a valuable resource for researchers seeking to analyze, categorize, and compare different toxic memes. Furthermore, our taxonomy addresses discrepancies and nuances in existing frameworks, enabling researchers to guide their investigations effectively and explore the complex relationships between different toxicity categories with clarity and precision.
This resulting framework has also significant practical implications since it identifies complex features that may be used by developers and practitioners to automatically or semi-automatically analyze memes to classify their toxicity. The datasets described and compared in Section 6 provide a map for possible training sets to be used to create such (semi-)automated models.

Another practical contribution is the synthesis of annotation guidelines across diverse toxic meme datasets detailed in Section 6.2. 
By characterizing the annotation methodologies and the range of computational task definitions 
utilized across various datasets for different meme attributes, we furnish researchers with a useful tool to inform their future investigations. 
Furthermore, this resource empowers researchers to pose targeted inquiries about meme attributes, including assessing the efficacy of different annotation schemes in capturing nuanced meme characteristics.

Meme toxicity definitions predominantly focus on three core elements: the targeted entity, the underlying intent behind meme creation or dissemination, and the rhetorical techniques used (see Section \ref{sec:dimensions}).  Building upon these observations, we pinpoint three key dimensions of meme toxicity—target, intent, and rhetorical tactics—that are frequently examined in isolation within computational studies. 
Researchers can utilize these dimensions and the relationships among them to systematically categorize and analyze various types of toxic content found in memes. 

In Section \ref{sec:issues_trends}, we identified trends and research directions in automatic toxic meme detection and understanding. These trends encompass tackling complex cross-modal entailments by leveraging multi-modal deep learning models, attention frameworks, and multidisciplinary approaches. Also, there is a focus on addressing contextuality and cultural background knowledge through named entity recognition and connection with commonsense knowledge bases, and the development of meme-specific external knowledge bases. We observed an emphasis on interpretability and explainability. Furthermore, attention is directed towards low-resource languages and the potential of generative AI and Large Language Models (LLMs) for detecting and interpreting toxic memes.

\subsection{Future Research Directions}

The recent emphasis on providing explanations alongside toxic meme detection marks a shift towards more transparent and accountable AI systems. However, there is a need for further clarification regarding the types of interpretability required and the appropriate methods for achieving them. It is crucial to differentiate between methods ensuring transparency of computation (e.g., explaining what the machine computed) and those providing users with understandable explanations for the toxicity label assigned to a meme (e.g., explaining why a meme is considered propagandistic). We conclude that the field is shifting to fulfill the latter—offering reasonable explanations for assigned labels. Interpretable methods help build trust in automatic moderation systems, ensuring that toxicity labels are properly explained. Based on the contributions identified, we propose the following future research directions:

\paragraph{Semiotics-Informed Cross-Modal Reasoning} Leveraging insights from semiotic research can significantly enhance the development of an advanced automatic understanding of cross-modality. Interdisciplinary collaboration with semiotics experts can operationalize these insights, improving AI-driven meme analysis systems. Future research should focus on advancing these models to effectively capture the multiplicative nature of meaning-making in multimodal content. This includes exploring semiotically relevant features with feature engineering and utilizing advanced fusion techniques like modality-specific gating and attention mechanisms to balance inter-modal and intra-modal relationships. By developing cross-modal reasoning models, researchers can enable machines to perform critical semiotic-based analyses, providing valuable insights into toxic message construction and dissemination. Using these insights in explanations, automatic systems can help users understand how toxic messages are constructed and conveyed in multimodal content. The integration of semiotic insights can empower content moderators and users with clues about toxic message construction, enhancing AI interpretability and informed decision-making. Prioritizing semiotic-informed cross-modal reasoning models can create more effective and transparent solutions for addressing toxic content.

\paragraph{Incorporating Toxicity-Specific Background Knowledge}Building upon the current trend of integrating external knowledge bases into meme analysis, there are several promising research directions to explore. While resources like ConceptNet, WikiData, and Probase have enhanced meme interpretation, a critical gap remains: incorporating databases with cultural knowledge specifically related to toxicity. For instance, the Global Extremist Symbols Database\footnote{\url{https://globalextremism.org/global-extremist-symbols-database/}} provided by the Global Project Against Hate and Extremism (GPAHE) offers a wealth of information on cultural icons associated with toxicity and hate, such as hate symbols, visual and numerical icons, flags, hand gestures, acronyms, and salutes used by extremist groups. Research efforts could focus on integrating such specialized knowledge bases into meme analysis frameworks and exploring hybrid approaches that combine external knowledge with pre-trained vision-language models, similar to KERMIT \cite{Grasso2024KERMIT}, to enhance classification performance by leveraging the strengths of both approaches. Additionally, given the dynamic nature of online content, developing adaptive learning mechanisms that continuously update knowledge representations to reflect evolving trends and cultural contexts would be beneficial. Overall, by broadening the scope of incorporated knowledge, future research can enhance the contextual understanding and analysis of memes within their socio-cultural contexts, leading to more effective meme classification.

\paragraph{Embracing Linguistic Diversity in Toxic Meme Detection}Expanding toxic meme detection beyond English to include other widely spoken languages is imperative, along with annotating datasets from diverse linguistic and cultural backgrounds. While leveraging models trained in English may provide a foundation, incorporating annotations from other languages is crucial to ensure the effectiveness and cultural relevance of detection methods. Additionally, exploring novel approaches that consider the unique linguistic and cultural characteristics of different languages can enhance the accuracy and applicability of toxic meme detection systems. Further work should investigate the transferability of existing models across languages and develop techniques to adapt them to new linguistic contexts, as embracing linguistic diversity and cultural nuances in toxic meme detection research can lead to more robust and inclusive frameworks.

\paragraph{Leveraging LLMs}Recent research highlights the potential of Large Language Models (LLMs) and generative AI in identifying and interpreting toxic memes. However, further investigation is needed to understand their strengths and weaknesses in meme detection, including their ability to differentiate between harmless and toxic memes and any inherent cultural or modal biases. Additionally, exploring LLMs' capacity for cross-modal analysis of meme content with multimodal data containing nuanced cross-modal interplays is crucial. Future research should also focus on assessing LLMs' cultural knowledge and understanding, exploring efficient methods like Retrieval-Augmented Generation (RAG) for incorporating background knowledge. Understanding how LLMs can adapt to evolving cultural references and incorporate diverse perspectives in meme interpretation is crucial. Equally important is further investigating the effectiveness and ethical implications of safeguard rails in LLMs for toxic meme detection.

\paragraph{Multi-Dimensional Exploration of Meme Toxicity} Recognizing that, for AI to detect multimodal toxicity, ``it must learn to understand content the way that people do: holistically'' \cite[p. 201]{woah2021shared}, we advocate for a multidimensional approach. Future research should focus on simultaneously analyzing multiple dimensions using an evidential reasoning approach grounded in decision theory, allowing for the accumulation and scrutiny of evidence to guide judgments or decisions. Additionally, there is a need to explore potential common relationships between meme toxicity types and the tactics used to convey them. Furthermore, further investigation into additional dimensions, such as the context of posting (user, forum, platform) and propagation features \cite{beskow2020evolution}, is warranted, as these factors can provide a more comprehensive understanding of meme toxicity dynamics.

\paragraph{Legal, Ethical and Collateral Impacts} Exploring legal and ethical considerations in automatic moderation is crucial. Automatic moderation systems must navigate legal and ethical landscapes to discern illegal content from harmful but legal content. Determining computational tasks based on content legality and harm potential raises complex ethical dilemmas that require careful consideration. Further research is also needed to address other issues, such as collateral damage in moderation algorithms. Evaluation of these algorithms should account for false positives and the potential unintended consequences they may have, particularly concerning the censorship of speech from marginalized groups or the unintentional suppression of resistance narratives. Other ethical challenges in meme analysis include addressing subjectivity, combating the use of AI to generate toxic content, and navigating the complexities of toxic positivity.

%% file: src/conclusion.tex
Toxic memes represent a spectrum of harmful or malevolent multimodal content disseminated across online platforms, often with the intention of promoting harm and hate, spreading disinformation, or promoting exploitation. Understanding the conceptualizations and distinguishing features of toxic memes is crucial for developing effective computational models capable of detecting and moderating such content. Our survey provides readers with a comprehensive understanding of toxic memes from a computational, content-based perspective, covering key developments up to early 2024. Our survey identified that the computational field has used a wide variety of terminology to refer to toxic memes, highlighting the increasing demand for automatic tools for identifying fine-grained toxicity types beyond a simple toxic/non-toxic classification. This includes specifying whether a meme is misogynistic, spreading disinformation, involved in cyberbullying, and so on. This complexity necessitates harmonizing term definitions and the establishment of a clearer taxonomy that delineates how these terms relate to each other. In response to this need, we provide a harmonized set of definitions and introduce a novel taxonomy in Section \ref{sec:meme_toxicities}. We offer insights into various dimensions of meme toxicity, including intent, target, and conveyance tactics, along with a standardized taxonomy for categorizing meme toxicity types. Also, we catalog datasets containing toxic memes, analyze prevalent challenges, and identify emerging trends in computational approaches to toxic meme detection and interpretation. Through synthesizing existing knowledge and identifying research gaps, we aim to promote interdisciplinary collaboration and innovation to foster media literacy and potentially a safer, higher quality, and more inclusive online ecosystem. Our survey offers pathways for computational advancement in the field, including enhancing interpretability through sophisticated cross-modal reasoning, background knowledge integration, attention on low-resource languages, and refining the usage of LLMs.

%% file: src/appendix.tex
\label{sec:appendix:venn}

\begin{figure}[!ht]
\centering
\includegraphics[width=0.53\textwidth]{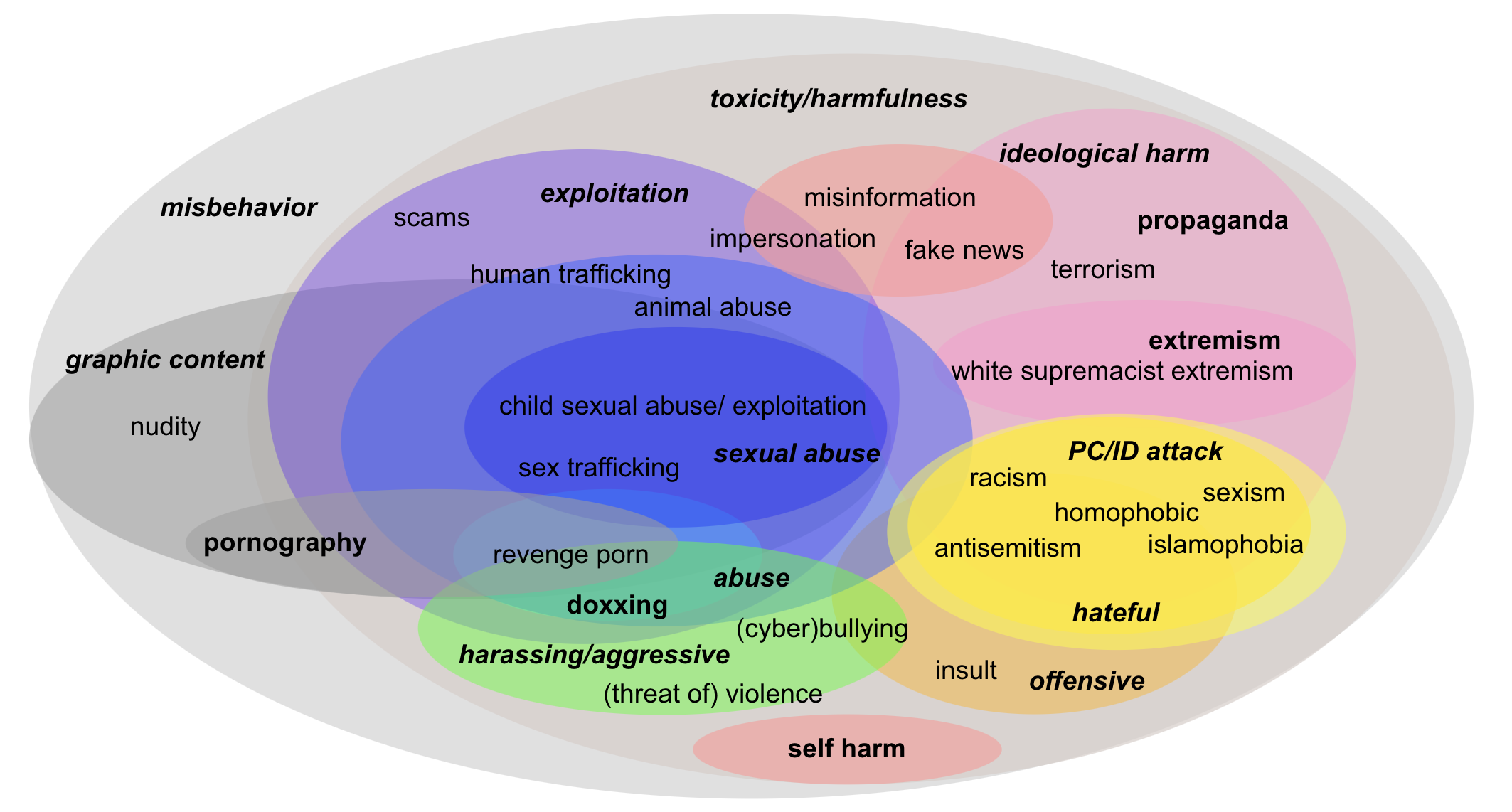}
\caption{Toxicity-related terms derived from our investigation of harmfulness and toxicity in multimodal data, illustrating the complex and overlapping nature of multimodal toxicities.}
\label{fig:venn}
\end{figure}
\begin{table}[!ht]
\scriptsize
\begin{tabular}{|l|{c}|{c}|{c}|{c}|{c}|{c}|{c}|{c}|{c}|{c}|{c}|{c}|{c}|{c}|{c}|{c}|{c}|}
\hline
\multirow{2}{*}{Dataset} &
  \multicolumn{4}{c|}{\textbf{Search Engines}} &
  \multicolumn{2}{c|}{\textbf{Img Platforms}} &
  \multicolumn{7}{c|}{\textbf{Social Media Platforms}} &
  \multicolumn{4}{c|}{\textbf{Meme Platforms}} \\ \cline{2-18} 
 &
  \multicolumn{1}{c|}{\textbf{B}} &
  \multicolumn{1}{c|}{\textbf{G}} &
  \multicolumn{1}{c|}{\textbf{D}} &
  \multicolumn{1}{c|}{\textbf{B}} &
  \multicolumn{1}{c|}{\textbf{Pin}} &
  \multicolumn{1}{c|}{\textbf{Img}} &
  \multicolumn{1}{c|}{\textbf{FB}} &
  \multicolumn{1}{c|}{\textbf{Gab}} &
  \multicolumn{1}{c|}{\textbf{IG}} &
  \multicolumn{1}{c|}{\textbf{RD}} &
  \multicolumn{1}{c|}{\textbf{TW}} &
  \multicolumn{1}{c|}{\textbf{Wei}} &
  \textbf{WA} &
  \multicolumn{1}{c|}{\textbf{KYM}} &
  \multicolumn{1}{c|}{\textbf{MD}} &
  \multicolumn{1}{c|}{\textbf{MG}} &
    \multicolumn{1}{c|}{\textbf{9gag}} \\ \hline
AOMD Gab &
  \multicolumn{1}{c|}{} &
  \multicolumn{1}{c|}{} &
  \multicolumn{1}{c|}{} &
   &
  \multicolumn{1}{c|}{} &
   &
  \multicolumn{1}{c|}{} &
  \multicolumn{1}{c|}{\checkmark} &
  \multicolumn{1}{c|}{} &
  \multicolumn{1}{c|}{} &
  \multicolumn{1}{c|}{} &
  \multicolumn{1}{c|}{} &
   &
  \multicolumn{1}{c|}{} &
  \multicolumn{1}{c|}{} &
  \multicolumn{1}{c|}{} &
   \\ \hline
AOMD Reddit &
  \multicolumn{1}{c|}{} &
  \multicolumn{1}{c|}{} &
  \multicolumn{1}{c|}{} &
   &
  \multicolumn{1}{c|}{} &
   &
  \multicolumn{1}{c|}{} &
  \multicolumn{1}{c|}{} &
  \multicolumn{1}{c|}{} &
  \multicolumn{1}{c|}{\textbf{\checkmark}} &
  \multicolumn{1}{c|}{} &
  \multicolumn{1}{c|}{} &
   &
  \multicolumn{1}{c|}{} &
  \multicolumn{1}{c|}{} &
  \multicolumn{1}{c|}{} &
   \\ \hline
BanglaAbuseMeme &
  \multicolumn{1}{c|}{\checkmark} &
  \multicolumn{1}{c|}{\checkmark} &
  \multicolumn{1}{c|}{} &
   &
  \multicolumn{1}{c|}{} &
   &
  \multicolumn{1}{c|}{\checkmark} &
  \multicolumn{1}{c|}{} &
  \multicolumn{1}{c|}{\checkmark} &
  \multicolumn{1}{c|}{} &
  \multicolumn{1}{c|}{} &
  \multicolumn{1}{c|}{} &
   &
  \multicolumn{1}{c|}{} &
  \multicolumn{1}{c|}{} &
  \multicolumn{1}{c|}{} &
   \\ \hline
CrisisHateMM &
  \multicolumn{1}{c|}{} &
  \multicolumn{1}{c|}{} &
  \multicolumn{1}{c|}{} &
   &
  \multicolumn{1}{c|}{} &
   &
  \multicolumn{1}{c|}{\checkmark} &
  \multicolumn{1}{c|}{} &
  \multicolumn{1}{c|}{} &
  \multicolumn{1}{c|}{\checkmark} &
  \multicolumn{1}{c|}{\checkmark} &
  \multicolumn{1}{c|}{} &
   &
  \multicolumn{1}{c|}{} &
  \multicolumn{1}{c|}{} &
  \multicolumn{1}{c|}{} &
   \\ \hline
Derogatory Fb-Meme &
  \multicolumn{1}{c|}{} &
  \multicolumn{1}{c|}{} &
  \multicolumn{1}{c|}{} &
   &
  \multicolumn{1}{c|}{} &
   &
  \multicolumn{1}{c|}{\checkmark} &
  \multicolumn{1}{c|}{} &
  \multicolumn{1}{c|}{} &
  \multicolumn{1}{c|}{} &
  \multicolumn{1}{c|}{} &
  \multicolumn{1}{c|}{} &
   &
  \multicolumn{1}{c|}{} &
  \multicolumn{1}{c|}{} &
  \multicolumn{1}{c|}{} &
   \\ \hline
DisinfoMeme &
  \multicolumn{1}{c|}{} &
  \multicolumn{1}{c|}{} &
  \multicolumn{1}{c|}{} &
   &
  \multicolumn{1}{c|}{} &
   &
  \multicolumn{1}{c|}{\checkmark} &
  \multicolumn{1}{c|}{} &
  \multicolumn{1}{c|}{} &
  \multicolumn{1}{c|}{} &
  \multicolumn{1}{c|}{} &
  \multicolumn{1}{c|}{} &
   &
  \multicolumn{1}{c|}{} &
  \multicolumn{1}{c|}{} &
  \multicolumn{1}{c|}{} &
   \\ \hline
ELEMENT &
  \multicolumn{1}{c|}{} &
  \multicolumn{1}{c|}{\checkmark} &
  \multicolumn{1}{c|}{} &
   &
  \multicolumn{1}{c|}{} &
   &
  \multicolumn{1}{c|}{} &
  \multicolumn{1}{c|}{} &
  \multicolumn{1}{c|}{} &
  \multicolumn{1}{c|}{\checkmark} &
  \multicolumn{1}{c|}{} &
  \multicolumn{1}{c|}{} &
   &
  \multicolumn{1}{c|}{} &
  \multicolumn{1}{c|}{} &
  \multicolumn{1}{c|}{\checkmark} &
   \\ \hline
Emoffmeme &
  \multicolumn{1}{c|}{} &
  \multicolumn{1}{c|}{\checkmark} &
  \multicolumn{1}{c|}{} &
   &
  \multicolumn{1}{c|}{} &
   &
  \multicolumn{1}{c|}{} &
  \multicolumn{1}{c|}{} &
  \multicolumn{1}{c|}{} &
  \multicolumn{1}{c|}{} &
  \multicolumn{1}{c|}{} &
  \multicolumn{1}{c|}{} &
   &
  \multicolumn{1}{c|}{} &
  \multicolumn{1}{c|}{} &
  \multicolumn{1}{c|}{} &
   \\ \hline
Ext-Harm-P &
  \multicolumn{1}{c|}{} &
  \multicolumn{1}{c|}{\checkmark} &
  \multicolumn{1}{c|}{} &
   &
  \multicolumn{1}{c|}{} &
   &
  \multicolumn{1}{c|}{\checkmark} &
  \multicolumn{1}{c|}{} &
  \multicolumn{1}{c|}{\checkmark} &
  \multicolumn{1}{c|}{\checkmark} &
  \multicolumn{1}{c|}{} &
  \multicolumn{1}{c|}{} &
   &
  \multicolumn{1}{c|}{} &
  \multicolumn{1}{c|}{} &
  \multicolumn{1}{c|}{} &
   \\ \hline
Facebook HM &
  \multicolumn{1}{c|}{} &
  \multicolumn{1}{c|}{} &
  \multicolumn{1}{c|}{} &
   &
  \multicolumn{1}{c|}{} &
   &
  \multicolumn{1}{c|}{\checkmark} &
  \multicolumn{1}{c|}{} &
  \multicolumn{1}{c|}{} &
  \multicolumn{1}{c|}{} &
  \multicolumn{1}{c|}{} &
  \multicolumn{1}{c|}{} &
   &
  \multicolumn{1}{c|}{} &
  \multicolumn{1}{c|}{} &
  \multicolumn{1}{c|}{} &
   \\ \hline
FAME dataset &
  \multicolumn{1}{c|}{} &
  \multicolumn{1}{c|}{} &
  \multicolumn{1}{c|}{\checkmark} &
   &
  \multicolumn{1}{c|}{} &
   &
  \multicolumn{1}{c|}{} &
  \multicolumn{1}{c|}{} &
  \multicolumn{1}{c|}{} &
  \multicolumn{1}{c|}{} &
  \multicolumn{1}{c|}{} &
  \multicolumn{1}{c|}{} &
   &
  \multicolumn{1}{c|}{} &
  \multicolumn{1}{c|}{} &
  \multicolumn{1}{c|}{} &
   \\ \hline
Fine grained HM &
  \multicolumn{1}{c|}{} &
  \multicolumn{1}{c|}{} &
  \multicolumn{1}{c|}{} &
   &
  \multicolumn{1}{c|}{} &
   &
  \multicolumn{1}{c|}{\checkmark} &
  \multicolumn{1}{c|}{} &
  \multicolumn{1}{c|}{} &
  \multicolumn{1}{c|}{\checkmark} &
  \multicolumn{1}{c|}{} &
  \multicolumn{1}{c|}{} &
   &
  \multicolumn{1}{c|}{} &
  \multicolumn{1}{c|}{} &
  \multicolumn{1}{c|}{\checkmark} &
   \\ \hline
GOAT-Bench &
  \multicolumn{1}{c|}{} &
  \multicolumn{1}{c|}{\checkmark} &
  \multicolumn{1}{c|}{} &
   &
  \multicolumn{1}{c|}{} &
  \checkmark &
  \multicolumn{1}{c|}{\checkmark} &
  \multicolumn{1}{c|}{} &
  \multicolumn{1}{c|}{\checkmark} &
  \multicolumn{1}{c|}{\checkmark} &
  \multicolumn{1}{c|}{\checkmark} &
  \multicolumn{1}{c|}{} &
   &
  \multicolumn{1}{c|}{\checkmark} &
  \multicolumn{1}{c|}{} &
  \multicolumn{1}{c|}{} &
  \checkmark \\ \hline
Harm-C &
  \multicolumn{1}{c|}{} &
  \multicolumn{1}{c|}{\checkmark} &
  \multicolumn{1}{c|}{} &
   &
  \multicolumn{1}{c|}{} &
   &
  \multicolumn{1}{c|}{\checkmark} &
  \multicolumn{1}{c|}{} &
  \multicolumn{1}{c|}{\checkmark} &
  \multicolumn{1}{c|}{\checkmark} &
  \multicolumn{1}{c|}{} &
  \multicolumn{1}{c|}{} &
   &
  \multicolumn{1}{c|}{} &
  \multicolumn{1}{c|}{} &
  \multicolumn{1}{c|}{} &
   \\ \hline
Harm-P &
  \multicolumn{1}{c|}{} &
  \multicolumn{1}{c|}{\checkmark} &
  \multicolumn{1}{c|}{} &
   &
  \multicolumn{1}{c|}{} &
   &
  \multicolumn{1}{c|}{\checkmark} &
  \multicolumn{1}{c|}{} &
  \multicolumn{1}{c|}{\checkmark} &
  \multicolumn{1}{c|}{\checkmark} &
  \multicolumn{1}{c|}{} &
  \multicolumn{1}{c|}{} &
   &
  \multicolumn{1}{c|}{} &
  \multicolumn{1}{c|}{} &
  \multicolumn{1}{c|}{} &
   \\ \hline
Hate Speech in Pixels &
  \multicolumn{1}{c|}{} &
  \multicolumn{1}{c|}{\checkmark} &
  \multicolumn{1}{c|}{} &
   &
  \multicolumn{1}{c|}{} &
   &
  \multicolumn{1}{c|}{} &
  \multicolumn{1}{c|}{} &
  \multicolumn{1}{c|}{} &
  \multicolumn{1}{c|}{\checkmark} &
  \multicolumn{1}{c|}{} &
  \multicolumn{1}{c|}{} &
   &
  \multicolumn{1}{c|}{} &
  \multicolumn{1}{c|}{} &
  \multicolumn{1}{c|}{} &
   \\ \hline
HatReD &
  \multicolumn{1}{c|}{} &
  \multicolumn{1}{c|}{} &
  \multicolumn{1}{c|}{} &
   &
  \multicolumn{1}{c|}{} &
   &
  \multicolumn{1}{c|}{\checkmark} &
  \multicolumn{1}{c|}{} &
  \multicolumn{1}{c|}{} &
  \multicolumn{1}{c|}{} &
  \multicolumn{1}{c|}{} &
  \multicolumn{1}{c|}{} &
   &
  \multicolumn{1}{c|}{} &
  \multicolumn{1}{c|}{} &
  \multicolumn{1}{c|}{} &
   \\ \hline
HVVMemes &
  \multicolumn{1}{c|}{} &
  \multicolumn{1}{c|}{\checkmark} &
  \multicolumn{1}{c|}{} &
   &
  \multicolumn{1}{c|}{} &
   &
  \multicolumn{1}{c|}{\checkmark} &
  \multicolumn{1}{c|}{} &
  \multicolumn{1}{c|}{\checkmark} &
  \multicolumn{1}{c|}{\checkmark} &
  \multicolumn{1}{c|}{} &
  \multicolumn{1}{c|}{} &
   &
  \multicolumn{1}{c|}{} &
  \multicolumn{1}{c|}{} &
  \multicolumn{1}{c|}{} &
   \\ \hline
Indian Political Memes &
  \multicolumn{1}{c|}{} &
  \multicolumn{1}{c|}{\checkmark} &
  \multicolumn{1}{c|}{} &
   &
  \multicolumn{1}{c|}{} &
   &
  \multicolumn{1}{c|}{} &
  \multicolumn{1}{c|}{} &
  \multicolumn{1}{c|}{} &
  \multicolumn{1}{c|}{} &
  \multicolumn{1}{c|}{} &
  \multicolumn{1}{c|}{} &
   &
  \multicolumn{1}{c|}{} &
  \multicolumn{1}{c|}{} &
  \multicolumn{1}{c|}{} &
   \\ \hline
Innopolis Hateful Memes &
  \multicolumn{1}{c|}{} &
  \multicolumn{1}{c|}{} &
  \multicolumn{1}{c|}{\checkmark} &
   &
  \multicolumn{1}{c|}{} &
   &
  \multicolumn{1}{c|}{} &
  \multicolumn{1}{c|}{} &
  \multicolumn{1}{c|}{} &
  \multicolumn{1}{c|}{} &
  \multicolumn{1}{c|}{\checkmark} &
  \multicolumn{1}{c|}{} &
   &
  \multicolumn{1}{c|}{} &
  \multicolumn{1}{c|}{\checkmark} &
  \multicolumn{1}{c|}{} &
   \\ \hline
KAU-Memes &
  \multicolumn{1}{c|}{} &
  \multicolumn{1}{c|}{} &
  \multicolumn{1}{c|}{} &
   &
  \multicolumn{1}{c|}{} &
   &
  \multicolumn{1}{c|}{\checkmark} &
  \multicolumn{1}{c|}{} &
  \multicolumn{1}{c|}{\checkmark} &
  \multicolumn{1}{c|}{\checkmark} &
  \multicolumn{1}{c|}{\checkmark} &
  \multicolumn{1}{c|}{} &
   &
  \multicolumn{1}{c|}{} &
  \multicolumn{1}{c|}{} &
  \multicolumn{1}{c|}{} &
   \\ \hline
Meme-Merge &
  \multicolumn{1}{c|}{} &
  \multicolumn{1}{c|}{} &
  \multicolumn{1}{c|}{} &
  \checkmark &
  \multicolumn{1}{c|}{} &
   &
  \multicolumn{1}{c|}{} &
  \multicolumn{1}{c|}{} &
  \multicolumn{1}{c|}{} &
  \multicolumn{1}{c|}{} &
  \multicolumn{1}{c|}{\checkmark} &
  \multicolumn{1}{c|}{\checkmark} &
   &
  \multicolumn{1}{c|}{} &
  \multicolumn{1}{c|}{} &
  \multicolumn{1}{c|}{} &
   \\ \hline
Memotion 1 &
  \multicolumn{1}{c|}{} &
  \multicolumn{1}{c|}{\checkmark} &
  \multicolumn{1}{c|}{} &
   &
  \multicolumn{1}{c|}{} &
   &
  \multicolumn{1}{c|}{} &
  \multicolumn{1}{c|}{} &
  \multicolumn{1}{c|}{} &
  \multicolumn{1}{c|}{} &
  \multicolumn{1}{c|}{} &
  \multicolumn{1}{c|}{} &
   &
  \multicolumn{1}{c|}{} &
  \multicolumn{1}{c|}{} &
  \multicolumn{1}{c|}{} &
   \\ \hline
Memotion 2 &
  \multicolumn{1}{c|}{} &
  \multicolumn{1}{c|}{} &
  \multicolumn{1}{c|}{} &
   &
  \multicolumn{1}{c|}{} &
  \checkmark &
  \multicolumn{1}{c|}{\checkmark} &
  \multicolumn{1}{c|}{} &
  \multicolumn{1}{c|}{\checkmark} &
  \multicolumn{1}{c|}{\checkmark} &
  \multicolumn{1}{c|}{} &
  \multicolumn{1}{c|}{} &
   &
  \multicolumn{1}{c|}{} &
  \multicolumn{1}{c|}{} &
  \multicolumn{1}{c|}{} &
   \\ \hline
MET-Meme &
  \multicolumn{1}{c|}{} &
  \multicolumn{1}{c|}{\checkmark} &
  \multicolumn{1}{c|}{} &
  \checkmark &
  \multicolumn{1}{c|}{} &
   &
  \multicolumn{1}{c|}{} &
  \multicolumn{1}{c|}{} &
  \multicolumn{1}{c|}{} &
  \multicolumn{1}{c|}{} &
  \multicolumn{1}{c|}{\checkmark} &
  \multicolumn{1}{c|}{\checkmark} &
   &
  \multicolumn{1}{c|}{} &
  \multicolumn{1}{c|}{} &
  \multicolumn{1}{c|}{} &
   \\ \hline
Misogynistic-MEME &
  \multicolumn{1}{c|}{} &
  \multicolumn{1}{c|}{} &
  \multicolumn{1}{c|}{} &
   &
  \multicolumn{1}{c|}{} &
   &
  \multicolumn{1}{c|}{\checkmark} &
  \multicolumn{1}{c|}{} &
  \multicolumn{1}{c|}{\checkmark} &
  \multicolumn{1}{c|}{\checkmark} &
  \multicolumn{1}{c|}{\checkmark} &
  \multicolumn{1}{c|}{} &
   &
  \multicolumn{1}{c|}{} &
  \multicolumn{1}{c|}{} &
  \multicolumn{1}{c|}{} &
   \\ \hline
MultiBully &
  \multicolumn{1}{c|}{} &
  \multicolumn{1}{c|}{} &
  \multicolumn{1}{c|}{} &
   &
  \multicolumn{1}{c|}{} &
   &
  \multicolumn{1}{c|}{} &
  \multicolumn{1}{c|}{} &
  \multicolumn{1}{c|}{} &
  \multicolumn{1}{c|}{\checkmark} &
  \multicolumn{1}{c|}{\checkmark} &
  \multicolumn{1}{c|}{} &
   &
  \multicolumn{1}{c|}{} &
  \multicolumn{1}{c|}{} &
  \multicolumn{1}{c|}{} &
   \\ \hline
MultiBully-Ex &
  \multicolumn{1}{c|}{} &
  \multicolumn{1}{c|}{} &
  \multicolumn{1}{c|}{} &
   &
  \multicolumn{1}{c|}{} &
   &
  \multicolumn{1}{c|}{} &
  \multicolumn{1}{c|}{} &
  \multicolumn{1}{c|}{} &
  \multicolumn{1}{c|}{\checkmark} &
  \multicolumn{1}{c|}{\checkmark} &
  \multicolumn{1}{c|}{} &
   &
  \multicolumn{1}{c|}{} &
  \multicolumn{1}{c|}{} &
  \multicolumn{1}{c|}{} &
   \\ \hline
MAMI  &
  \multicolumn{1}{c|}{} &
  \multicolumn{1}{c|}{} &
  \multicolumn{1}{c|}{} &
   &
  \multicolumn{1}{c|}{} &
  \checkmark &
  \multicolumn{1}{c|}{} &
  \multicolumn{1}{c|}{} &
  \multicolumn{1}{c|}{} &
  \multicolumn{1}{c|}{\checkmark} &
  \multicolumn{1}{c|}{\checkmark} &
  \multicolumn{1}{c|}{} &
   &
  \multicolumn{1}{c|}{\checkmark} &
  \multicolumn{1}{c|}{} &
  \multicolumn{1}{c|}{} &
  \checkmark \\ \hline
MultiOFF &
  \multicolumn{1}{c|}{} &
  \multicolumn{1}{c|}{} &
  \multicolumn{1}{c|}{} &
   &
  \multicolumn{1}{c|}{} &
   &
  \multicolumn{1}{c|}{\checkmark} &
  \multicolumn{1}{c|}{} &
  \multicolumn{1}{c|}{\checkmark} &
  \multicolumn{1}{c|}{\checkmark} &
  \multicolumn{1}{c|}{\checkmark} &
  \multicolumn{1}{c|}{} &
   &
  \multicolumn{1}{c|}{} &
  \multicolumn{1}{c|}{} &
  \multicolumn{1}{c|}{} &
   \\ \hline
Pol\_Off\_Meme &
  \multicolumn{1}{c|}{} &
  \multicolumn{1}{c|}{\checkmark} &
  \multicolumn{1}{c|}{} &
   &
  \multicolumn{1}{c|}{} &
   &
  \multicolumn{1}{c|}{} &
  \multicolumn{1}{c|}{} &
  \multicolumn{1}{c|}{} &
  \multicolumn{1}{c|}{} &
  \multicolumn{1}{c|}{} &
  \multicolumn{1}{c|}{} &
   &
  \multicolumn{1}{c|}{} &
  \multicolumn{1}{c|}{} &
  \multicolumn{1}{c|}{} &
   \\ \hline
SemEval-2021 Task 6 &
  \multicolumn{1}{c|}{} &
  \multicolumn{1}{c|}{} &
  \multicolumn{1}{c|}{} &
   &
  \multicolumn{1}{c|}{} &
   &
  \multicolumn{1}{c|}{\checkmark} &
  \multicolumn{1}{c|}{} &
  \multicolumn{1}{c|}{} &
  \multicolumn{1}{c|}{} &
  \multicolumn{1}{c|}{} &
  \multicolumn{1}{c|}{} &
   &
  \multicolumn{1}{c|}{} &
  \multicolumn{1}{c|}{} &
  \multicolumn{1}{c|}{} &
   \\ \hline
TamilMemes &
  \multicolumn{1}{c|}{} &
  \multicolumn{1}{c|}{} &
  \multicolumn{1}{c|}{} &
   &
  \multicolumn{1}{c|}{\checkmark} &
   &
  \multicolumn{1}{c|}{\checkmark} &
  \multicolumn{1}{c|}{} &
  \multicolumn{1}{c|}{\checkmark} &
  \multicolumn{1}{c|}{} &
  \multicolumn{1}{c|}{} &
  \multicolumn{1}{c|}{} &
  \checkmark &
  \multicolumn{1}{c|}{} &
  \multicolumn{1}{c|}{} &
  \multicolumn{1}{c|}{} &
   \\ \hline
TrollsWithOpinion &
  \multicolumn{1}{c|}{} &
  \multicolumn{1}{c|}{\checkmark} &
  \multicolumn{1}{c|}{} &
   &
  \multicolumn{1}{c|}{} &
   &
  \multicolumn{1}{c|}{} &
  \multicolumn{1}{c|}{} &
  \multicolumn{1}{c|}{} &
  \multicolumn{1}{c|}{} &
  \multicolumn{1}{c|}{} &
  \multicolumn{1}{c|}{} &
   &
  \multicolumn{1}{c|}{} &
  \multicolumn{1}{l|}{} &
  \multicolumn{1}{l|}{} &
  \multicolumn{1}{l|}{} \\ \hline
\end{tabular}
\captionsetup{width=0.9\textwidth}
\caption{Sources of meme acquisition, as documented in the papers introducing the datasets. Rows are datasets, while columns are sources categorized into search engines, image hosting platforms, social media platforms, or meme creation and sharing platforms. Abbreviations used: B: Bing, G: Google, D: DuckDuckGo, B: Baidu, Pin: Pinterest, Img Platforms: Image Hosting Platforms, Img: Imgur, FB: Facebook, IG: Instagram, Meme Platforms: Meme Creation and Sharing Platforms, RD: Reddit, TW: Twitter, Wei: Weibo, WA: WhatsApp, KYM: KnowYourMeme, MD: Memedroid, MG: MemeGenerator.}
\label{sec:appendix:dataset_sources}
\end{table}

%% file: cas-sc-template.bbl
\begin{thebibliography}{252}
\expandafter\ifx\csname natexlab\endcsname\relax\def\natexlab#1{#1}\fi
\providecommand{\url}[1]{\texttt{#1}}
\providecommand{\href}[2]{#2}
\providecommand{\path}[1]{#1}
\providecommand{\DOIprefix}{doi:}
\providecommand{\ArXivprefix}{arXiv:}
\providecommand{\URLprefix}{URL: }
\providecommand{\Pubmedprefix}{pmid:}
\providecommand{\doi}[1]{\href{http://dx.doi.org/#1}{\path{#1}}}
\providecommand{\Pubmed}[1]{\href{pmid:#1}{\path{#1}}}
\providecommand{\bibinfo}[2]{#2}
\ifx\xfnm\relax \def\xfnm[#1]{\unskip,\space#1}\fi
\bibitem[{Dawkins(1989)}]{dawkins1989selfish}
\bibinfo{author}{R.~Dawkins}, \bibinfo{title}{The Selfish Gene}, \bibinfo{edition}{new} ed., \bibinfo{publisher}{Oxford University Press}, \bibinfo{year}{1989}.
\bibitem[{{D. Dennet}(2007)}]{ted2007dangerous}
\bibinfo{author}{{D. Dennet}}, \bibinfo{title}{Dangerous memes - a ted talk}, \bibinfo{howpublished}{Youtube}, \bibinfo{year}{2007}. \URLprefix \url{https://www.youtube.com/watch?v=KzGjEkp772s}.
\bibitem[{Koutlis et~al.(2023)Koutlis, Schinas, and Papadopoulos}]{koutlis2023memetector}
\bibinfo{author}{C.~Koutlis}, \bibinfo{author}{M.~Schinas}, \bibinfo{author}{S.~Papadopoulos},
\newblock \bibinfo{title}{Memetector: Enforcing deep focus for meme detection},
\newblock \bibinfo{journal}{International Journal of Multimedia Information Retrieval} \bibinfo{volume}{12} (\bibinfo{year}{2023}) \bibinfo{pages}{11}.
\bibitem[{Peeters et~al.(2021)Peeters, Tuters, Willaert, and De~Zeeuw}]{peeters2021vernacular}
\bibinfo{author}{S.~Peeters}, \bibinfo{author}{M.~Tuters}, \bibinfo{author}{T.~Willaert}, \bibinfo{author}{D.~De~Zeeuw},
\newblock \bibinfo{title}{On the vernacular language games of an antagonistic online subculture},
\newblock \bibinfo{journal}{Frontiers in big Data} \bibinfo{volume}{4} (\bibinfo{year}{2021}) \bibinfo{pages}{718368}.
\bibitem[{Arkenbout(2022)}]{arkenbout2022political}
\bibinfo{author}{C.~Arkenbout},
\newblock \bibinfo{title}{Political meme toolkit: leftist dutch meme makers share their trade secrets},
\newblock in: \bibinfo{booktitle}{Critical meme reader II: memetic tacticality}, \bibinfo{publisher}{Institute of Network Cultures}, \bibinfo{year}{2022}, pp. \bibinfo{pages}{20--31}.
\bibitem[{Wagener(2023)}]{wagener2023semiotic}
\bibinfo{author}{A.~Wagener},
\newblock \bibinfo{title}{Semiotic excess in memes: From postdigital creativity to social violence},
\newblock \bibinfo{journal}{Internet Pragmatics} \bibinfo{volume}{6} (\bibinfo{year}{2023}) \bibinfo{pages}{239--258}.
\bibitem[{Kiela et~al.(2020)Kiela, Firooz, Mohan, Goswami, Singh, Fitzpatrick, Bull, Lipstein, Nelli, Zhu, Muennighoff, Velioglu, Rose, Lippe, Holla, Chandra, Rajamanickam, Antoniou, Shutova, Yannakoudakis, Sandulescu, Ozertem, Pantel, Specia, and Parikh}]{Kiela2020hateful_report}
\bibinfo{author}{D.~Kiela}, \bibinfo{author}{H.~Firooz}, \bibinfo{author}{A.~Mohan}, \bibinfo{author}{V.~Goswami}, \bibinfo{author}{A.~Singh}, \bibinfo{author}{C.~Fitzpatrick}, \bibinfo{author}{P.~Bull}, \bibinfo{author}{G.~Lipstein}, \bibinfo{author}{T.~Nelli}, \bibinfo{author}{R.~Zhu}, \bibinfo{author}{N.~Muennighoff}, \bibinfo{author}{R.~Velioglu}, \bibinfo{author}{J.~Rose}, \bibinfo{author}{P.~Lippe}, \bibinfo{author}{N.~Holla}, \bibinfo{author}{S.~Chandra}, \bibinfo{author}{S.~Rajamanickam}, \bibinfo{author}{G.~Antoniou}, \bibinfo{author}{E.~Shutova}, \bibinfo{author}{H.~Yannakoudakis}, \bibinfo{author}{V.~Sandulescu}, \bibinfo{author}{U.~Ozertem}, \bibinfo{author}{P.~Pantel}, \bibinfo{author}{L.~Specia}, \bibinfo{author}{D.~Parikh},
\newblock \bibinfo{title}{The hateful memes challenge: Competition report},
\newblock in: \bibinfo{booktitle}{Proceedings of Machine Learning Res.}, volume \bibinfo{volume}{133}, \bibinfo{year}{2020}.
\bibitem[{Kirk et~al.(2021)Kirk, Jun, Rauba, Wachtel, Li, Bai, Broestl, Doff-Sotta, Shtedritski, and Asano}]{Kirk2021Memes}
\bibinfo{author}{H.~Kirk}, \bibinfo{author}{Y.~Jun}, \bibinfo{author}{P.~Rauba}, \bibinfo{author}{G.~Wachtel}, \bibinfo{author}{R.~Li}, \bibinfo{author}{X.~Bai}, \bibinfo{author}{N.~Broestl}, \bibinfo{author}{M.~Doff-Sotta}, \bibinfo{author}{A.~Shtedritski}, \bibinfo{author}{Y.~Asano},
\newblock \bibinfo{title}{Memes in the wild: Assessing the generalizability of the hateful memes challenge dataset},
\newblock in: \bibinfo{booktitle}{Proceedings of WOAH 2021 - 5th Workshop on Online Abuse and Harms}, volume~\bibinfo{volume}{26}, \bibinfo{year}{2021}.
\bibitem[{Kumar and Nandakumar(2022)}]{Kumar2022Hate-CLIPper}
\bibinfo{author}{G.~Kumar}, \bibinfo{author}{K.~Nandakumar},
\newblock \bibinfo{title}{Hate-clipper: Multimodal hateful meme classification based on cross-modal interaction of clip features},
\newblock in: \bibinfo{booktitle}{NLP4PI 2022 - 2nd Workshop on NLP for Positive Impact, Proceedings of the Workshop}, volume \bibinfo{volume}{171}, \bibinfo{year}{2022}.
\bibitem[{Sharma et~al.(2022)Sharma, Alam, Akhtar, Dimitrov, Da~San~Martino, Firooz, Halevy, Silvestri, Nakov, and Chakraborty}]{sharma2022detecting}
\bibinfo{author}{S.~Sharma}, \bibinfo{author}{F.~Alam}, \bibinfo{author}{M.~S. Akhtar}, \bibinfo{author}{D.~Dimitrov}, \bibinfo{author}{G.~Da~San~Martino}, \bibinfo{author}{H.~Firooz}, \bibinfo{author}{A.~Halevy}, \bibinfo{author}{F.~Silvestri}, \bibinfo{author}{P.~Nakov}, \bibinfo{author}{T.~Chakraborty},
\newblock \bibinfo{title}{Detecting and understanding harmful memes: A survey},
\newblock \bibinfo{journal}{arXiv preprint arXiv:2205.04274} \bibinfo{volume}{1} (\bibinfo{year}{2022}) \bibinfo{pages}{1--9}.
\bibitem[{Grasso et~al.(2024)Grasso, La~Gatta, Moscato, and Sperl\`{\i}}]{Grasso2024KERMIT}
\bibinfo{author}{B.~Grasso}, \bibinfo{author}{V.~La~Gatta}, \bibinfo{author}{V.~Moscato}, \bibinfo{author}{G.~Sperl\`{\i}},
\newblock \bibinfo{title}{Kermit: Knowledge-empowered model in harmful meme detection},
\newblock \bibinfo{journal}{Information Fusion} \bibinfo{volume}{106} (\bibinfo{year}{2024}).
\bibitem[{Yang et~al.(2023)Yang, Zhu, Han, and Hu}]{Yang2023Invariant}
\bibinfo{author}{C.~Yang}, \bibinfo{author}{F.~Zhu}, \bibinfo{author}{J.~Han}, \bibinfo{author}{S.~Hu},
\newblock \bibinfo{title}{Invariant meets specific: A scalable harmful memes detection framework},
\newblock \bibinfo{journal}{MM 2023 - Proceedings of the 31st ACM International Conference on Multimedia} \bibinfo{volume}{4788} (\bibinfo{year}{2023}).
\bibitem[{Das and Mukherjee(2023)}]{das2023banglaabusememe}
\bibinfo{author}{M.~Das}, \bibinfo{author}{A.~Mukherjee},
\newblock \bibinfo{title}{Banglaabusememe: A dataset for bengali abusive meme classification},
\newblock \bibinfo{journal}{arXiv preprint arXiv:2310.11748}  (\bibinfo{year}{2023}) \bibinfo{pages}{15498--15512}.
\bibitem[{Jain et~al.(2023)Jain, Maity, Jha, and Saha}]{jain2023generative}
\bibinfo{author}{R.~Jain}, \bibinfo{author}{K.~Maity}, \bibinfo{author}{P.~Jha}, \bibinfo{author}{S.~Saha},
\newblock \bibinfo{title}{Generative models vs discriminative models: Which performs better in detecting cyberbullying in memes?},
\newblock in: \bibinfo{booktitle}{Proceedings of the International Joint Conference on Neural Networks}, \bibinfo{year}{2023}. \DOIprefix\doi{10.1109/ijcnn54540.2023.10191363}.
\bibitem[{Shang et~al.(2021)Shang, Zhang, Zha, Chen, Youn, and Wang}]{shang2021aomd}
\bibinfo{author}{L.~Shang}, \bibinfo{author}{Y.~Zhang}, \bibinfo{author}{Y.~Zha}, \bibinfo{author}{Y.~Chen}, \bibinfo{author}{C.~Youn}, \bibinfo{author}{D.~Wang},
\newblock \bibinfo{title}{Aomd: An analogy-aware approach to offensive meme detection on social media},
\newblock \bibinfo{journal}{Information Processing \& Management}  (\bibinfo{year}{2021}).
\bibitem[{Sharma et~al.(2020)Sharma, Bhageria, Scott, Pykl, Das, Chakraborty, Pulabaigari, and Gamback}]{sharma2020semeval}
\bibinfo{author}{C.~Sharma}, \bibinfo{author}{D.~Bhageria}, \bibinfo{author}{W.~Scott}, \bibinfo{author}{S.~Pykl}, \bibinfo{author}{A.~Das}, \bibinfo{author}{T.~Chakraborty}, \bibinfo{author}{V.~Pulabaigari}, \bibinfo{author}{B.~Gamback},
\newblock \bibinfo{title}{Semeval-2020 task 8: Memotion analysis--the visuo-lingual metaphor!},
\newblock \bibinfo{journal}{arXiv preprint arXiv:2008.03781}  (\bibinfo{year}{2020}) \bibinfo{pages}{759--773}.
\bibitem[{Arora et~al.(2023)Arora, Nakov, Hardalov, Sarwar, Nayak, Dinkov, Zlatkova, Dent, Bhatawdekar, Bouchard, and Augenstein}]{arora2023detecting}
\bibinfo{author}{A.~Arora}, \bibinfo{author}{P.~Nakov}, \bibinfo{author}{M.~Hardalov}, \bibinfo{author}{S.~M. Sarwar}, \bibinfo{author}{V.~Nayak}, \bibinfo{author}{Y.~Dinkov}, \bibinfo{author}{D.~Zlatkova}, \bibinfo{author}{K.~Dent}, \bibinfo{author}{A.~Bhatawdekar}, \bibinfo{author}{G.~Bouchard}, \bibinfo{author}{I.~Augenstein}, \bibinfo{title}{Detecting harmful content on online platforms: What platforms need vs. where research efforts go}, \bibinfo{year}{2023}. \href{http://arxiv.org/abs/2103.00153}{\tt arXiv:2103.00153}.
\bibitem[{Williams and Dupuis(2020)}]{williams2020don}
\bibinfo{author}{A.~Williams}, \bibinfo{author}{M.~Dupuis},
\newblock \bibinfo{title}{I don't always spread disinformation on the web, but when i do i like to use memes: An examination of memes in the spread of disinformation},
\newblock in: \bibinfo{booktitle}{Proceedings of the 11th International Multi-Conferences on Complexity, Informatics and Cybernetics: IMCIC}, \bibinfo{year}{2020}, pp. \bibinfo{pages}{165--172}.
\bibitem[{Dimitrov et~al.(2021)Dimitrov, Ali, Shaar, Alam, Silvestri, Firooz, and Martino}]{Dimitrov2021detecting}
\bibinfo{author}{D.~Dimitrov}, \bibinfo{author}{B.~B. Ali}, \bibinfo{author}{S.~Shaar}, \bibinfo{author}{F.~Alam}, \bibinfo{author}{F.~Silvestri}, \bibinfo{author}{H.~Firooz}, \bibinfo{author}{G.~D.~S. Martino},
\newblock \bibinfo{title}{Detecting propaganda techniques in memes},
\newblock \bibinfo{journal}{arXiv preprint arXiv:2109.08013}  (\bibinfo{year}{2021}) \bibinfo{pages}{6603--6617}.
\bibitem[{Abdullah et~al.(2023)Abdullah, Abujaber, Al-Qarqaz, Abbott, and Hadzikadic}]{abdullah2023combating}
\bibinfo{author}{M.~Abdullah}, \bibinfo{author}{D.~Abujaber}, \bibinfo{author}{A.~Al-Qarqaz}, \bibinfo{author}{R.~Abbott}, \bibinfo{author}{M.~Hadzikadic},
\newblock \bibinfo{title}{Combating propaganda texts using transfer learning},
\newblock \bibinfo{journal}{IAES International Journal of Artificial Intelligence} \bibinfo{volume}{12} (\bibinfo{year}{2023}) \bibinfo{pages}{956--965}.
\bibitem[{Rodr\'{\i}guez et~al.(2023)Rodr\'{\i}guez, Nakov, Dankers, and Shutova}]{rodriguez2023paper}
\bibinfo{author}{D.~Rodr\'{\i}guez}, \bibinfo{author}{P.~Nakov}, \bibinfo{author}{V.~Dankers}, \bibinfo{author}{E.~Shutova},
\newblock \bibinfo{title}{Paper bullets: Modeling propaganda with the help of metaphor},
\newblock \bibinfo{journal}{European Chapter of the Association for Computational Linguistics, Findings of EACL 2023}  (\bibinfo{year}{2023}) \bibinfo{pages}{472--489}.
\bibitem[{Shridara et~al.(2023)Shridara, Hl{\'a}dek, Pleva, and Halu{\v{s}}ka}]{shridara2023identification}
\bibinfo{author}{M.~G. Shridara}, \bibinfo{author}{D.~Hl{\'a}dek}, \bibinfo{author}{M.~Pleva}, \bibinfo{author}{R.~Halu{\v{s}}ka},
\newblock \bibinfo{title}{Identification of trolling in memes using convolutional neural networks},
\newblock in: \bibinfo{booktitle}{2023 33rd International Conference Radioelektronika (RADIOELEKTRONIKA)}, \bibinfo{organization}{IEEE}, \bibinfo{year}{2023}, pp. \bibinfo{pages}{1--6}.
\bibitem[{Suryawanshi et~al.(2023)Suryawanshi, Chakravarthi, Arcan, and Buitelaar}]{suryawanshi2023trollswithopinion}
\bibinfo{author}{S.~Suryawanshi}, \bibinfo{author}{B.~R. Chakravarthi}, \bibinfo{author}{M.~Arcan}, \bibinfo{author}{P.~Buitelaar},
\newblock \bibinfo{title}{Trollswithopinion: A taxonomy and dataset for predicting domain-specific opinion manipulation in troll memes},
\newblock \bibinfo{journal}{Multimedia Tools and Applications} \bibinfo{volume}{82} (\bibinfo{year}{2023}) \bibinfo{pages}{9137--9171}.
\bibitem[{Bebi\'{c} and Volarevic(2018)}]{bebic2018do}
\bibinfo{author}{D.~Bebi\'{c}}, \bibinfo{author}{M.~Volarevic},
\newblock \bibinfo{title}{Do not mess with a meme: the use of viral content in communicating politics},
\newblock \bibinfo{journal}{Communication \& Society} \bibinfo{volume}{31} (\bibinfo{year}{2018}) \bibinfo{pages}{43--56}.
\bibitem[{Mazambani et~al.(2015)Mazambani, Carlson, Reysen, and Hempelmann}]{mazambani2015impact}
\bibinfo{author}{G.~Mazambani}, \bibinfo{author}{M.~A. Carlson}, \bibinfo{author}{S.~Reysen}, \bibinfo{author}{C.~F. Hempelmann},
\newblock \bibinfo{title}{Impact of status and meme content on the spread of memes in virtual communities},
\newblock \bibinfo{journal}{Human Technology: An Interdisciplinary Journal on Humans in ICT Environments} \bibinfo{volume}{11} (\bibinfo{year}{2015}) \bibinfo{pages}{148--164}.
\bibitem[{Netanya(2019)}]{DangersMemes}
\bibinfo{author}{Netanya}, \bibinfo{title}{{T}he {D}angers of {M}emes --- netanyataitague}, \bibinfo{howpublished}{\url{https://medium.com/@netanyataitague/the-dangers-of-memes-b1bb67e10083}}, \bibinfo{year}{2019}. \bibinfo{note}{[Accessed 17-04-2024]}.
\bibitem[{Vang(2021)}]{thecurrentmsuMemeCulture}
\bibinfo{author}{M.~Vang}, \bibinfo{title}{{I}s {M}eme {C}ulture {P}roblematic - {T}he {C}urrent --- thecurrentmsu.com}, \bibinfo{howpublished}{\url{https://thecurrentmsu.com/2021/08/07/is-meme-culture-problematic/}}, \bibinfo{year}{2021}. \bibinfo{note}{[Accessed 17-04-2024]}.
\bibitem[{Rojas(2022)}]{studybreaksToxicityOnline}
\bibinfo{author}{K.~Rojas}, \bibinfo{title}{{T}he {T}oxicity of {O}nline {M}eme {C}ulture: {W}hen {I}s {I}t {T}oo {F}ar? --- studybreaks.com}, \bibinfo{howpublished}{\url{https://studybreaks.com/thoughts/meme-culture-2/}}, \bibinfo{year}{2022}. \bibinfo{note}{[Accessed 17-04-2024]}.
\bibitem[{Roberts(2023)}]{progressiveFreeHelicopter}
\bibinfo{author}{Z.~D. Roberts}, \bibinfo{title}{{H}ow the ‘{F}ree {H}elicopter {R}ides’ {M}eme {W}ent {V}iral --- progressive.org}, \bibinfo{howpublished}{\url{https://progressive.org/magazine/how-the-free-helicopter-rides-meme-went-viral-roberts-20230907/}}, \bibinfo{year}{2023}. \bibinfo{note}{[Accessed 17-04-2024]}.
\bibitem[{Serna(2024)}]{serna2024memes}
\bibinfo{author}{F.~J.~A. Serna},
\newblock \bibinfo{title}{Los memes como simbolos del discurso de odio: La influencia del humor gr{\'a}fico en la libertad de expresi{\'o}n y la pol{\'\i}tica},
\newblock \bibinfo{journal}{VISUAL REVIEW. International Visual Culture Review/Revista Internacional de Cultura Visual} \bibinfo{volume}{16} (\bibinfo{year}{2024}) \bibinfo{pages}{241--253}.
\bibitem[{Needham(2019)}]{unimelbToxicWent}
\bibinfo{author}{L.~Needham}, \bibinfo{title}{{H}ow the toxic went mainstream --- pursuit.unimelb.edu.au}, \bibinfo{howpublished}{\url{https://pursuit.unimelb.edu.au/articles/how-the-toxic-went-mainstream}}, \bibinfo{year}{2019}. \bibinfo{note}{[Accessed 17-04-2024]}.
\bibitem[{Duchscherer and Dovidio(2016)}]{duchscherer2016when}
\bibinfo{author}{K.~M. Duchscherer}, \bibinfo{author}{J.~F. Dovidio},
\newblock \bibinfo{title}{When memes are mean: Appraisals of and objections to stereotypic memes},
\newblock \bibinfo{journal}{Translational Issues in Psychological Science} \bibinfo{volume}{2} (\bibinfo{year}{2016}) \bibinfo{pages}{335--345}.
\bibitem[{Bennet(2020)}]{bennet2020moderates}
\bibinfo{author}{P.~M. Bennet}, \bibinfo{title}{Who Moderates the Social Media Giants? A Call to End Outsourcing}, \bibinfo{type}{Technical Report}, NYU STERN, Center for Business and Human Rights, \bibinfo{year}{2020}.
\bibitem[{Nondo(2023)}]{nondo2023facing}
\bibinfo{author}{N.~Nondo},
\newblock \bibinfo{title}{Facing disturbing content daily, online moderators in africa want better protections and a fair wage},
\newblock \bibinfo{journal}{CBC Radio}  (\bibinfo{year}{2023}).
\bibitem[{Perrigo(2022)}]{perrigo2022inside}
\bibinfo{author}{B.~Perrigo},
\newblock \bibinfo{title}{Inside facebook's african sweatshop},
\newblock \bibinfo{journal}{Time}  (\bibinfo{year}{2022}).
\bibitem[{Rowe(2023)}]{rowe2023destroyed}
\bibinfo{author}{N.~Rowe},
\newblock \bibinfo{title}{"it's destroyed me completely": Kenyan moderators decry toll of training of ai models},
\newblock \bibinfo{journal}{The Guardian}  (\bibinfo{year}{2023}).
\bibitem[{Mbagathi(2023)}]{mbagathi2023africa}
\bibinfo{author}{N.~Mbagathi},
\newblock \bibinfo{title}{In africa, taking on viral hate},
\newblock \bibinfo{journal}{Open Society Foundations}  (\bibinfo{year}{2023}).
\bibitem[{Afridi et~al.(2021)Afridi, Alam, Khan, Khan, and Lee}]{afridi2021multimodal}
\bibinfo{author}{T.~H. Afridi}, \bibinfo{author}{A.~Alam}, \bibinfo{author}{M.~N. Khan}, \bibinfo{author}{J.~Khan}, \bibinfo{author}{Y.-K. Lee},
\newblock \bibinfo{title}{A multimodal memes classification: A survey and open research issues},
\newblock in: \bibinfo{booktitle}{Innovations in Smart Cities Applications Volume 4: The Proceedings of the 5th International Conference on Smart City Applications}, \bibinfo{organization}{Springer}, \bibinfo{year}{2021}, pp. \bibinfo{pages}{1451--1466}.
\bibitem[{Hermida and Santos(2023)}]{hermida2023detecting}
\bibinfo{author}{P.~C. d.~Q. Hermida}, \bibinfo{author}{E.~M.~D. Santos},
\newblock \bibinfo{title}{Detecting hate speech in memes: a review},
\newblock \bibinfo{journal}{Artificial Intelligence Review} \bibinfo{volume}{56} (\bibinfo{year}{2023}) \bibinfo{pages}{1--19}.
\bibitem[{Zannettou et~al.(2018)Zannettou, Caulfield, Blackburn, De~Cristofaro, Sirivianos, Stringhini, and Suarez-Tangil}]{zannettou2018origins}
\bibinfo{author}{S.~Zannettou}, \bibinfo{author}{T.~Caulfield}, \bibinfo{author}{J.~Blackburn}, \bibinfo{author}{E.~De~Cristofaro}, \bibinfo{author}{M.~Sirivianos}, \bibinfo{author}{G.~Stringhini}, \bibinfo{author}{G.~Suarez-Tangil},
\newblock \bibinfo{title}{On the origins of memes by means of fringe web communities},
\newblock in: \bibinfo{booktitle}{Proceedings of the internet measurement conference 2018}, \bibinfo{year}{2018}, pp. \bibinfo{pages}{188--202}.
\bibitem[{Chakraborty and Masud(2022)}]{chakraborty2022nipping}
\bibinfo{author}{T.~Chakraborty}, \bibinfo{author}{S.~Masud},
\newblock \bibinfo{title}{Nipping in the bud: detection, diffusion and mitigation of hate speech on social media},
\newblock \bibinfo{journal}{ACM SIGWEB Newsletter} \bibinfo{volume}{2022} (\bibinfo{year}{2022}) \bibinfo{pages}{1--9}.
\bibitem[{Polli and Sindoni(2023)}]{polli2023multimodal}
\bibinfo{author}{C.~Polli}, \bibinfo{author}{M.~G. Sindoni},
\newblock \bibinfo{title}{Multimodal computation or interpretation? automatic vs. critical understanding of text-image relations in racist memes},
\newblock \bibinfo{journal}{Ssrn}  (\bibinfo{year}{2023}).
\bibitem[{Solon(2013)}]{solon2013richard}
\bibinfo{author}{O.~Solon},
\newblock \bibinfo{title}{Richard dawkins on the internet's hijacking of the word `meme'},
\newblock \bibinfo{journal}{Wired UK} \bibinfo{volume}{20} (\bibinfo{year}{2013}).
\bibitem[{Shifman(2013)}]{shifman2013memes}
\bibinfo{author}{L.~Shifman}, \bibinfo{title}{Memes in Digital Culture}, MIT Press Essential Knowledge Series, \bibinfo{publisher}{MIT Press}, \bibinfo{year}{2013}.
\bibitem[{Dynel(2022)}]{dynel2022life}
\bibinfo{author}{M.~Dynel},
\newblock \bibinfo{title}{The life of covid-19 mask memes: a diachronic study of the pandemic memescape},
\newblock \bibinfo{journal}{Comunicar} \bibinfo{volume}{30} (\bibinfo{year}{2022}) \bibinfo{pages}{73--85}.
\bibitem[{Thakur et~al.(2023)Thakur, Ilievski, Sandlin, Sourati, Luceri, Tommasini, and Mermoud}]{thakur2023explainable}
\bibinfo{author}{A.~Thakur}, \bibinfo{author}{F.~Ilievski}, \bibinfo{author}{H.-A. Sandlin}, \bibinfo{author}{Z.~Sourati}, \bibinfo{author}{L.~Luceri}, \bibinfo{author}{R.~Tommasini}, \bibinfo{author}{A.~Mermoud},
\newblock \bibinfo{title}{Explainable classification of internet memes},
\newblock in: \bibinfo{booktitle}{CEUR Workshop Proceedings}, volume \bibinfo{volume}{3432}, \bibinfo{year}{2023}, pp. \bibinfo{pages}{395--409}.
\bibitem[{Xie et~al.(2011)Xie, Natsev, Kender, Hill, and Smith}]{xie2011visual}
\bibinfo{author}{L.~Xie}, \bibinfo{author}{A.~Natsev}, \bibinfo{author}{J.~R. Kender}, \bibinfo{author}{M.~Hill}, \bibinfo{author}{J.~R. Smith},
\newblock \bibinfo{title}{Visual memes in social media: tracking real-world news in youtube videos},
\newblock in: \bibinfo{booktitle}{Proceedings of the 19th ACM international conference on Multimedia}, \bibinfo{year}{2011}, pp. \bibinfo{pages}{53--62}.
\bibitem[{Du et~al.(2020)Du, Masood, and Joseph}]{du2020understanding}
\bibinfo{author}{Y.~Du}, \bibinfo{author}{M.~A. Masood}, \bibinfo{author}{K.~Joseph},
\newblock \bibinfo{title}{Understanding visual memes: An empirical analysis of text superimposed on memes shared on twitter},
\newblock in: \bibinfo{booktitle}{Proceedings of the International AAAI Conference on Web and Social Media}, volume~\bibinfo{volume}{14}, \bibinfo{organization}{Aaai}, \bibinfo{year}{2020}, pp. \bibinfo{pages}{153--164}.
\bibitem[{Pramanick et~al.(2021)Pramanick, Dimitrov, Mukherjee, Sharma, Akhtar, Nakov, and Chakraborty}]{pramanick2021detecting}
\bibinfo{author}{S.~Pramanick}, \bibinfo{author}{D.~Dimitrov}, \bibinfo{author}{R.~Mukherjee}, \bibinfo{author}{S.~Sharma}, \bibinfo{author}{M.~S. Akhtar}, \bibinfo{author}{P.~Nakov}, \bibinfo{author}{T.~Chakraborty},
\newblock \bibinfo{title}{Detecting harmful memes and their targets},
\newblock \bibinfo{journal}{arXiv preprint arXiv:2110.00413}  (\bibinfo{year}{2021}).
\bibitem[{Knobel and Lankshear(2007)}]{knobel2007online}
\bibinfo{author}{M.~Knobel}, \bibinfo{author}{C.~Lankshear},
\newblock \bibinfo{title}{Online memes, affinities, and cultural production},
\newblock \bibinfo{journal}{A new literacies sampler} \bibinfo{volume}{29} (\bibinfo{year}{2007}) \bibinfo{pages}{199--227}.
\bibitem[{Sharma et~al.(2020)Sharma, Pulabaigari, and Das}]{sharma2020meme}
\bibinfo{author}{C.~Sharma}, \bibinfo{author}{V.~Pulabaigari}, \bibinfo{author}{A.~Das},
\newblock \bibinfo{title}{Meme vs. non-meme classification using visuo-linguistic association.},
\newblock in: \bibinfo{booktitle}{WEBIST}, \bibinfo{year}{2020}, pp. \bibinfo{pages}{353--360}.
\bibitem[{Sherratt et~al.(2023)Sherratt, Pimbblet, and Dethlefs}]{sherratt2023multi}
\bibinfo{author}{V.~Sherratt}, \bibinfo{author}{K.~Pimbblet}, \bibinfo{author}{N.~Dethlefs},
\newblock \bibinfo{title}{Multi-channel convolutional neural network for precise meme classification},
\newblock in: \bibinfo{booktitle}{Proceedings of the 2023 ACM International Conference on Multimedia Retrieval}, \bibinfo{year}{2023}, pp. \bibinfo{pages}{190--198}.
\bibitem[{Kiela et~al.(2020)Kiela, Firooz, Mohan, Goswami, Singh, Ringshia, and Testuggine}]{Kiela2020}
\bibinfo{author}{D.~Kiela}, \bibinfo{author}{H.~Firooz}, \bibinfo{author}{A.~Mohan}, \bibinfo{author}{V.~Goswami}, \bibinfo{author}{A.~Singh}, \bibinfo{author}{P.~Ringshia}, \bibinfo{author}{D.~Testuggine},
\newblock \bibinfo{title}{The hateful memes challenge: Detecting hate speech in multimodal memes},
\newblock in: \bibinfo{booktitle}{Advances in neural information processing systems}, volume~\bibinfo{volume}{33}, \bibinfo{year}{2020}, pp. \bibinfo{pages}{2611--2624}.
\bibitem[{Kostadinovska-Stojchevska and Shalevska(2018)}]{kostadinovska2018internet}
\bibinfo{author}{B.~Kostadinovska-Stojchevska}, \bibinfo{author}{E.~Shalevska},
\newblock \bibinfo{title}{Internet memes and their socio-linguistic features},
\newblock \bibinfo{journal}{European journal of literature, language and linguistics studies} \bibinfo{volume}{2} (\bibinfo{year}{2018}).
\bibitem[{Lapidot-Lefler and Barak(2012)}]{lapidot2012effects}
\bibinfo{author}{N.~Lapidot-Lefler}, \bibinfo{author}{A.~Barak},
\newblock \bibinfo{title}{Effects of anonymity, invisibility, and lack of eye-contact on toxic online disinhibition},
\newblock \bibinfo{journal}{Computers in human behavior} \bibinfo{volume}{28} (\bibinfo{year}{2012}) \bibinfo{pages}{434--443}.
\bibitem[{Gordeev and Potapov(2020)}]{gordeev2020toxicity}
\bibinfo{author}{D.~Gordeev}, \bibinfo{author}{V.~Potapov},
\newblock \bibinfo{title}{Toxicity in texts and images on the internet},
\newblock in: \bibinfo{booktitle}{International Conference on Speech and Computer}, \bibinfo{organization}{Springer}, \bibinfo{year}{2020}, pp. \bibinfo{pages}{156--165}.
\bibitem[{Sheth et~al.(2022)Sheth, Shalin, and Kursuncu}]{sheth2022defining}
\bibinfo{author}{A.~Sheth}, \bibinfo{author}{V.~L. Shalin}, \bibinfo{author}{U.~Kursuncu},
\newblock \bibinfo{title}{Defining and detecting toxicity on social media: context and knowledge are key},
\newblock \bibinfo{journal}{Neurocomputing} \bibinfo{volume}{490} (\bibinfo{year}{2022}) \bibinfo{pages}{312--318}.
\bibitem[{Carlisle(2022)}]{powerof0ToxicityMemes}
\bibinfo{author}{N.~Carlisle}, \bibinfo{title}{{T}oxicity, memes and raids - {P}ower of {Z}ero --- powerof0.org}, \bibinfo{howpublished}{\url{https://powerof0.org/toxicity-memes-and-raids/}}, \bibinfo{year}{2022}. \bibinfo{note}{[Accessed 17-04-2024]}.
\bibitem[{Mosleh et~al.(2024)Mosleh, Cole, and Rand}]{mosleh2024misinformation}
\bibinfo{author}{M.~Mosleh}, \bibinfo{author}{R.~Cole}, \bibinfo{author}{D.~G. Rand},
\newblock \bibinfo{title}{Misinformation and harmful language are interconnected, rather than distinct, challenges},
\newblock \bibinfo{journal}{PNAS nexus}  (\bibinfo{year}{2024}) \bibinfo{pages}{pgae111}.
\bibitem[{Ghosh et~al.(2021)Ghosh, Lepcha, Sakshi, Shah, and Umesh}]{ghosh2021detoxy}
\bibinfo{author}{S.~Ghosh}, \bibinfo{author}{S.~Lepcha}, \bibinfo{author}{S.~Sakshi}, \bibinfo{author}{R.~R. Shah}, \bibinfo{author}{S.~Umesh},
\newblock \bibinfo{title}{Detoxy: A large-scale multimodal dataset for toxicity classification in spoken utterances},
\newblock \bibinfo{journal}{arXiv preprint arXiv:2110.07592}  (\bibinfo{year}{2021}).
\bibitem[{Waseem et~al.(2017)Waseem, Davidson, Warmsley, and Weber}]{waseem2017understanding}
\bibinfo{author}{Z.~Waseem}, \bibinfo{author}{T.~Davidson}, \bibinfo{author}{D.~Warmsley}, \bibinfo{author}{I.~Weber},
\newblock \bibinfo{title}{Understanding abuse: A typology of abusive language detection subtasks},
\newblock \bibinfo{journal}{arXiv preprint arXiv:1705.09899}  (\bibinfo{year}{2017}).
\bibitem[{Wiegand et~al.(2019)Wiegand, Siegel, and Ruppenhofer}]{wiegand2018overview}
\bibinfo{author}{M.~Wiegand}, \bibinfo{author}{M.~Siegel}, \bibinfo{author}{J.~Ruppenhofer},
\newblock \bibinfo{title}{Overview of the germeval 2018 shared task on the identification of offensive language},
\newblock in: \bibinfo{booktitle}{Proceedings of GermEval 2018, 14th Conference on Natural Language Processing (KONVENS 2018)}, \bibinfo{publisher}{Austrian Academy of Sciences}, \bibinfo{year}{2019}, pp. \bibinfo{pages}{1 -- 10}. \URLprefix \url{https://nbn-resolving.org/urn:nbn:de:bsz:mh39-84935}.
\bibitem[{Zampieri et~al.(2019)Zampieri, Malmasi, Nakov, Rosenthal, Farra, and Kumar}]{zampieri2019semeval}
\bibinfo{author}{M.~Zampieri}, \bibinfo{author}{S.~Malmasi}, \bibinfo{author}{P.~Nakov}, \bibinfo{author}{S.~Rosenthal}, \bibinfo{author}{N.~Farra}, \bibinfo{author}{R.~Kumar},
\newblock \bibinfo{title}{Semeval-2019 task 6: Identifying and categorizing offensive language in social media (offenseval)},
\newblock \bibinfo{journal}{arXiv preprint arXiv:1903.08983}  (\bibinfo{year}{2019}).
\bibitem[{Piot et~al.(2024)Piot, Mart{\'\i}n-Rodilla, and Parapar}]{piot2024metahate}
\bibinfo{author}{P.~Piot}, \bibinfo{author}{P.~Mart{\'\i}n-Rodilla}, \bibinfo{author}{J.~Parapar},
\newblock \bibinfo{title}{Metahate: A dataset for unifying efforts on hate speech detection},
\newblock \bibinfo{journal}{arXiv preprint arXiv:2401.06526}  (\bibinfo{year}{2024}).
\bibitem[{Adams et~al.(2017)Adams, Sorensen, Elliott, Dixon, McDonald, n., and Cukierski}]{jigsaw}
\bibinfo{author}{C.~Adams}, \bibinfo{author}{J.~Sorensen}, \bibinfo{author}{J.~Elliott}, \bibinfo{author}{L.~Dixon}, \bibinfo{author}{M.~McDonald}, \bibinfo{author}{n.}, \bibinfo{author}{W.~Cukierski}, \bibinfo{title}{Toxic comment classification challenge}, \bibinfo{year}{2017}. \URLprefix \url{{https://kaggle.com/competitions/jigsaw-toxic-comment-classification-challenge}}.
\bibitem[{Davidson et~al.(2017)Davidson, Warmsley, Macy, and Weber}]{davidson2017automated}
\bibinfo{author}{T.~Davidson}, \bibinfo{author}{D.~Warmsley}, \bibinfo{author}{M.~Macy}, \bibinfo{author}{I.~Weber},
\newblock \bibinfo{title}{Automated hate speech detection and the problem of offensive language},
\newblock in: \bibinfo{booktitle}{Proceedings of the international AAAI conference on web and social media}, volume~\bibinfo{volume}{11}, \bibinfo{year}{2017}, pp. \bibinfo{pages}{512--515}.
\bibitem[{Ayo et~al.(2020)Ayo, Folorunso, Ibharalu, and Osinuga}]{ayo2020machine}
\bibinfo{author}{F.~E. Ayo}, \bibinfo{author}{O.~Folorunso}, \bibinfo{author}{F.~T. Ibharalu}, \bibinfo{author}{I.~A. Osinuga},
\newblock \bibinfo{title}{Machine learning techniques for hate speech classification of twitter data: State-of-the-art, future challenges and research directions},
\newblock \bibinfo{journal}{Computer Science Review} \bibinfo{volume}{38} (\bibinfo{year}{2020}) \bibinfo{pages}{100311}.
\bibitem[{Alkomah and Ma(2022)}]{alkomah2022literature}
\bibinfo{author}{F.~Alkomah}, \bibinfo{author}{X.~Ma},
\newblock \bibinfo{title}{A literature review of textual hate speech detection methods and datasets},
\newblock \bibinfo{journal}{Information} \bibinfo{volume}{13} (\bibinfo{year}{2022}) \bibinfo{pages}{273}.
\bibitem[{Chhabra and Vishwakarma(2023)}]{chhabra2023literature}
\bibinfo{author}{A.~Chhabra}, \bibinfo{author}{D.~K. Vishwakarma},
\newblock \bibinfo{title}{A literature survey on multimodal and multilingual automatic hate speech identification},
\newblock \bibinfo{journal}{Multimedia Systems} \bibinfo{volume}{29} (\bibinfo{year}{2023}) \bibinfo{pages}{1203--1230}.
\bibitem[{Fortuna et~al.(2020)Fortuna, Soler, and Wanner}]{fortuna2020toxic}
\bibinfo{author}{P.~Fortuna}, \bibinfo{author}{J.~Soler}, \bibinfo{author}{L.~Wanner},
\newblock \bibinfo{title}{Toxic, hateful, offensive or abusive? what are we really classifying? an empirical analysis of hate speech datasets},
\newblock in: \bibinfo{booktitle}{Proceedings of the Twelfth Language Resources and Evaluation Conference}, \bibinfo{year}{2020}, pp. \bibinfo{pages}{6786--6794}.
\bibitem[{Yankoski et~al.(2021)Yankoski, Scheirer, and Weninger}]{yankoski2021meme}
\bibinfo{author}{M.~Yankoski}, \bibinfo{author}{W.~Scheirer}, \bibinfo{author}{T.~Weninger},
\newblock \bibinfo{title}{Meme warfare: Ai countermeasures to disinformation should focus on popular, not perfect, fakes},
\newblock \bibinfo{journal}{Bulletin of the atomic scientists} \bibinfo{volume}{77} (\bibinfo{year}{2021}) \bibinfo{pages}{119--123}.
\bibitem[{Banko et~al.(2020)Banko, MacKeen, and Ray}]{banko2020unified}
\bibinfo{author}{M.~Banko}, \bibinfo{author}{B.~MacKeen}, \bibinfo{author}{L.~Ray},
\newblock \bibinfo{title}{A unified taxonomy of harmful content},
\newblock in: \bibinfo{booktitle}{Proceedings of the fourth workshop on online abuse and harms}, \bibinfo{year}{2020}, pp. \bibinfo{pages}{125--137}.
\bibitem[{Nakov et~al.(2021)Nakov, Nayak, Dent, Bhatawdekar, Sarwar, Hardalov, Dinkov, Zlatkova, Bouchard, and Augenstein}]{nakov2021detecting}
\bibinfo{author}{P.~Nakov}, \bibinfo{author}{V.~Nayak}, \bibinfo{author}{K.~Dent}, \bibinfo{author}{A.~Bhatawdekar}, \bibinfo{author}{S.~M. Sarwar}, \bibinfo{author}{M.~Hardalov}, \bibinfo{author}{Y.~Dinkov}, \bibinfo{author}{D.~Zlatkova}, \bibinfo{author}{G.~Bouchard}, \bibinfo{author}{I.~Augenstein},
\newblock \bibinfo{title}{Detecting abusive language on online platforms: A critical analysis},
\newblock \bibinfo{journal}{arXiv preprint arXiv:2103.00153}  (\bibinfo{year}{2021}).
\bibitem[{Halevy et~al.(2022)Halevy, Canton-Ferrer, Ma, Ozertem, Pantel, Saeidi, Silvestri, and Stoyanov}]{halevy2022preserving}
\bibinfo{author}{A.~Halevy}, \bibinfo{author}{C.~Canton-Ferrer}, \bibinfo{author}{H.~Ma}, \bibinfo{author}{U.~Ozertem}, \bibinfo{author}{P.~Pantel}, \bibinfo{author}{M.~Saeidi}, \bibinfo{author}{F.~Silvestri}, \bibinfo{author}{V.~Stoyanov},
\newblock \bibinfo{title}{Preserving integrity in online social networks},
\newblock \bibinfo{journal}{Communications of the ACM} \bibinfo{volume}{65} (\bibinfo{year}{2022}) \bibinfo{pages}{92--98}.
\bibitem[{Alam et~al.(2021)Alam, Cresci, Chakraborty, Silvestri, Dimitrov, Martino, Shaar, Firooz, and Nakov}]{alam2021survey}
\bibinfo{author}{F.~Alam}, \bibinfo{author}{S.~Cresci}, \bibinfo{author}{T.~Chakraborty}, \bibinfo{author}{F.~Silvestri}, \bibinfo{author}{D.~Dimitrov}, \bibinfo{author}{G.~D.~S. Martino}, \bibinfo{author}{S.~Shaar}, \bibinfo{author}{H.~Firooz}, \bibinfo{author}{P.~Nakov},
\newblock \bibinfo{title}{A survey on multimodal disinformation detection},
\newblock \bibinfo{journal}{arXiv preprint arXiv:2103.12541}  (\bibinfo{year}{2021}).
\bibitem[{Anjum and Katarya(2024)}]{anjum2024hate}
\bibinfo{author}{Anjum}, \bibinfo{author}{R.~Katarya},
\newblock \bibinfo{title}{Hate speech, toxicity detection in online social media: a recent survey of state of the art and opportunities},
\newblock \bibinfo{journal}{International Journal of Information Security} \bibinfo{volume}{23} (\bibinfo{year}{2024}) \bibinfo{pages}{577--608}.
\bibitem[{Lewandowska-Tomaszczyk et~al.(2023)Lewandowska-Tomaszczyk, B{\k{a}}czkowska, Liebeskind, Valunaite~O., and {\v{Z}}itnik}]{lewandowska2023integrated}
\bibinfo{author}{B.~Lewandowska-Tomaszczyk}, \bibinfo{author}{A.~B{\k{a}}czkowska}, \bibinfo{author}{C.~Liebeskind}, \bibinfo{author}{G.~Valunaite~O.}, \bibinfo{author}{S.~{\v{Z}}itnik},
\newblock \bibinfo{title}{An integrated explicit and implicit offensive language taxonomy},
\newblock \bibinfo{journal}{Lodz Papers in Pragmatics} \bibinfo{volume}{19} (\bibinfo{year}{2023}) \bibinfo{pages}{7--48}.
\bibitem[{Garg et~al.(2023)Garg, Masud, Suresh, and Chakraborty}]{garg2023handling}
\bibinfo{author}{T.~Garg}, \bibinfo{author}{S.~Masud}, \bibinfo{author}{T.~Suresh}, \bibinfo{author}{T.~Chakraborty},
\newblock \bibinfo{title}{Handling bias in toxic speech detection: A survey},
\newblock \bibinfo{journal}{ACM Computing Surveys} \bibinfo{volume}{55} (\bibinfo{year}{2023}) \bibinfo{pages}{1--32}.
\bibitem[{Lin et~al.(2024)Lin, Luo, Gao, Ma, Wang, and Yang}]{lin2024towards}
\bibinfo{author}{H.~Lin}, \bibinfo{author}{Z.~Luo}, \bibinfo{author}{W.~Gao}, \bibinfo{author}{J.~Ma}, \bibinfo{author}{B.~Wang}, \bibinfo{author}{R.~Yang},
\newblock \bibinfo{title}{Towards explainable harmful meme detection through multimodal debate between large language models},
\newblock \bibinfo{journal}{arXiv preprint arXiv:2401.13298}  (\bibinfo{year}{2024}).
\bibitem[{Jha et~al.(2024)Jha, Maity, Jain, Verma, Saha, and Bhattacharyya}]{jha2024meme}
\bibinfo{author}{P.~Jha}, \bibinfo{author}{K.~Maity}, \bibinfo{author}{R.~Jain}, \bibinfo{author}{A.~Verma}, \bibinfo{author}{S.~Saha}, \bibinfo{author}{P.~Bhattacharyya},
\newblock \bibinfo{title}{Meme-ingful analysis: Enhanced understanding of cyberbullying in memes through multimodal explanations},
\newblock \bibinfo{journal}{arXiv preprint arXiv:2401.09899}  (\bibinfo{year}{2024}).
\bibitem[{Lin et~al.(2023)Lin, Luo, Ma, and Chen}]{lin2023beneath}
\bibinfo{author}{H.~Lin}, \bibinfo{author}{Z.~Luo}, \bibinfo{author}{J.~Ma}, \bibinfo{author}{L.~Chen},
\newblock \bibinfo{title}{Beneath the surface: Unveiling harmful memes with multimodal reasoning distilled from large language models},
\newblock \bibinfo{journal}{arXiv preprint arXiv:2312.05434} \bibinfo{volume}{9114} (\bibinfo{year}{2023}).
\bibitem[{Chen et~al.(2023)Chen, Zhu, Shen, Li, Liu, Zhang, Krishnamoorthi, Chandra, Xiong, and Elhoseiny}]{chen2023minigpt}
\bibinfo{author}{J.~Chen}, \bibinfo{author}{D.~Zhu}, \bibinfo{author}{X.~Shen}, \bibinfo{author}{X.~Li}, \bibinfo{author}{Z.~Liu}, \bibinfo{author}{P.~Zhang}, \bibinfo{author}{R.~Krishnamoorthi}, \bibinfo{author}{V.~Chandra}, \bibinfo{author}{Y.~Xiong}, \bibinfo{author}{M.~Elhoseiny},
\newblock \bibinfo{title}{Minigpt-v2: large language model as a unified interface for vision-language multi-task learning},
\newblock \bibinfo{journal}{arXiv preprint arXiv:2310.09478}  (\bibinfo{year}{2023}).
\bibitem[{Dai et~al.(2023)Dai, Li, Li, Tiong, Zhao, Wang, Li, Fung, and Hoi}]{dai2023instructblip}
\bibinfo{author}{W.~Dai}, \bibinfo{author}{J.~Li}, \bibinfo{author}{D.~Li}, \bibinfo{author}{A.~Tiong}, \bibinfo{author}{J.~Zhao}, \bibinfo{author}{W.~Wang}, \bibinfo{author}{B.~Li}, \bibinfo{author}{P.~N. Fung}, \bibinfo{author}{S.~Hoi},
\newblock \bibinfo{title}{Instructblip: Towards general-purpose vision-language models with instruction tuning},
\newblock in: \bibinfo{booktitle}{Poster at Advances in Neural Information Processing Systems}, \bibinfo{year}{2023}.
\bibitem[{Hu et~al.(2023)Hu, Stretcu, Lu, Viswanathan, Hata, Luo, Krishna, and Fuxman}]{hu2023visual}
\bibinfo{author}{Y.~Hu}, \bibinfo{author}{O.~Stretcu}, \bibinfo{author}{C.-T. Lu}, \bibinfo{author}{K.~Viswanathan}, \bibinfo{author}{K.~Hata}, \bibinfo{author}{E.~Luo}, \bibinfo{author}{R.~Krishna}, \bibinfo{author}{A.~Fuxman},
\newblock \bibinfo{title}{Visual program distillation: Distilling tools and programmatic reasoning into vision-language models},
\newblock \bibinfo{journal}{arXiv preprint arXiv:2312.03052}  (\bibinfo{year}{2023}).
\bibitem[{Wang et~al.(2023)Wang, Luo, Yang, Hong, and Luo}]{wang2023gpt}
\bibinfo{author}{J.~Wang}, \bibinfo{author}{J.~Luo}, \bibinfo{author}{G.~Yang}, \bibinfo{author}{A.~Hong}, \bibinfo{author}{F.~Luo},
\newblock \bibinfo{title}{Is gpt powerful enough to analyze the emotions of memes?},
\newblock \bibinfo{journal}{arXiv preprint arXiv:2311.00223}  (\bibinfo{year}{2023}).
\bibitem[{Wu et~al.(2023)Wu, Yu, Backes, Shen, and Zhang}]{wu2023proactive}
\bibinfo{author}{Y.~Wu}, \bibinfo{author}{N.~Yu}, \bibinfo{author}{M.~Backes}, \bibinfo{author}{Y.~Shen}, \bibinfo{author}{Y.~Zhang},
\newblock \bibinfo{title}{On the proactive generation of unsafe images from text-to-image models using benign prompts},
\newblock \bibinfo{journal}{arXiv:2310.16613}  (\bibinfo{year}{2023}).
\bibitem[{Lin et~al.(2024)Lin, Luo, Wang, Yang, and Ma}]{lin2024goat}
\bibinfo{author}{H.~Lin}, \bibinfo{author}{Z.~Luo}, \bibinfo{author}{B.~Wang}, \bibinfo{author}{R.~Yang}, \bibinfo{author}{J.~Ma},
\newblock \bibinfo{title}{Goat-bench: Safety insights to large multimodal models through meme-based social abuse},
\newblock \bibinfo{journal}{arXiv preprint arXiv:2401.01523}  (\bibinfo{year}{2024}).
\bibitem[{Hossain et~al.(2022)Hossain, Sharif, Hoque, Akber~Dewan, Siddique, and Hossain}]{hossain2022identification}
\bibinfo{author}{E.~Hossain}, \bibinfo{author}{O.~Sharif}, \bibinfo{author}{M.~Hoque}, \bibinfo{author}{M.~Akber~Dewan}, \bibinfo{author}{N.~Siddique}, \bibinfo{author}{M.~Hossain},
\newblock \bibinfo{title}{Identification of multilingual offense and troll from social media memes using weighted ensemble of multimodal features},
\newblock in: \bibinfo{booktitle}{Journal of King Saud University - Computer and Information Sciences}, volume~\bibinfo{volume}{34}, \bibinfo{year}{2022}, pp. \bibinfo{pages}{6605--6623}. \DOIprefix\doi{10.1016/j.jksuci.2022.06.010}.
\bibitem[{Bhandari et~al.(2023)Bhandari, Shah, Thapa, Naseem, and Nasim}]{bhandari2023crisishatemm}
\bibinfo{author}{A.~Bhandari}, \bibinfo{author}{S.~B. Shah}, \bibinfo{author}{S.~Thapa}, \bibinfo{author}{U.~Naseem}, \bibinfo{author}{M.~Nasim},
\newblock \bibinfo{title}{Crisishatemm: Multimodal analysis of directed and undirected hate speech in text-embedded images from russia-ukraine conflict},
\newblock in: \bibinfo{booktitle}{Proceedings of the IEEE/CVF Conference on Computer Vision and Pattern Recognition}, \bibinfo{year}{2023}, pp. \bibinfo{pages}{1993--2002}.
\bibitem[{Bhowmick et~al.(2022)Bhowmick, Ganguli, Paul, and Sil}]{bhowmick2022multimodal}
\bibinfo{author}{R.~Bhowmick}, \bibinfo{author}{I.~Ganguli}, \bibinfo{author}{J.~Paul}, \bibinfo{author}{J.~Sil},
\newblock \bibinfo{title}{A multimodal deep framework for derogatory social media post identification of a recognized person},
\newblock \bibinfo{journal}{ACM Transactions on Asian and Low-Resource Language Information Processing} \bibinfo{volume}{21} (\bibinfo{year}{2022}) \bibinfo{pages}{3447651}.
\bibitem[{Qu et~al.(2022)Qu, Li, Zhao, Dev, and Chang}]{qu2022disinfomeme}
\bibinfo{author}{J.~Qu}, \bibinfo{author}{L.~H. Li}, \bibinfo{author}{J.~Zhao}, \bibinfo{author}{S.~Dev}, \bibinfo{author}{K.-W. Chang},
\newblock \bibinfo{title}{Disinfomeme: A multimodal dataset for detecting meme intentionally spreading out disinformation},
\newblock \bibinfo{journal}{arXiv preprint arXiv:2205.12617}  (\bibinfo{year}{2022}).
\bibitem[{Zhang et~al.(2023)Zhang, Feng, Gu, and Chang}]{zhang2023mvlp}
\bibinfo{author}{N.~Zhang}, \bibinfo{author}{X.~Feng}, \bibinfo{author}{T.~Gu}, \bibinfo{author}{L.~Chang},
\newblock \bibinfo{title}{Mvlp: Multi-perspective vision-language pre-training model for ethically aligned meme detection},
\newblock \bibinfo{journal}{Authorea Preprints}  (\bibinfo{year}{2023}).
\bibitem[{Kumari et~al.(2023)Kumari, Bandyopadhyay, and Ekbal}]{kumari2023emoffmeme}
\bibinfo{author}{G.~Kumari}, \bibinfo{author}{D.~Bandyopadhyay}, \bibinfo{author}{A.~Ekbal},
\newblock \bibinfo{title}{Emoffmeme: identifying offensive memes by leveraging underlying emotions},
\newblock in: \bibinfo{booktitle}{Multimedia Tools and Applications}, volume~\bibinfo{volume}{82}, \bibinfo{year}{2023}, pp. \bibinfo{pages}{45061--45096}. \DOIprefix\doi{10.1007/s11042-023-14807-1}.
\bibitem[{Sharma et~al.(2022)Sharma, Akhtar, Nakov, and Chakraborty}]{sharma2022disarm}
\bibinfo{author}{S.~Sharma}, \bibinfo{author}{M.~S. Akhtar}, \bibinfo{author}{P.~Nakov}, \bibinfo{author}{T.~Chakraborty},
\newblock \bibinfo{title}{Disarm: Detecting the victims targeted by harmful memes},
\newblock \bibinfo{journal}{arXiv preprint arXiv:2205.05738}  (\bibinfo{year}{2022}) \bibinfo{pages}{1572--1588}.
\bibitem[{Jabiyev et~al.(2021)Jabiyev, Onaolapo, Stringhini, and Kirda}]{jabiyev2021game}
\bibinfo{author}{B.~Jabiyev}, \bibinfo{author}{J.~Onaolapo}, \bibinfo{author}{G.~Stringhini}, \bibinfo{author}{E.~Kirda},
\newblock \bibinfo{title}{e-game of fame: Automatic detection of fake memes.},
\newblock in: \bibinfo{booktitle}{TTO}, \bibinfo{year}{2021}, pp. \bibinfo{pages}{1--11}.
\bibitem[{Mathias et~al.(2021)Mathias, Nie, Davani, Kiela, Prabhakaran, Vidgen, and Waseem}]{woah2021shared}
\bibinfo{author}{L.~Mathias}, \bibinfo{author}{S.~Nie}, \bibinfo{author}{A.~M. Davani}, \bibinfo{author}{D.~Kiela}, \bibinfo{author}{V.~Prabhakaran}, \bibinfo{author}{B.~Vidgen}, \bibinfo{author}{Z.~Waseem},
\newblock \bibinfo{title}{Findings of the woah 5 shared task on fine grained hateful memes detection},
\newblock in: \bibinfo{booktitle}{Proceedings of the 5th Workshop on Online Abuse and Harms (WOAH 2021)}, \bibinfo{year}{2021}, pp. \bibinfo{pages}{201--206}.
\bibitem[{Pramanick et~al.(2021)Pramanick, Sharma, Dimitrov, Nakov, and Chakraborty}]{pramanick2021momenta}
\bibinfo{author}{S.~Pramanick}, \bibinfo{author}{S.~Sharma}, \bibinfo{author}{D.~Dimitrov}, \bibinfo{author}{P.~Nakov}, \bibinfo{author}{T.~Chakraborty},
\newblock \bibinfo{title}{Momenta: A multimodal framework for detecting harmful memes and their targets},
\newblock \bibinfo{journal}{arXiv}  (\bibinfo{year}{2021}).
\bibitem[{Sabat et~al.(2019)Sabat, Ferrer, and Giro-I-Nieto}]{sabat2019hate}
\bibinfo{author}{B.~O. Sabat}, \bibinfo{author}{C.~C. Ferrer}, \bibinfo{author}{X.~Giro-I-Nieto},
\newblock \bibinfo{title}{Hate speech in pixels: Detection of offensive memes towards automatic moderation},
\newblock \bibinfo{journal}{arXiv}  (\bibinfo{year}{2019}).
\bibitem[{Hee et~al.(2023)Hee, Chong, and Lee}]{hee2023decoding}
\bibinfo{author}{M.~Hee}, \bibinfo{author}{W.-H. Chong}, \bibinfo{author}{R.-W. Lee},
\newblock \bibinfo{title}{Decoding the underlying meaning of multimodal hateful memes},
\newblock in: \bibinfo{booktitle}{IJCAI International Joint Conference on Artificial Intelligence}, volume \bibinfo{volume}{2023-August}, \bibinfo{year}{2023}.
\bibitem[{Sharma et~al.(2022)Sharma, Suresh, Kulkarni, Mathur, Nakov, Akhtar, and Chakraborty}]{sharma2022findings}
\bibinfo{author}{S.~Sharma}, \bibinfo{author}{T.~Suresh}, \bibinfo{author}{A.~Kulkarni}, \bibinfo{author}{H.~Mathur}, \bibinfo{author}{P.~Nakov}, \bibinfo{author}{M.~S. Akhtar}, \bibinfo{author}{T.~Chakraborty},
\newblock \bibinfo{title}{Findings of the constraint 2022 shared task on detecting the hero, the villain, and the victim in memes},
\newblock in: \bibinfo{booktitle}{Proceedings of the Workshop on Combating Online Hostile Posts in Regional Languages during Emergency Situations}, \bibinfo{organization}{Constraint}, \bibinfo{year}{2022}, pp. \bibinfo{pages}{1--11}.
\bibitem[{Rajput et~al.(2022)Rajput, Kapoor, Rai, and Kaur}]{rajput2022hate}
\bibinfo{author}{K.~Rajput}, \bibinfo{author}{R.~Kapoor}, \bibinfo{author}{K.~K. Rai}, \bibinfo{author}{P.~Kaur},
\newblock \bibinfo{title}{Hate me not: Detecting hate inducing memes in code switched languages},
\newblock \bibinfo{journal}{arXiv}  (\bibinfo{year}{2022}).
\bibitem[{Badour and Brown(2021)}]{Badour2021Hateful}
\bibinfo{author}{J.~Badour}, \bibinfo{author}{J.~Brown},
\newblock \bibinfo{title}{Hateful memes classification using machine learning},
\newblock in: \bibinfo{booktitle}{2021 IEEE Symposium Series on Computational Intelligence, SSCI 2021 - Proceedings}, volume~\bibinfo{volume}{2}, \bibinfo{year}{2021}. \DOIprefix\doi{10.1109/ssci50451.2021.9659896}.
\bibitem[{Bacha et~al.(2023)Bacha, Ullah, Khan, Sardar, and Lee}]{bacha2023deep}
\bibinfo{author}{J.~Bacha}, \bibinfo{author}{F.~Ullah}, \bibinfo{author}{J.~Khan}, \bibinfo{author}{A.~Sardar}, \bibinfo{author}{S.~Lee},
\newblock \bibinfo{title}{A deep learning-based framework for offensive text detection in unstructured data for heterogeneous social media},
\newblock in: \bibinfo{booktitle}{IEEE Access}, volume~\bibinfo{volume}{11}, \bibinfo{year}{2023}, pp. \bibinfo{pages}{124484--124498}. \DOIprefix\doi{10.1109/access.2023.3330081}.
\bibitem[{Alzu’bi et~al.(2023)Alzu’bi, Bani~Younis, Abuarqoub, and Hammoudeh}]{alzu2023multimodal}
\bibinfo{author}{A.~Alzu’bi}, \bibinfo{author}{L.~Bani~Younis}, \bibinfo{author}{A.~Abuarqoub}, \bibinfo{author}{M.~Hammoudeh},
\newblock \bibinfo{title}{Multimodal deep learning with discriminant descriptors for offensive memes detection},
\newblock \bibinfo{journal}{ACM Journal of Data and Information Quality} \bibinfo{volume}{15} (\bibinfo{year}{2023}) \bibinfo{pages}{1--16}.
\bibitem[{Ramamoorthy et~al.(2022)Ramamoorthy, Gunti, Mishra, Suryavardan, Reganti, Patwa, DaS, Chakraborty, Sheth, Ekbal, and A}]{ramamoorthy2022memotion}
\bibinfo{author}{S.~Ramamoorthy}, \bibinfo{author}{N.~Gunti}, \bibinfo{author}{S.~Mishra}, \bibinfo{author}{S.~Suryavardan}, \bibinfo{author}{A.~Reganti}, \bibinfo{author}{P.~Patwa}, \bibinfo{author}{A.~DaS}, \bibinfo{author}{T.~Chakraborty}, \bibinfo{author}{A.~Sheth}, \bibinfo{author}{A.~Ekbal}, \bibinfo{author}{C.~A},
\newblock \bibinfo{title}{Memotion 2: Dataset on sentiment and emotion analysis of memes},
\newblock in: \bibinfo{booktitle}{Proceedings of De-Factify: Workshop on Multimodal Fact Checking and Hate Speech Detection, CEUR}, \bibinfo{year}{2022}.
\bibitem[{Xu et~al.(2022)Xu, Li, Zheng, Naseriparsa, Zhao, Lin, and Xia}]{xu2022met}
\bibinfo{author}{B.~Xu}, \bibinfo{author}{T.~Li}, \bibinfo{author}{J.~Zheng}, \bibinfo{author}{M.~Naseriparsa}, \bibinfo{author}{Z.~Zhao}, \bibinfo{author}{H.~Lin}, \bibinfo{author}{F.~Xia},
\newblock \bibinfo{title}{Met-meme: A multimodal meme dataset rich in metaphors},
\newblock in: \bibinfo{booktitle}{Proceedings of the 45th International ACM SIGIR Conference on Research and Development in Information Retrieval}, \bibinfo{year}{2022}, pp. \bibinfo{pages}{2887--2899}.
\bibitem[{Gasparini et~al.(2022)Gasparini, Rizzi, Saibene, and Fersini}]{gasparini2022benchmark}
\bibinfo{author}{F.~Gasparini}, \bibinfo{author}{G.~Rizzi}, \bibinfo{author}{A.~Saibene}, \bibinfo{author}{E.~Fersini},
\newblock \bibinfo{title}{Benchmark dataset of memes with text transcriptions for automatic detection of multi-modal misogynistic content},
\newblock \bibinfo{journal}{Data in brief} \bibinfo{volume}{44} (\bibinfo{year}{2022}) \bibinfo{pages}{108526}.
\bibitem[{Maity et~al.(2022)Maity, Jha, Saha, and Bhattacharyya}]{maity2022multitask}
\bibinfo{author}{K.~Maity}, \bibinfo{author}{P.~Jha}, \bibinfo{author}{S.~Saha}, \bibinfo{author}{P.~Bhattacharyya},
\newblock \bibinfo{title}{A multitask framework for sentiment, emotion and sarcasm aware cyberbullying detection from multi-modal code-mixed memes},
\newblock in: \bibinfo{booktitle}{Proceedings of the 45th International ACM SIGIR Conference on Research and Development in Information Retrieval}, \bibinfo{year}{2022}, pp. \bibinfo{pages}{1739--1749}.
\bibitem[{Fersini et~al.(2022)Fersini, Gasparini, Rizzi, Saibene, Chulvi, Rosso, Lees, and Sorensen}]{fersini2022semeval}
\bibinfo{author}{E.~Fersini}, \bibinfo{author}{F.~Gasparini}, \bibinfo{author}{G.~Rizzi}, \bibinfo{author}{A.~Saibene}, \bibinfo{author}{B.~Chulvi}, \bibinfo{author}{P.~Rosso}, \bibinfo{author}{A.~Lees}, \bibinfo{author}{J.~Sorensen},
\newblock \bibinfo{title}{Semeval-2022 task 5: Multimedia automatic misogyny identification},
\newblock in: \bibinfo{booktitle}{Proceedings of the 16th International Workshop on Semantic Evaluation (SemEval-2022)}, \bibinfo{year}{2022}, pp. \bibinfo{pages}{533--549}.
\bibitem[{Suryawanshi et~al.(2020)Suryawanshi, Chakravarthi, Arcan, and Buitelaar}]{suryawanshi2020multimodal}
\bibinfo{author}{S.~Suryawanshi}, \bibinfo{author}{B.~R. Chakravarthi}, \bibinfo{author}{M.~Arcan}, \bibinfo{author}{P.~Buitelaar},
\newblock \bibinfo{title}{Multimodal meme dataset (multioff) for identifying offensive content in image and text},
\newblock in: \bibinfo{booktitle}{Proceedings of the second workshop on trolling, aggression and cyberbullying}, \bibinfo{year}{2020}, pp. \bibinfo{pages}{32--41}.
\bibitem[{Kumari et~al.(2024)Kumari, Sinha, Ekbal, Chatterjee, and Vinutha}]{kumari2024enhancing}
\bibinfo{author}{G.~Kumari}, \bibinfo{author}{A.~Sinha}, \bibinfo{author}{A.~Ekbal}, \bibinfo{author}{A.~Chatterjee}, \bibinfo{author}{B.~Vinutha},
\newblock \bibinfo{title}{Enhancing the fairness of offensive memes detection models by mitigating unintended political bias},
\newblock \bibinfo{journal}{Journal of Intelligent Information Systems}  (\bibinfo{year}{2024}) \bibinfo{pages}{1}.
\bibitem[{Suryawanshi et~al.(2020)Suryawanshi, Chakravarthi, Verma, Arcan, McCrae, and Buitelaar}]{suryawanshi2020dataset}
\bibinfo{author}{S.~Suryawanshi}, \bibinfo{author}{B.~R. Chakravarthi}, \bibinfo{author}{P.~Verma}, \bibinfo{author}{M.~Arcan}, \bibinfo{author}{J.~P. McCrae}, \bibinfo{author}{P.~Buitelaar},
\newblock \bibinfo{title}{A dataset for troll classification of tamilmemes},
\newblock in: \bibinfo{booktitle}{Proceedings of the WILDRE5--5th workshop on indian language data: resources and evaluation}, \bibinfo{year}{2020}, pp. \bibinfo{pages}{7--13}.
\bibitem[{Fersini et~al.(2022)Fersini, Rizzi, Saibene, and Gasparini}]{fersini2022misogynous}
\bibinfo{author}{E.~Fersini}, \bibinfo{author}{G.~Rizzi}, \bibinfo{author}{A.~Saibene}, \bibinfo{author}{F.~Gasparini},
\newblock \bibinfo{title}{Misogynous meme recognition: A preliminary study},
\newblock in: \bibinfo{booktitle}{Lecture Notes in Computer Science}, volume \bibinfo{volume}{13196} of \textit{\bibinfo{series}{Lecture Notes in Computer Science}}, \bibinfo{year}{2022}, pp. \bibinfo{pages}{279--293}. \DOIprefix\doi{10.1007/978-3-031-08421-8\_19}.
\bibitem[{Kiela et~al.(2020)Kiela, Firooz, Mohan, Goswami, Singh, Ringshia, and Testuggine}]{Kiela2020hateful_challenge}
\bibinfo{author}{D.~Kiela}, \bibinfo{author}{H.~Firooz}, \bibinfo{author}{A.~Mohan}, \bibinfo{author}{V.~Goswami}, \bibinfo{author}{A.~Singh}, \bibinfo{author}{P.~Ringshia}, \bibinfo{author}{D.~Testuggine},
\newblock \bibinfo{title}{The hateful memes challenge: Detecting hate speech in multimodal memes},
\newblock in: \bibinfo{booktitle}{Advances in Neural Information Processing Systems}, volume \bibinfo{volume}{2020-December}, \bibinfo{year}{2020}.
\bibitem[{Das and Mukherjee(2023)}]{das2023transfer}
\bibinfo{author}{M.~Das}, \bibinfo{author}{A.~Mukherjee},
\newblock \bibinfo{title}{Transfer learning for multilingual abusive meme detection},
\newblock in: \bibinfo{booktitle}{ACM International Conference Proceeding Series}, \bibinfo{year}{2023}, pp. \bibinfo{pages}{245--250}. \DOIprefix\doi{10.1145/3578503.3583607}.
\bibitem[{Titli and Paul(2023)}]{titli2023automated}
\bibinfo{author}{S.~R. Titli}, \bibinfo{author}{S.~Paul},
\newblock \bibinfo{title}{Automated bengali abusive text classification: Using deep learning techniques},
\newblock in: \bibinfo{booktitle}{2023 International Conference on Advances in Electronics, Communication, Computing and Intelligent Information Systems (ICAECIS)}, \bibinfo{organization}{Ieee}, \bibinfo{year}{2023}, pp. \bibinfo{pages}{1--6}.
\bibitem[{Smith et~al.(2008)Smith, Mahdavi, Carvalho, Fisher, Russell, and Tippett}]{smith2008cyberbullying}
\bibinfo{author}{P.~K. Smith}, \bibinfo{author}{J.~Mahdavi}, \bibinfo{author}{M.~Carvalho}, \bibinfo{author}{S.~Fisher}, \bibinfo{author}{S.~Russell}, \bibinfo{author}{N.~Tippett},
\newblock \bibinfo{title}{Cyberbullying: Its nature and impact in secondary school pupils},
\newblock \bibinfo{journal}{Journal of child psychology and psychiatry} \bibinfo{volume}{49} (\bibinfo{year}{2008}) \bibinfo{pages}{376--385}.
\bibitem[{Kumar et~al.(2023)Kumar, Ahmed, Prashanth, Vinaykumar, Babu, and Kumar}]{kumar2023efficient}
\bibinfo{author}{M.~N. Kumar}, \bibinfo{author}{D.~M. Ahmed}, \bibinfo{author}{J.~Prashanth}, \bibinfo{author}{V.~Vinaykumar}, \bibinfo{author}{J.~A. Babu}, \bibinfo{author}{T.~K. Kumar},
\newblock \bibinfo{title}{An efficient deep learning approach to deal with cyberbullying},
\newblock in: \bibinfo{booktitle}{2023 2nd International Conference on Computational Modelling, Simulation and Optimization (ICCMSO)}, \bibinfo{organization}{IEEE}, \bibinfo{year}{2023}, pp. \bibinfo{pages}{253--258}.
\bibitem[{Ji et~al.(2023)Ji, Ren, and Naseem}]{Ji2023Identifying}
\bibinfo{author}{J.~Ji}, \bibinfo{author}{W.~Ren}, \bibinfo{author}{U.~Naseem},
\newblock \bibinfo{title}{Identifying creative harmful memes via prompt based approach},
\newblock in: \bibinfo{booktitle}{ACM Web Conference 2023 - Proceedings of the World Wide Web Conference, WWW 2023}, volume \bibinfo{volume}{3868}, \bibinfo{year}{2023}.
\bibitem[{Fharook et~al.(2022)Fharook, Ahmed, Rithika, Budde, Saumya, and Biradar}]{fharook2022are}
\bibinfo{author}{S.~Fharook}, \bibinfo{author}{S.~Ahmed}, \bibinfo{author}{G.~Rithika}, \bibinfo{author}{S.~Budde}, \bibinfo{author}{S.~Saumya}, \bibinfo{author}{S.~Biradar},
\newblock \bibinfo{title}{Are you a hero or a villain? a semantic role labelling approach for detecting harmful memes},
\newblock in: \bibinfo{booktitle}{CONSTRAINT 2022 - 2nd Workshop on Combating Online Hostile Posts in Regional Languages during Emergency Situation, Proceedings of the Workshop}, \bibinfo{organization}{Constraint}, \bibinfo{year}{2022}, pp. \bibinfo{pages}{19--23}.
\bibitem[{Chakraborty et~al.(2022)Chakraborty, Akhtar, Shu, Bernard, Liakata, and Nakov}]{constraint2022}
\bibinfo{editor}{T.~Chakraborty}, \bibinfo{editor}{M.~S. Akhtar}, \bibinfo{editor}{K.~Shu}, \bibinfo{editor}{H.~R. Bernard}, \bibinfo{editor}{M.~Liakata}, \bibinfo{editor}{P.~Nakov} (Eds.), \bibinfo{title}{CONSTRAINT 2022 - 2nd Workshop on Combating Online Hostile Posts in Regional Languages during Emergency Situation, Proceedings of the Workshop}, \bibinfo{year}{2022}.
\bibitem[{Nandi et~al.(2022)Nandi, Alam, and Nakov}]{nandi2022detecting}
\bibinfo{author}{R.~Nandi}, \bibinfo{author}{F.~Alam}, \bibinfo{author}{P.~Nakov},
\newblock \bibinfo{title}{Detecting the role of an entity in harmful memes: Techniques and their limitations},
\newblock \bibinfo{journal}{CONSTRAINT 2022 - 2nd Workshop on Combating Online Hostile Posts in Regional Languages during Emergency Situation, Proceedings of the Workshop}  (\bibinfo{year}{2022}) \bibinfo{pages}{43--54}.
\bibitem[{Sharma et~al.(2023)Sharma, Kulkarni, Suresh, Mathur, Nakov, Akhtar, and Chakraborty}]{sharma2023characterizing}
\bibinfo{author}{S.~Sharma}, \bibinfo{author}{A.~Kulkarni}, \bibinfo{author}{T.~Suresh}, \bibinfo{author}{H.~Mathur}, \bibinfo{author}{P.~Nakov}, \bibinfo{author}{M.~Akhtar}, \bibinfo{author}{T.~Chakraborty},
\newblock \bibinfo{title}{Characterizing the entities in harmful memes: Who is the hero, the villain, the victim?},
\newblock in: \bibinfo{booktitle}{EACL 2023 - 17th Conference of the European Chapter of the Association for Computational Linguistics, Proceedings of the Conference}, \bibinfo{year}{2023}, pp. \bibinfo{pages}{2141--2155}.
\bibitem[{Singh et~al.(2022)Singh, Maladry, and Lefever}]{singh2022combining}
\bibinfo{author}{P.~Singh}, \bibinfo{author}{A.~Maladry}, \bibinfo{author}{E.~Lefever},
\newblock \bibinfo{title}{Combining language models and linguistic information to label entities in memes},
\newblock in: \bibinfo{booktitle}{CONSTRAINT 2022 - 2nd Workshop on Combating Online Hostile Posts in Regional Languages during Emergency Situation, Proceedings of the Workshop}, \bibinfo{organization}{Constraint}, \bibinfo{year}{2022}, pp. \bibinfo{pages}{35--42}.
\bibitem[{Zhou et~al.(2022)Zhou, Zhao, Dong, Gao, and Liu}]{zhou2022ddtig}
\bibinfo{author}{Z.~Zhou}, \bibinfo{author}{H.~Zhao}, \bibinfo{author}{J.~Dong}, \bibinfo{author}{J.~Gao}, \bibinfo{author}{X.~Liu},
\newblock \bibinfo{title}{Dd-tig at constraint\@acl2022: Multimodal understanding and reasoning for role labeling of entities in hateful memes},
\newblock in: \bibinfo{booktitle}{Proceedings of CONSTRAINT 2022 - 2nd Workshop on Combating Online Hostile Posts in Regional Languages during Emergency Situation}, \bibinfo{organization}{Constraint}, \bibinfo{year}{2022}, pp. \bibinfo{pages}{12--18}.
\bibitem[{Sharma et~al.(2022)Sharma, Siddiqui, Akhtar, and Chakraborty}]{sharma2022domain}
\bibinfo{author}{S.~Sharma}, \bibinfo{author}{M.~K. Siddiqui}, \bibinfo{author}{M.~S. Akhtar}, \bibinfo{author}{T.~Chakraborty},
\newblock \bibinfo{title}{Domain-aware self-supervised pre-training for label-efficient meme analysis},
\newblock \bibinfo{journal}{arXiv}  (\bibinfo{year}{2022}).
\bibitem[{Koutlis et~al.(2023)Koutlis, Schinas, and Papadopoulos}]{koutlis2023memefier}
\bibinfo{author}{C.~Koutlis}, \bibinfo{author}{M.~Schinas}, \bibinfo{author}{S.~Papadopoulos},
\newblock \bibinfo{title}{Memefier: Dual-stage modality fusion for image meme},
\newblock in: \bibinfo{booktitle}{ICMR 2023 - Proceedings of the 2023 ACM International Conference on Multimedia Retrieval}, \bibinfo{year}{2023}, pp. \bibinfo{pages}{586--591}. \DOIprefix\doi{10.1145/3591106.3592254}.
\bibitem[{Zia et~al.(2021)Zia, Castro, and Tyson}]{zia2021racist}
\bibinfo{author}{H.~B. Zia}, \bibinfo{author}{I.~Castro}, \bibinfo{author}{G.~Tyson},
\newblock \bibinfo{title}{Racist or sexist meme? classifying memes beyond hateful},
\newblock in: \bibinfo{booktitle}{Proceedings of the 5th Workshop on Online Abuse and Harms (WOAH 2021)}, volume \bibinfo{volume}{215}, \bibinfo{year}{2021}, pp. \bibinfo{pages}{215--219}.
\bibitem[{Armenta-Segura et~al.(2023)Armenta-Segura, N\'{u}\~{n}ez Prado, Sidorov, Gelbukh, and Rom\'{a}n-God\'{\i}nez}]{Armenta-Segura2023Ometeotl}
\bibinfo{author}{J.~Armenta-Segura}, \bibinfo{author}{C.-J. N\'{u}\~{n}ez Prado}, \bibinfo{author}{G.~Sidorov}, \bibinfo{author}{A.~Gelbukh}, \bibinfo{author}{R.~Rom\'{a}n-God\'{\i}nez},
\newblock \bibinfo{title}{Ometeotl{\@}multimodal hate speech event detection 2023: Hate speech and text-image correlation detection in real life memes using pre-trained bert models over text},
\newblock in: \bibinfo{booktitle}{CASE 2023 - Proceedings of the 6th Workshop on Challenges and Applications of Automated Extraction of Socio-Political Events from Text at RANLP}, volume~\bibinfo{volume}{53}, \bibinfo{year}{2023}. \DOIprefix\doi{10.26615/978-954-452-089-2\_007}.
\bibitem[{Aggarwal et~al.(2021)Aggarwal, Sharma, Trivedi, Yadav, Agrawal, Singh, Mishra, and Gritli}]{Aggarwal2021Two}
\bibinfo{author}{A.~Aggarwal}, \bibinfo{author}{V.~Sharma}, \bibinfo{author}{A.~Trivedi}, \bibinfo{author}{M.~Yadav}, \bibinfo{author}{C.~Agrawal}, \bibinfo{author}{D.~Singh}, \bibinfo{author}{V.~Mishra}, \bibinfo{author}{H.~Gritli},
\newblock \bibinfo{title}{Two-way feature extraction using sequential and multimodal approach for hateful meme classification},
\newblock \bibinfo{journal}{Complexity} \bibinfo{volume}{2021} (\bibinfo{year}{2021}).
\bibitem[{Arya et~al.(2024)Arya, Hasan, Bagwari, Safie, Islam, Ahmed, De, Khan, and Ghazal}]{arya2024multimodal}
\bibinfo{author}{G.~Arya}, \bibinfo{author}{M.~Hasan}, \bibinfo{author}{A.~Bagwari}, \bibinfo{author}{N.~Safie}, \bibinfo{author}{S.~Islam}, \bibinfo{author}{F.~Ahmed}, \bibinfo{author}{A.~De}, \bibinfo{author}{M.~Khan}, \bibinfo{author}{T.~Ghazal},
\newblock \bibinfo{title}{Multimodal hate speech detection in memes using contrastive language-image pre-training},
\newblock \bibinfo{journal}{IEEE Access} \bibinfo{volume}{12} (\bibinfo{year}{2024}) \bibinfo{pages}{22359--22375}.
\bibitem[{Aggarwal et~al.(2023)Aggarwal, Chawla, Das, Saha, Mathew, Zesch, and Mukherjee}]{Aggarwal2023HateProof}
\bibinfo{author}{P.~Aggarwal}, \bibinfo{author}{P.~Chawla}, \bibinfo{author}{M.~Das}, \bibinfo{author}{P.~Saha}, \bibinfo{author}{B.~Mathew}, \bibinfo{author}{T.~Zesch}, \bibinfo{author}{A.~Mukherjee},
\newblock \bibinfo{title}{Hateproof: Are hateful meme detection systems really robust?},
\newblock in: \bibinfo{booktitle}{ACM Web Conference 2023 - Proceedings of the World Wide Web Conference, WWW 2023}, volume \bibinfo{volume}{3734}, \bibinfo{year}{2023}. \DOIprefix\doi{10.1145/3543507.3583356}.
\bibitem[{Aggarwal et~al.(2021)Aggarwal, Liman, Gold, and Zesch}]{Aggarwal2021VL}
\bibinfo{author}{P.~Aggarwal}, \bibinfo{author}{M.~Liman}, \bibinfo{author}{D.~Gold}, \bibinfo{author}{T.~Zesch},
\newblock \bibinfo{title}{Vl-bert+: Detecting protected groups in hateful multimodal memes},
\newblock in: \bibinfo{booktitle}{WOAH 2021 {-} 5th Workshop on Online Abuse and Harms, Proceedings of the Workshop}, volume \bibinfo{volume}{207}, \bibinfo{year}{2021}.
\bibitem[{Ahmed et~al.(2021)Ahmed, Bhadani, and Chakraborty}]{Ahmed2021Hateful}
\bibinfo{author}{M.~Ahmed}, \bibinfo{author}{N.~Bhadani}, \bibinfo{author}{I.~Chakraborty},
\newblock \bibinfo{title}{Hateful meme prediction model using multimodal deep learning},
\newblock in: \bibinfo{booktitle}{2021 International Conference on Computing, Communication and Green Engineering, CCGE 2021}, volume~\bibinfo{volume}{2}, \bibinfo{year}{2021}. \DOIprefix\doi{10.1109/ccge50943.2021.9776440}.
\bibitem[{Bhat et~al.(2022)Bhat, Varshney, Bajlotra, and Gupta}]{Bhat2022Detection}
\bibinfo{author}{A.~Bhat}, \bibinfo{author}{V.~Varshney}, \bibinfo{author}{V.~Bajlotra}, \bibinfo{author}{V.~Gupta},
\newblock \bibinfo{title}{Detection of hatefulness in memes using unimodal and multimodal techniques},
\newblock in: \bibinfo{booktitle}{Proceedings - 2022 6th International Conference on Intelligent Computing and Control Systems, ICICCS 2022}, volume~\bibinfo{volume}{65}, \bibinfo{year}{2022}. \DOIprefix\doi{10.1109/iciccs53718.2022.9788376}.
\bibitem[{Bhat et~al.(2023)Bhat, Vashisht, Sahni, and Meena}]{Bhat2023Hate}
\bibinfo{author}{A.~Bhat}, \bibinfo{author}{V.~Vashisht}, \bibinfo{author}{V.~Sahni}, \bibinfo{author}{S.~Meena},
\newblock \bibinfo{title}{Hate speech detection using multimodal meme analysis},
\newblock in: \bibinfo{booktitle}{Proceedings of the 2nd International Conference on Applied Artificial Intelligence and Computing, ICAAIC 2023}, volume \bibinfo{volume}{1137}, \bibinfo{year}{2023}. \DOIprefix\doi{10.1109/icaaic56838.2023.10140393}.
\bibitem[{Bi et~al.(2023)Bi, Huang, Han, and Hsu}]{Bi2023You}
\bibinfo{author}{N.~Bi}, \bibinfo{author}{Y.-C. Huang}, \bibinfo{author}{C.-C. Han}, \bibinfo{author}{J.-J. Hsu},
\newblock \bibinfo{title}{You know what i meme: Enhancing people's understanding and awareness of hateful memes using crowdsourced explanations},
\newblock in: \bibinfo{booktitle}{Proceedings of the ACM on Human-Computer Interaction}, volume~\bibinfo{volume}{7}, \bibinfo{year}{2023}. \DOIprefix\doi{10.1145/3579593}.
\bibitem[{Blaier et~al.(2021)Blaier, Malkiel, and Wolf}]{Blaier2021Caption}
\bibinfo{author}{E.~Blaier}, \bibinfo{author}{I.~Malkiel}, \bibinfo{author}{L.~Wolf},
\newblock \bibinfo{title}{Caption enriched samples for improving hateful memes detection},
\newblock in: \bibinfo{booktitle}{EMNLP 2021 - 2021 Conference on Empirical Methods in Natural Language Processing, Proceedings}, volume \bibinfo{volume}{9350}, \bibinfo{year}{2021}.
\bibitem[{Cao et~al.(2022)Cao, Lee, Chong, and Jiang}]{Cao2022Prompting}
\bibinfo{author}{R.~Cao}, \bibinfo{author}{R.-W. Lee}, \bibinfo{author}{W.-H. Chong}, \bibinfo{author}{J.~Jiang},
\newblock \bibinfo{title}{Prompting for multimodal hateful meme classification},
\newblock in: \bibinfo{booktitle}{Proceedings of the 2022 Conference on Empirical Methods in Natural Language Processing, EMNLP 2022}, volume \bibinfo{volume}{321}, \bibinfo{year}{2022}.
\bibitem[{Cao et~al.(2023)Cao, Hee, Kuek, Chong, Lee, and Jiang}]{Cao2023Pro-Cap}
\bibinfo{author}{R.~Cao}, \bibinfo{author}{M.~Hee}, \bibinfo{author}{A.~Kuek}, \bibinfo{author}{W.-H. Chong}, \bibinfo{author}{R.-W. Lee}, \bibinfo{author}{J.~Jiang},
\newblock \bibinfo{title}{Pro-cap: Leveraging a frozen vision-language model for hateful meme detection},
\newblock in: \bibinfo{booktitle}{MM 2023 - Proceedings of the 31st ACM International Conference on Multimedia}, volume \bibinfo{volume}{5244}, \bibinfo{publisher}{Association for Computing Machinery}, \bibinfo{year}{2023}. \DOIprefix\doi{10.1145/3581783.3612498}.
\bibitem[{Chhabra and Vishwakarma(2023)}]{Chhabra2023Multimodal}
\bibinfo{author}{A.~Chhabra}, \bibinfo{author}{D.~Vishwakarma},
\newblock \bibinfo{title}{Multimodal hate speech detection via multi-scale visual kernels and knowledge distillation architecture},
\newblock \bibinfo{journal}{Engineering Applications of Artificial Intelligence} \bibinfo{volume}{126} (\bibinfo{year}{2023}).
\bibitem[{Constantin et~al.(2021)Constantin, Parvu, Stanciu, Ionascu, and Ionescu}]{Constantin2021Hateful}
\bibinfo{author}{M.~Constantin}, \bibinfo{author}{D.-S. Parvu}, \bibinfo{author}{C.~Stanciu}, \bibinfo{author}{D.~Ionascu}, \bibinfo{author}{B.~Ionescu},
\newblock \bibinfo{title}{Hateful meme detection with multimodal deep neural networks},
\newblock in: \bibinfo{booktitle}{ISSCS 2021 - International Symposium on Signals, Circuits and Systems}, volume \bibinfo{volume}{9497374}, \bibinfo{year}{2021}. \DOIprefix\doi{10.1109/isscs52333.2021.9497374}.
\bibitem[{Deshpande and Mani(2021)}]{Deshpande2021interpretable}
\bibinfo{author}{T.~Deshpande}, \bibinfo{author}{N.~Mani},
\newblock \bibinfo{title}{An interpretable approach to hateful meme detection},
\newblock in: \bibinfo{booktitle}{ICMI 2021 - Proceedings of the 2021 International Conference on Multimodal Interaction}, volume \bibinfo{volume}{723}, \bibinfo{year}{2021}. \DOIprefix\doi{10.1145/3462244.3479949}.
\bibitem[{Fang et~al.(2022)Fang, Zhu, Han, and Hu}]{Fang2022Multimodal}
\bibinfo{author}{H.~Fang}, \bibinfo{author}{F.~Zhu}, \bibinfo{author}{J.~Han}, \bibinfo{author}{S.~Hu},
\newblock \bibinfo{title}{Multimodal hateful memes detection via image caption supervision},
\newblock in: \bibinfo{booktitle}{Proceedings - 2022 IEEE SmartWorld, Ubiquitous Intelligence and Computing, Autonomous and Trusted Vehicles, Scalable Computing and Communications, Digital Twin, Privacy Computing, Metaverse, SmartWorld/UIC/ATC/ScalCom/DigitalTwin/PriComp/Metaverse 2022}, volume \bibinfo{volume}{1530}, \bibinfo{year}{2022}. \DOIprefix\doi{10.1109/SmartWorld-UIC-ATC-ScalCom-DigitalTwin-PriComp-Metaverse56740.2022.00221}.
\bibitem[{Gaikwad et~al.(2022)Gaikwad, Kurma, Patwardhan, Karande, and Pedanekar}]{Gaikwad2022Can}
\bibinfo{author}{B.~Gaikwad}, \bibinfo{author}{B.~Kurma}, \bibinfo{author}{M.~Patwardhan}, \bibinfo{author}{S.~Karande}, \bibinfo{author}{N.~Pedanekar},
\newblock \bibinfo{title}{Can a pretrained language model make sense with pretrained neural extractors? an application to multimodal classification},
\newblock in: \bibinfo{booktitle}{CEUR Workshop Proceedings}, volume \bibinfo{volume}{3168}, \bibinfo{year}{2022}. \DOIprefix\doi{10.1109/iceic57457.2023.10049865}.
\bibitem[{Goswami et~al.(2023)Goswami, Rawat, Tongaria, and Jhingran}]{Goswami2023Detection}
\bibinfo{author}{A.~Goswami}, \bibinfo{author}{A.~Rawat}, \bibinfo{author}{S.~Tongaria}, \bibinfo{author}{S.~Jhingran},
\newblock \bibinfo{title}{Detection of hate speech in multi-modal social post},
\newblock in: \bibinfo{booktitle}{Artificial Intelligence, Blockchain, Computing and Security: Volume 1}, volume~\bibinfo{volume}{1}, \bibinfo{year}{2023}. \DOIprefix\doi{10.1201/9781003393580-50}.
\bibitem[{Hee et~al.(2022)Hee, Lee, and Chong}]{Hee2022On}
\bibinfo{author}{M.~Hee}, \bibinfo{author}{R.-W. Lee}, \bibinfo{author}{W.-H. Chong},
\newblock \bibinfo{title}{On explaining multimodal hateful meme detection models},
\newblock in: \bibinfo{booktitle}{WWW 2022 - Proceedings of the ACM Web Conference 2022}, volume \bibinfo{volume}{3651}, \bibinfo{year}{2022}. \DOIprefix\doi{10.1145/3485447.3512260}.
\bibitem[{Kiran et~al.(2023)Kiran, Shetty, Shukla, Kerenalli, and Das}]{Kiran2023Getting}
\bibinfo{author}{A.~Kiran}, \bibinfo{author}{M.~Shetty}, \bibinfo{author}{S.~Shukla}, \bibinfo{author}{V.~Kerenalli}, \bibinfo{author}{B.~Das},
\newblock \bibinfo{title}{Getting around the semantics challenge in hateful memes},
\newblock in: \bibinfo{booktitle}{Lecture Notes on Data Engineering and Communications Technologies}, volume \bibinfo{volume}{142}, \bibinfo{year}{2023}. \DOIprefix\doi{10.1007/978-981-19-3391-2\_26}.
\bibitem[{Kougia and Pavlopoulos(2021)}]{Kougia2021Multimodal}
\bibinfo{author}{V.~Kougia}, \bibinfo{author}{J.~Pavlopoulos},
\newblock \bibinfo{title}{Multimodal or text? retrieval or bert? benchmarking classifiers for the shared task on hateful memes},
\newblock in: \bibinfo{booktitle}{WOAH 2021 - 5th Workshop on Online Abuse and Harms, Proceedings of the Workshop}, volume \bibinfo{volume}{220}, \bibinfo{year}{2021}.
\bibitem[{Kougia et~al.(2023)Kougia, Fetzel, Kirchmair, \c{C}ano, Baharlou, Sharifzadeh, and Roth}]{kougia2023memegraphs}
\bibinfo{author}{V.~Kougia}, \bibinfo{author}{S.~Fetzel}, \bibinfo{author}{T.~Kirchmair}, \bibinfo{author}{E.~\c{C}ano}, \bibinfo{author}{S.~Baharlou}, \bibinfo{author}{S.~Sharifzadeh}, \bibinfo{author}{B.~Roth},
\newblock \bibinfo{title}{Memegraphs: Linking memes to knowledge graphs},
\newblock in: \bibinfo{booktitle}{Lecture Notes in Computer Science}, volume \bibinfo{volume}{14187} of \textit{\bibinfo{series}{Lecture Notes in Computer Science (including subseries Lecture Notes in Artificial Intelligence and Lecture Notes in Bioinformatics)}}, \bibinfo{organization}{Springer}, \bibinfo{year}{2023}, pp. \bibinfo{pages}{534--551}. \DOIprefix\doi{10.1007/978-3-031-41676-7\_31}.
\bibitem[{Lee et~al.(2021)Lee, Cao, Fan, Jiang, and Chong}]{Lee2021Disentangling}
\bibinfo{author}{R.~K.-W. Lee}, \bibinfo{author}{R.~Cao}, \bibinfo{author}{Z.~Fan}, \bibinfo{author}{J.~Jiang}, \bibinfo{author}{W.-H. Chong},
\newblock \bibinfo{title}{Disentangling hate in online memes},
\newblock in: \bibinfo{booktitle}{Proceedings of the 29th ACM International Conference on Multimedia}, volume \bibinfo{volume}{5138}, \bibinfo{organization}{Association for Computing Machinery}, \bibinfo{year}{2021}, pp. \bibinfo{pages}{5138--5147}. \DOIprefix\doi{10.1145/3474085.3475625}.
\bibitem[{Liang et~al.(2022)Liang, Huang, Liu, Zhu, Liang, and Chen}]{Liang2022TRICAN}
\bibinfo{author}{X.~Liang}, \bibinfo{author}{Y.-C. Huang}, \bibinfo{author}{W.~Liu}, \bibinfo{author}{H.~Zhu}, \bibinfo{author}{Z.~Liang}, \bibinfo{author}{L.~Chen},
\newblock \bibinfo{title}{Trican: Multi-modal hateful memes detection with triplet-relation information cross-attention network},
\newblock in: \bibinfo{booktitle}{Proceedings of the International Joint Conference on Neural Networks}, volume \bibinfo{volume}{2022-July}, \bibinfo{year}{2022}. \DOIprefix\doi{10.1109/ijcnn55064.2022.9892164}.
\bibitem[{Ma et~al.(2022)Ma, Yao, Wu, Gao, and Zhang}]{Ma2022Hateful}
\bibinfo{author}{Z.~Ma}, \bibinfo{author}{S.~Yao}, \bibinfo{author}{L.~Wu}, \bibinfo{author}{S.~Gao}, \bibinfo{author}{Y.~Zhang},
\newblock \bibinfo{title}{Hateful memes detection based on multi-task learning},
\newblock \bibinfo{journal}{Mathematics} \bibinfo{volume}{10} (\bibinfo{year}{2022}).
\bibitem[{MacRayo et~al.(2023)MacRayo, Casino, Dalangin, Gabriel~Gahoy, Christian~Reyes, Vitto, Abisado, Lor Huyo-A, and Avelino~Sampedro}]{macrayo2023please}
\bibinfo{author}{G.~MacRayo}, \bibinfo{author}{W.~Casino}, \bibinfo{author}{J.~Dalangin}, \bibinfo{author}{J.~Gabriel~Gahoy}, \bibinfo{author}{A.~Christian~Reyes}, \bibinfo{author}{C.~Vitto}, \bibinfo{author}{M.~Abisado}, \bibinfo{author}{S.~Lor Huyo-A}, \bibinfo{author}{G.~Avelino~Sampedro},
\newblock \bibinfo{title}{Please be nice: A deep learning based approach to content moderation of internet memes},
\newblock in: \bibinfo{booktitle}{2023 International Conference on Electronics, Information, and Communication, ICEIC 2023}, volume~\bibinfo{volume}{0}, \bibinfo{organization}{Ieee}, \bibinfo{year}{2023}, pp. \bibinfo{pages}{1--5}. \DOIprefix\doi{10.1109/iceic57457.2023.10049865}.
\bibitem[{Mookdarsanit and Mookdarsanit(2021)}]{Mookdarsanit2021Combating}
\bibinfo{author}{L.~Mookdarsanit}, \bibinfo{author}{P.~Mookdarsanit},
\newblock \bibinfo{title}{Combating the hate speech in thai textual memes},
\newblock \bibinfo{journal}{Indonesian Journal of Electrical Engineering and Computer Science} \bibinfo{volume}{21} (\bibinfo{year}{2021}) \bibinfo{pages}{1493--1502}.
\bibitem[{Nayak and Agrawal(2022)}]{Nayak2022Detection}
\bibinfo{author}{A.~Nayak}, \bibinfo{author}{A.~Agrawal},
\newblock \bibinfo{title}{Detection of hate speech in social media memes: A comparative analysis},
\newblock in: \bibinfo{booktitle}{Proceedings of the 2022 3rd International Conference on Intelligent Computing, Instrumentation and Control Technologies: Computational Intelligence for Smart Systems, ICICICT 2022}, volume \bibinfo{volume}{1179}, \bibinfo{year}{2022}. \DOIprefix\doi{10.1109/icicict54557.2022.9917633}.
\bibitem[{Qu et~al.(2023{\natexlab{a}})Qu, He, Pierson, Backes, Zhang, and Zannettou}]{Qu2023On}
\bibinfo{author}{Y.~Qu}, \bibinfo{author}{X.~He}, \bibinfo{author}{S.~Pierson}, \bibinfo{author}{M.~Backes}, \bibinfo{author}{Y.~Zhang}, \bibinfo{author}{S.~Zannettou},
\newblock \bibinfo{title}{On the evolution of (hateful) memes by means of multimodal contrastive learning},
\newblock in: \bibinfo{booktitle}{Proceedings - IEEE Symposium on Security and Privacy}, volume \bibinfo{volume}{2023-May}, \bibinfo{year}{2023}{\natexlab{a}}. \DOIprefix\doi{10.1109/sp46215.2023.10179315}.
\bibitem[{Qu et~al.(2023{\natexlab{b}})Qu, Shen, He, Backes, Zannettou, and Zhang}]{qu2023unsafe}
\bibinfo{author}{Y.~Qu}, \bibinfo{author}{X.~Shen}, \bibinfo{author}{X.~He}, \bibinfo{author}{M.~Backes}, \bibinfo{author}{S.~Zannettou}, \bibinfo{author}{Y.~Zhang},
\newblock \bibinfo{title}{Unsafe diffusion: On the generation of unsafe images and hateful memes from text-to-image models},
\newblock in: \bibinfo{booktitle}{Proceedings of the 2023 ACM SIGSAC Conference on Computer and Communications Security}, Ccs '23, \bibinfo{publisher}{Association for Computing Machinery}, \bibinfo{year}{2023}{\natexlab{b}}, p. \bibinfo{pages}{3403–3417}. \DOIprefix\doi{10.1145/3576915.3616679}.
\bibitem[{Sethi et~al.(2021)Sethi, Kuchhal, Anjum, and Katarya}]{Sethi2021Study}
\bibinfo{author}{A.~Sethi}, \bibinfo{author}{U.~Kuchhal}, \bibinfo{author}{Anjum}, \bibinfo{author}{R.~Katarya},
\newblock \bibinfo{title}{Study of various techniques for the classification of hateful memes},
\newblock in: \bibinfo{booktitle}{2021 6th International Conference on Recent Trends on Electronics, Information, Communication and Technology, RTEICT 2021}, volume \bibinfo{volume}{675}, \bibinfo{year}{2021}. \DOIprefix\doi{10.1109/rteict52294.2021.9573926}.
\bibitem[{Wanbo and Suying(2021)}]{Wanbo2021Research}
\bibinfo{author}{L.~Wanbo}, \bibinfo{author}{L.~Suying},
\newblock \bibinfo{title}{Research on multi-modal hateful meme detection},
\newblock in: \bibinfo{booktitle}{ACM International Conference Proceeding Series}, volume \bibinfo{volume}{3470385}, \bibinfo{year}{2021}. \DOIprefix\doi{10.1145/3469213.3470385}.
\bibitem[{Wu and Mebane(2022)}]{Wu2022MARMOT}
\bibinfo{author}{P.~Wu}, \bibinfo{author}{W.~Mebane},
\newblock \bibinfo{title}{Marmot a deep learning framework for constructing multimodal representations for vision-and-language tasks},
\newblock \bibinfo{journal}{Computational Communication Research} \bibinfo{volume}{4} (\bibinfo{year}{2022}).
\bibitem[{Zhang et~al.(2023)Zhang, Jin, Sun, Xu, Zhang, Li, Liu, Liu, and Yan}]{Zhang2023TOT}
\bibinfo{author}{L.~Zhang}, \bibinfo{author}{L.~Jin}, \bibinfo{author}{X.~Sun}, \bibinfo{author}{G.~Xu}, \bibinfo{author}{Z.~Zhang}, \bibinfo{author}{X.~Li}, \bibinfo{author}{N.~Liu}, \bibinfo{author}{Q.~Liu}, \bibinfo{author}{S.~Yan},
\newblock \bibinfo{title}{Tot: Topology-aware optimal transport for multimodal hate detection},
\newblock in: \bibinfo{booktitle}{Proceedings of the 37th AAAI Conference on Artificial Intelligence, AAAI 2023}, volume~\bibinfo{volume}{37}, \bibinfo{year}{2023}.
\bibitem[{Zhou et~al.(2021)Zhou, Chen, and Yang}]{Zhou2021Multimodal}
\bibinfo{author}{Y.~Zhou}, \bibinfo{author}{Z.~Chen}, \bibinfo{author}{H.~Yang},
\newblock \bibinfo{title}{Multimodal learning for hateful memes detection},
\newblock \bibinfo{journal}{2021 IEEE International Conference on Multimedia and Expo Workshops, ICMEW 2021} \bibinfo{volume}{28} (\bibinfo{year}{2021}).
\bibitem[{Zhu et~al.(2022)Zhu, Lee, and Chong}]{Zhu2022Multimodal}
\bibinfo{author}{J.~Zhu}, \bibinfo{author}{R.-W. Lee}, \bibinfo{author}{W.~Chong},
\newblock \bibinfo{title}{Multimodal zero-shot hateful meme detection},
\newblock in: \bibinfo{booktitle}{ACM International Conference Proceeding Series}, volume \bibinfo{volume}{382}, \bibinfo{year}{2022}. \DOIprefix\doi{10.1145/3501247.3531557}.
\bibitem[{Aggarwal et~al.(2024)Aggarwal, Mehrabanian, Huang, Alacam, and Zesch}]{aggarwal2024text}
\bibinfo{author}{P.~Aggarwal}, \bibinfo{author}{J.~Mehrabanian}, \bibinfo{author}{W.~Huang}, \bibinfo{author}{O.~Alacam}, \bibinfo{author}{T.~Zesch},
\newblock \bibinfo{title}{Text or image? what is more important in cross-domain generalization capabilities of hate meme detection models?},
\newblock \bibinfo{journal}{arXiv}  (\bibinfo{year}{2024}).
\bibitem[{Chen and Pan(2022)}]{chen2022multimodal}
\bibinfo{author}{Y.~Chen}, \bibinfo{author}{F.~Pan},
\newblock \bibinfo{title}{Multimodal detection of hateful messages using visual-linguistic pre-trained deep learning models},
\newblock \bibinfo{journal}{Research Square}  (\bibinfo{year}{2022}).
\bibitem[{Das et~al.(2020)Das, Wahi, and Li}]{da2020detecting}
\bibinfo{author}{A.~Das}, \bibinfo{author}{J.~S. Wahi}, \bibinfo{author}{S.~Li},
\newblock \bibinfo{title}{Detecting hate speech in multi-modal memes},
\newblock \bibinfo{journal}{arXiv}  (\bibinfo{year}{2020}).
\bibitem[{Evtimov et~al.(2020)Evtimov, Howes, Dolhansky, Firooz, and Ferrer}]{evtimov2020adversarial}
\bibinfo{author}{I.~Evtimov}, \bibinfo{author}{R.~Howes}, \bibinfo{author}{B.~Dolhansky}, \bibinfo{author}{H.~Firooz}, \bibinfo{author}{C.~C. Ferrer},
\newblock \bibinfo{title}{Adversarial evaluation of multimodal models under realistic gray box assumptions},
\newblock \bibinfo{journal}{arXiv}  (\bibinfo{year}{2020}).
\bibitem[{Gao et~al.(2021)Gao, Wang, Yin, and Tian}]{Gao2021hateful}
\bibinfo{author}{A.~Gao}, \bibinfo{author}{B.~Wang}, \bibinfo{author}{J.~Yin}, \bibinfo{author}{Y.~Tian},
\newblock \bibinfo{title}{Hateful memes challenge: An enhanced multimodal framework},
\newblock \bibinfo{journal}{arXiv}  (\bibinfo{year}{2021}).
\bibitem[{Jennifer et~al.(2022)Jennifer, Tahmasbi, Blackburn, Zannettou, and De~Cristofaro}]{jennifer2022feels}
\bibinfo{author}{C.~Jennifer}, \bibinfo{author}{F.~Tahmasbi}, \bibinfo{author}{J.~Blackburn}, \bibinfo{author}{S.~Zannettou}, \bibinfo{author}{E.~De~Cristofaro},
\newblock \bibinfo{title}{Feels bad man: Dissecting automated hateful meme detection through the lens of facebook's challenge},
\newblock \bibinfo{journal}{arXiv}  (\bibinfo{year}{2022}).
\bibitem[{Jin and Wilhelm(2022)}]{jin2022hateful}
\bibinfo{author}{W.~Jin}, \bibinfo{author}{L.~Wilhelm},
\newblock \bibinfo{title}{The hateful memes challenge next move},
\newblock \bibinfo{journal}{arXiv}  (\bibinfo{year}{2022}).
\bibitem[{Li et~al.(2021)Li, Zhang, and Huang}]{li2021enhance}
\bibinfo{author}{Y.~Li}, \bibinfo{author}{Z.~Zhang}, \bibinfo{author}{H.~Huang},
\newblock \bibinfo{title}{Enhance multimodal model performance with data augmentation: facebook hateful meme challenge solution},
\newblock \bibinfo{journal}{arXiv}  (\bibinfo{year}{2021}).
\bibitem[{Lippe et~al.(2020)Lippe, Holla, Chandra, Rajamanickam, Antoniou, Shutova, and Yannakoudakis}]{Lippe2020multimodal}
\bibinfo{author}{P.~Lippe}, \bibinfo{author}{N.~Holla}, \bibinfo{author}{S.~Chandra}, \bibinfo{author}{S.~Rajamanickam}, \bibinfo{author}{G.~Antoniou}, \bibinfo{author}{E.~Shutova}, \bibinfo{author}{H.~Yannakoudakis},
\newblock \bibinfo{title}{A multimodal framework for the detection of hateful memes},
\newblock \bibinfo{journal}{arXiv preprint arXiv:2012.12871}  (\bibinfo{year}{2020}).
\bibitem[{Mei et~al.(2023)Mei, Chen, Lin, Byrne, and Tomalin}]{mei2023improving}
\bibinfo{author}{J.~Mei}, \bibinfo{author}{J.~Chen}, \bibinfo{author}{W.~Lin}, \bibinfo{author}{B.~Byrne}, \bibinfo{author}{M.~Tomalin},
\newblock \bibinfo{title}{Improving hateful memes detection via learning hatefulness-aware embedding space through retrieval-guided contrastive learning},
\newblock \bibinfo{journal}{arXiv}  (\bibinfo{year}{2023}).
\bibitem[{Miyanishi and Le~Nguyen(2023)}]{miyanishi2023causal}
\bibinfo{author}{Y.~Miyanishi}, \bibinfo{author}{M.~Le~Nguyen},
\newblock \bibinfo{title}{Causal intersectionality and dual form of gradient descent for multimodal analysis: a case study on hateful memes},
\newblock \bibinfo{journal}{arXiv}  (\bibinfo{year}{2023}).
\bibitem[{Muennighoff(2020)}]{Muennighoff2020vilio}
\bibinfo{author}{N.~Muennighoff},
\newblock \bibinfo{title}{Vilio: State-of-the-art visio-linguistic models applied to hateful memes},
\newblock \bibinfo{journal}{arXiv preprint arXiv:2012.07788}  (\bibinfo{year}{2020}).
\bibitem[{Sandulescu(2020)}]{sandulescu2020detecting}
\bibinfo{author}{V.~Sandulescu},
\newblock \bibinfo{title}{Detecting hateful memes using a multimodal deep ensemble},
\newblock \bibinfo{journal}{arXiv preprint arXiv:2012.13235}  (\bibinfo{year}{2020}).
\bibitem[{Van and Wu(2023)}]{van2023detecting}
\bibinfo{author}{M.-H. Van}, \bibinfo{author}{X.~Wu},
\newblock \bibinfo{title}{Detecting and correcting hate speech in multimodal memes with large visual language model},
\newblock \bibinfo{journal}{arXiv}  (\bibinfo{year}{2023}).
\bibitem[{Velioglu and Rose(2020)}]{Velioglu2020detecting}
\bibinfo{author}{R.~Velioglu}, \bibinfo{author}{J.~Rose},
\newblock \bibinfo{title}{Detecting hate speech in memes using multimodal deep learning approaches: Prize-winning solution to hateful memes challenge},
\newblock \bibinfo{journal}{arXiv preprint arXiv:2012.12975}  (\bibinfo{year}{2020}).
\bibitem[{Yuan et~al.(2023)Yuan, Yu, Mittal, Sajeev, and Chen}]{Yuan2023rethinking}
\bibinfo{author}{J.~Yuan}, \bibinfo{author}{Y.~Yu}, \bibinfo{author}{G.~Mittal}, \bibinfo{author}{S.~Sajeev}, \bibinfo{author}{M.~Chen},
\newblock \bibinfo{title}{Rethinking multimodal content moderation from an asymmetric angle with mixed-modality},
\newblock \bibinfo{journal}{arXiv}  (\bibinfo{year}{2023}).
\bibitem[{Zhang et~al.(2020)Zhang, Liu, Li, and Zhu}]{zhang2020hateful}
\bibinfo{author}{W.~Zhang}, \bibinfo{author}{G.~Liu}, \bibinfo{author}{Z.~Li}, \bibinfo{author}{F.~Zhu},
\newblock \bibinfo{title}{Hateful memes detection via complementary visual and linguistic networks},
\newblock \bibinfo{journal}{arXiv}  (\bibinfo{year}{2020}).
\bibitem[{Zhao et~al.(2023)Zhao, Zhang, Watson, Kearney, and Dale}]{zhao2023review}
\bibinfo{author}{B.~Zhao}, \bibinfo{author}{A.~Zhang}, \bibinfo{author}{B.~Watson}, \bibinfo{author}{G.~Kearney}, \bibinfo{author}{I.~Dale},
\newblock \bibinfo{title}{A review of vision-language models and their performance on the hateful memes challenge},
\newblock \bibinfo{journal}{arXiv}  (\bibinfo{year}{2023}).
\bibitem[{Zhong(2020)}]{zhong2020classification}
\bibinfo{author}{X.~Zhong},
\newblock \bibinfo{title}{Classification of multimodal hate speech -the winning solution of hateful memes challenge},
\newblock \bibinfo{journal}{arXiv}  (\bibinfo{year}{2020}).
\bibitem[{Zhou and Chen(2020)}]{zhou2020multimodal}
\bibinfo{author}{Y.~Zhou}, \bibinfo{author}{Z.~Chen},
\newblock \bibinfo{title}{Multimodal learning for hateful memes detection},
\newblock \bibinfo{journal}{arXiv}  (\bibinfo{year}{2020}).
\bibitem[{Zhu(2020)}]{zhu2020enhance}
\bibinfo{author}{R.~Zhu},
\newblock \bibinfo{title}{Enhance multimodal transformer with external label and in-domain pretrain: Hateful meme challenge winning solution},
\newblock \bibinfo{journal}{arXiv preprint arXiv:2012.08290}  (\bibinfo{year}{2020}).
\bibitem[{Burbi et~al.(2023)Burbi, Baldrati, Agnolucci, Bertini, and Del~Bimbo}]{burbi2023mapping}
\bibinfo{author}{G.~Burbi}, \bibinfo{author}{A.~Baldrati}, \bibinfo{author}{L.~Agnolucci}, \bibinfo{author}{M.~Bertini}, \bibinfo{author}{A.~Del~Bimbo},
\newblock \bibinfo{title}{Mapping memes to words for multimodal hateful meme classification},
\newblock in: \bibinfo{booktitle}{Proceedings of the IEEE/CVF International Conference on Computer Vision}, \bibinfo{year}{2023}, pp. \bibinfo{pages}{2832--2836}. \DOIprefix\doi{10.1109/iccvw60793.2023.00303}.
\bibitem[{Fersini et~al.(2019)Fersini, Gasparini, and Corchs}]{fersini2019detecting}
\bibinfo{author}{E.~Fersini}, \bibinfo{author}{F.~Gasparini}, \bibinfo{author}{S.~Corchs},
\newblock \bibinfo{title}{Detecting sexist meme on the web: A study on textual and visual cues},
\newblock in: \bibinfo{booktitle}{2019 8th International Conference on Affective Computing and Intelligent Interaction Workshops and Demos (ACIIW)}, \bibinfo{organization}{Ieee}, \bibinfo{year}{2019}, pp. \bibinfo{pages}{226--231}.
\bibitem[{Attanasio et~al.(2022)Attanasio, Nozza, and Bianchi}]{attanasio2022milanlp}
\bibinfo{author}{G.~Attanasio}, \bibinfo{author}{D.~Nozza}, \bibinfo{author}{F.~Bianchi},
\newblock \bibinfo{title}{Milanlp at semeval-2022 task 5: Using perceiver io for detecting misogynous memes with text and image modalities},
\newblock in: \bibinfo{booktitle}{Proceedings of SemEval 2022 - 16th International Workshop on Semantic Evaluation}, \bibinfo{year}{2022}, pp. \bibinfo{pages}{654--662}.
\bibitem[{Behzadi et~al.(2022)Behzadi, Derakhshan, and Harris}]{behzadi2022mitra}
\bibinfo{author}{M.~Behzadi}, \bibinfo{author}{A.~Derakhshan}, \bibinfo{author}{I.~Harris},
\newblock \bibinfo{title}{Mitra behzadi at semeval-2022 task 5: Multimedia automatic misogyny identification method based on clip},
\newblock in: \bibinfo{booktitle}{Proceedings of SemEval 2022 - 16th International Workshop on Semantic Evaluation}, \bibinfo{year}{2022}, pp. \bibinfo{pages}{724--727}.
\bibitem[{Chen and Chou(2022)}]{chen2022rit}
\bibinfo{author}{L.~Chen}, \bibinfo{author}{H.~Chou},
\newblock \bibinfo{title}{Rit boston at semeval-2022 task 5: Multimedia misogyny detection by using coherent visual and language features from clip model and data-centric ai principle},
\newblock in: \bibinfo{booktitle}{Proceedings of SemEval 2022 - 16th Intl. Workshop on Semantic Evaluation}, \bibinfo{year}{2022}, pp. \bibinfo{pages}{636--641}.
\bibitem[{Gu et~al.(2022)Gu, Castro, and Tyson}]{gu2022mmvae}
\bibinfo{author}{Y.~Gu}, \bibinfo{author}{I.~Castro}, \bibinfo{author}{G.~Tyson},
\newblock \bibinfo{title}{Mmvae at semeval-2022 task 5: A multi-modal multi-task vae on misogynous meme detection},
\newblock in: \bibinfo{booktitle}{Proceedings of SemEval 2022 - 16th International Workshop on Semantic Evaluation}, \bibinfo{year}{2022}, pp. \bibinfo{pages}{700--710}.
\bibitem[{Kalkenings and Mandl(2022)}]{kalkenings2022university}
\bibinfo{author}{M.~Kalkenings}, \bibinfo{author}{T.~Mandl},
\newblock \bibinfo{title}{University of hildesheim at semeval-2022 task 5: Combining deep text and image models for multimedia misogyny detection},
\newblock in: \bibinfo{booktitle}{Proceedings of SemEval 2022 - 16th International Workshop on Semantic Evaluation}, \bibinfo{year}{2022}, pp. \bibinfo{pages}{718--723}.
\bibitem[{Obeidat et~al.(2023)Obeidat, Nammas, and Abdullah}]{obeidat2023just_one}
\bibinfo{author}{D.~Obeidat}, \bibinfo{author}{H.~Nammas}, \bibinfo{author}{M.~Abdullah},
\newblock \bibinfo{title}{Just\_one at semeval-2023 task 10: Explainable detection of online sexism (edos)},
\newblock in: \bibinfo{booktitle}{Proceedings of the 17th International Workshop on Semantic Evaluation (SemEval-2023)}, \bibinfo{year}{2023}, pp. \bibinfo{pages}{526--531}.
\bibitem[{Paraschiv et~al.(2022)Paraschiv, Dascalu, and Cercel}]{paraschiv2022upb}
\bibinfo{author}{A.~Paraschiv}, \bibinfo{author}{M.~Dascalu}, \bibinfo{author}{D.-C. Cercel},
\newblock \bibinfo{title}{Upb at semeval-2022 task 5: Enhancing uniter with image sentiment and graph convolutional networks for multimedia automatic misogyny identification},
\newblock in: \bibinfo{booktitle}{Proceedings of SemEval 2022 - 16th International Workshop on Semantic Evaluation}, \bibinfo{year}{2022}, pp. \bibinfo{pages}{618--625}.
\bibitem[{Ravagli and Vaiani(2022)}]{ravagli2022jrlv}
\bibinfo{author}{J.~Ravagli}, \bibinfo{author}{L.~Vaiani},
\newblock \bibinfo{title}{Jrlv at semeval-2022 task 5: The importance of visual elements for misogyny identification in memes},
\newblock in: \bibinfo{booktitle}{SemEval 2022 - 16th International Workshop on Semantic Evaluation, Proceedings of the Workshop}, \bibinfo{year}{2022}, pp. \bibinfo{pages}{610--617}.
\bibitem[{Rizzi et~al.(2023)Rizzi, Gasparini, Saibene, Rosso, and Fersini}]{rizzi2023recognizing}
\bibinfo{author}{G.~Rizzi}, \bibinfo{author}{F.~Gasparini}, \bibinfo{author}{A.~Saibene}, \bibinfo{author}{P.~Rosso}, \bibinfo{author}{E.~Fersini},
\newblock \bibinfo{title}{Recognizing misogynous memes: Biased models and tricky archetypes},
\newblock \bibinfo{journal}{Information Processing and Management} \bibinfo{volume}{60} (\bibinfo{year}{2023}) \bibinfo{pages}{103474}.
\bibitem[{Singh et~al.(2023)Singh, Haridasan, and Mooney}]{singh2023female}
\bibinfo{author}{S.~Singh}, \bibinfo{author}{A.~Haridasan}, \bibinfo{author}{R.~Mooney},
\newblock \bibinfo{title}{``female astronaut: Because sandwiches won't make themselves up there!": Towards multi-modal misogyny detection in memes},
\newblock in: \bibinfo{booktitle}{Proceedings of the Annual Meeting of the Association for Computational Linguistics}, \bibinfo{year}{2023}, pp. \bibinfo{pages}{150--159}.
\bibitem[{Tung et~al.(2023)Tung, Viet, Anh, and Hung}]{tung2023semimemes}
\bibinfo{author}{P.~Tung}, \bibinfo{author}{N.~Viet}, \bibinfo{author}{N.~Anh}, \bibinfo{author}{P.~Hung},
\newblock \bibinfo{title}{Semimemes: A semi-supervised learning approach for multimodal memes analysis},
\newblock in: \bibinfo{booktitle}{Lecture Notes in Computer Science}, volume \bibinfo{volume}{14162} of \textit{\bibinfo{series}{Lecture Notes in Computer Science}}, \bibinfo{year}{2023}, pp. \bibinfo{pages}{565--577}. \DOIprefix\doi{10.1007/978-3-031-41456-5\_43}.
\bibitem[{Singh et~al.(2023)Singh, Das, Manderna, and Chand}]{singh2023devi}
\bibinfo{author}{N.~K. Singh}, \bibinfo{author}{P.~Das}, \bibinfo{author}{A.~Manderna}, \bibinfo{author}{S.~Chand},
\newblock \bibinfo{title}{Devi deep learning framework for misogyny identification in multimodal data},
\newblock \bibinfo{journal}{Research Square}  (\bibinfo{year}{2023}).
\bibitem[{Drakett et~al.(2018)Drakett, Rickett, Day, and Milnes}]{drakett2018old}
\bibinfo{author}{J.~Drakett}, \bibinfo{author}{B.~Rickett}, \bibinfo{author}{K.~Day}, \bibinfo{author}{K.~Milnes},
\newblock \bibinfo{title}{Old jokes, new media--online sexism and constructions of gender in internet memes},
\newblock \bibinfo{journal}{Feminism \& psychology} \bibinfo{volume}{28} (\bibinfo{year}{2018}) \bibinfo{pages}{109--127}.
\bibitem[{Alzu'bi et~al.(2023)Alzu'bi, Bani~Younis, Abuarqoub, and Hammoudeh}]{alzubi2023multimodal}
\bibinfo{author}{A.~Alzu'bi}, \bibinfo{author}{L.~Bani~Younis}, \bibinfo{author}{A.~Abuarqoub}, \bibinfo{author}{M.~Hammoudeh},
\newblock \bibinfo{title}{Multimodal deep learning with discriminant descriptors for offensive memes detection},
\newblock in: \bibinfo{booktitle}{Journal of Data and Information Quality}, volume~\bibinfo{volume}{15}, \bibinfo{year}{2023}, p. \bibinfo{pages}{3597308}. \DOIprefix\doi{10.1145/3597308}.
\bibitem[{Aman et~al.(2021)Aman, Krishna, Anand, and Lal}]{aman2021identification}
\bibinfo{author}{A.~Aman}, \bibinfo{author}{G.~Krishna}, \bibinfo{author}{T.~Anand}, \bibinfo{author}{A.~Lal},
\newblock \bibinfo{title}{Identification of offensive content in memes},
\newblock in: \bibinfo{booktitle}{Lecture Notes in Networks and Systems}, volume \bibinfo{volume}{290}, \bibinfo{year}{2021}, pp. \bibinfo{pages}{438--445}. \DOIprefix\doi{10.1007/978-981-16-4486-3\_49}.
\bibitem[{Baruah et~al.(2020)Baruah, Das, Barbhuiya, and Dey}]{baruah2020iiitg}
\bibinfo{author}{A.~Baruah}, \bibinfo{author}{K.~Das}, \bibinfo{author}{F.~Barbhuiya}, \bibinfo{author}{K.~Dey},
\newblock \bibinfo{title}{Iiitg-adbu at semeval-2020 task 8: A multimodal approach to detect offensive, sarcastic and humorous memes},
\newblock in: \bibinfo{booktitle}{Proceedings of SemEval 2020 - 14th International Workshops on Semantic Evaluation}, \bibinfo{year}{2020}, pp. \bibinfo{pages}{885--890}.
\bibitem[{Bejan(2020)}]{bejan2020memosys}
\bibinfo{author}{I.~Bejan},
\newblock \bibinfo{title}{Memosys at semeval-2020 task 8: Multimodal emotion analysis in memes},
\newblock in: \bibinfo{booktitle}{Proceedings of SemEval 2020 -- 14th International Workshops on Semantic Evaluation}, \bibinfo{year}{2020}, pp. \bibinfo{pages}{1172--1178}.
\bibitem[{Boinepelli et~al.(2020)Boinepelli, Shrivastava, and Varma}]{boinepelli2020sis}
\bibinfo{author}{S.~Boinepelli}, \bibinfo{author}{M.~Shrivastava}, \bibinfo{author}{V.~Varma},
\newblock \bibinfo{title}{Sis\@iiith at semeval-2020 task 8: An overview of simple text classification methods for meme analysis},
\newblock in: \bibinfo{booktitle}{Proceedings of SemEval 2020 -- 14th International Workshops on Semantic Evaluation}, \bibinfo{year}{2020}, pp. \bibinfo{pages}{1190--1194}.
\bibitem[{Bucur et~al.(2022)Bucur, Cosma, and Iordache}]{bucur2022blue}
\bibinfo{author}{A.-M. Bucur}, \bibinfo{author}{A.~Cosma}, \bibinfo{author}{I.-B. Iordache},
\newblock \bibinfo{title}{Blue at memotion 2.0 2022: You have my image, my text and my transformer},
\newblock in: \bibinfo{booktitle}{CEUR Workshop Proceedings}, volume \bibinfo{volume}{3168}, \bibinfo{year}{2022}.
\bibitem[{Sharma et~al.(2022)Sharma, Kushwaha, Jaiswal, and Nandi}]{cui2022meme}
\bibinfo{author}{V.~Sharma}, \bibinfo{author}{V.~Kushwaha}, \bibinfo{author}{S.~Jaiswal}, \bibinfo{author}{G.~Nandi},
\newblock \bibinfo{title}{Meme detection for sentiment analysis and human robot interactions using multiple modes},
\newblock in: \bibinfo{booktitle}{9th IEEE Uttar Pradesh Section International Conference on Electrical, Electronics and Computer Engineering, UPCON 2022}, \bibinfo{year}{2022}. \DOIprefix\doi{10.1109/upcon56432.2022.9986453}.
\bibitem[{de~la Pe\~{n}a Sarrac\'{e}n et~al.(2020)de~la Pe\~{n}a Sarrac\'{e}n, Rosso, and Giachanou}]{dela2020prhlt}
\bibinfo{author}{G.~de~la Pe\~{n}a Sarrac\'{e}n}, \bibinfo{author}{P.~Rosso}, \bibinfo{author}{A.~Giachanou},
\newblock \bibinfo{title}{Prhlt-upv at semeval-2020 task 8: Study of multimodal techniques for memes analysis},
\newblock in: \bibinfo{booktitle}{14th International Workshops on Semantic Evaluation, SemEval 2020 - co-located 28th International Conference on Computational Linguistics, COLING 2020, Proceedings of the Workshop}, \bibinfo{year}{2020}, pp. \bibinfo{pages}{908--915}.
\bibitem[{Giri et~al.(2021)Giri, Gupta, and Gupta}]{giri2021approach}
\bibinfo{author}{R.~Giri}, \bibinfo{author}{S.~Gupta}, \bibinfo{author}{U.~Gupta},
\newblock \bibinfo{title}{An approach to detect offence in memes using natural language processing(nlp) and deep learning},
\newblock in: \bibinfo{booktitle}{2021 International Conference on Computer Communication and Informatics, ICCCI 2021}, \bibinfo{year}{2021}, p. \bibinfo{pages}{9402406}. \DOIprefix\doi{10.1109/iccci50826.2021.9402406}.
\bibitem[{Gupta et~al.(2020)Gupta, Kataria, Mishra, Badal, and Mishra}]{gupta2020bennettnlp}
\bibinfo{author}{A.~Gupta}, \bibinfo{author}{H.~Kataria}, \bibinfo{author}{S.~Mishra}, \bibinfo{author}{T.~Badal}, \bibinfo{author}{V.~Mishra},
\newblock \bibinfo{title}{Bennettnlp at semeval-2020 task 8: Multimodal sentiment classification using hybrid hierarchical classifier},
\newblock in: \bibinfo{booktitle}{Proceedings of SemEval 2020 - 14th International Workshops on Semantic Evaluation}, \bibinfo{year}{2020}, pp. \bibinfo{pages}{1085--1093}.
\bibitem[{Hakimov et~al.(2022)Hakimov, Cheema, and Ewerth}]{hakimov2022tib}
\bibinfo{author}{S.~Hakimov}, \bibinfo{author}{G.~Cheema}, \bibinfo{author}{R.~Ewerth},
\newblock \bibinfo{title}{Tib-va at semeval-2022 task 5: A multimodal architecture for the detection and classification of misogynous memes},
\newblock in: \bibinfo{booktitle}{Proceedings of SemEval 2022 - 16th International Workshop on Semantic Evaluation}, \bibinfo{year}{2022}, pp. \bibinfo{pages}{756--760}.
\bibitem[{Hossain et~al.(2023)Hossain, Hoque, and Hossain}]{hossain2023inter}
\bibinfo{author}{E.~Hossain}, \bibinfo{author}{M.~Hoque}, \bibinfo{author}{M.~Hossain},
\newblock \bibinfo{title}{An inter-modal attention framework for multimodal offense detection},
\newblock in: \bibinfo{booktitle}{Lecture Notes in Networks and Systems}, volume \bibinfo{volume}{569 Lnns}, \bibinfo{year}{2023}, pp. \bibinfo{pages}{853--862}. \DOIprefix\doi{10.1007/978-3-031-19958-5\_81}.
\bibitem[{Myilvahanan et~al.(2023)Myilvahanan, Shashank, Raj, Attanti, and Sahay}]{myilvahanan2023study}
\bibinfo{author}{K.~Myilvahanan}, \bibinfo{author}{B.~Shashank}, \bibinfo{author}{T.~Raj}, \bibinfo{author}{C.~Attanti}, \bibinfo{author}{S.~Sahay},
\newblock \bibinfo{title}{A study on deep learning based classification and identification of offensive memes},
\newblock in: \bibinfo{booktitle}{Proceedings of the 3rd International Conference on Trends in Electronics and Informatics, ICOEI 2019}, \bibinfo{year}{2023}, pp. \bibinfo{pages}{214--218}. \DOIprefix\doi{10.1109/icoei.2019.8862647}.
\bibitem[{Nguyen et~al.(2022)Nguyen, Pham, Nguyen, Nguyen, Nguyen, and Kim}]{nguyen2022hcilab}
\bibinfo{author}{T.~Nguyen}, \bibinfo{author}{N.~Pham}, \bibinfo{author}{N.~Nguyen}, \bibinfo{author}{H.~Nguyen}, \bibinfo{author}{L.~Nguyen}, \bibinfo{author}{Y.-G. Kim},
\newblock \bibinfo{title}{Hcilab at memotion 2.0 2022: Analysis of sentiment, emotion and intensity of emotion classes from meme images using single and multi modalities},
\newblock in: \bibinfo{booktitle}{CEUR Workshop Proceedings}, volume \bibinfo{volume}{3168}, \bibinfo{year}{2022}.
\bibitem[{Phan et~al.(2022)Phan, Lee, Yang, and Kim}]{phan2022little}
\bibinfo{author}{K.~Phan}, \bibinfo{author}{G.-S. Lee}, \bibinfo{author}{H.-J. Yang}, \bibinfo{author}{S.-H. Kim},
\newblock \bibinfo{title}{Little flower at memotion 2.0 2022: Ensemble of multi-modal model using attention mechanism in memotion analysis},
\newblock in: \bibinfo{booktitle}{CEUR Workshop Proceedings}, volume \bibinfo{volume}{3168}, \bibinfo{year}{2022}.
\bibitem[{Shang et~al.(2021)Shang, Zhang, and Zha}]{shang2021knowmeme}
\bibinfo{author}{L.~Shang}, \bibinfo{author}{Y.~Zhang}, \bibinfo{author}{Y.~Zha},
\newblock \bibinfo{title}{Knowmeme: A knowledge-enriched graph neural network solution to offensive meme detection},
\newblock in: \bibinfo{booktitle}{Proceedings - IEEE 17th International Conference on eScience, eScience 2021}, \bibinfo{year}{2021}, pp. \bibinfo{pages}{186--195}. \DOIprefix\doi{10.1109/eScience51609.2021.00029}.
\bibitem[{Walinska and Potoniec(2020)}]{walinska2020urszula}
\bibinfo{author}{U.~Walinska}, \bibinfo{author}{J.~Potoniec},
\newblock \bibinfo{title}{Urszula wali\'{n}ska at semeval-2020 task 8: Fusion of text and image features using lstm and vgg16 for memotion analysis},
\newblock in: \bibinfo{booktitle}{Proceedings of SemEval 2020 -- 14th International Workshops on Semantic Evaluation}, \bibinfo{year}{2020}, pp. \bibinfo{pages}{1215--1220}.
\bibitem[{Yu and Kolossa(2022)}]{yu2022wentaorub}
\bibinfo{author}{W.~Yu}, \bibinfo{author}{D.~Kolossa},
\newblock \bibinfo{title}{wentaorub at memotion 3: Ensemble learning for multi-modal meme classification},
\newblock in: \bibinfo{booktitle}{CEUR Workshop Proceedings}, volume \bibinfo{volume}{3555}, \bibinfo{year}{2022}.
\bibitem[{Zhong et~al.(2022)Zhong, Wang, and Liu}]{zhong2022combining}
\bibinfo{author}{Q.~Zhong}, \bibinfo{author}{Q.~Wang}, \bibinfo{author}{J.~Liu},
\newblock \bibinfo{title}{Combining knowledge and multi-modal fusion for meme classification},
\newblock in: \bibinfo{booktitle}{Lecture Notes in Computer Science}, volume \bibinfo{volume}{13141}, \bibinfo{year}{2022}, pp. \bibinfo{pages}{599--611}. \DOIprefix\doi{10.1007/978-3-030-98358-1\_47}.
\bibitem[{Pramanick et~al.(2021)Pramanick, Akhtar, and Chakraborty}]{pramanick2021exercise}
\bibinfo{author}{S.~Pramanick}, \bibinfo{author}{M.~S. Akhtar}, \bibinfo{author}{T.~Chakraborty},
\newblock \bibinfo{title}{Exercise? i thought you said 'extra fries' ☺: Leveraging sentence demarcations and multi-hop attention for meme affect analysis},
\newblock \bibinfo{journal}{arXiv}  (\bibinfo{year}{2021}).
\bibitem[{Gaurav et~al.(2020)Gaurav, Shandilya, Tiwari, and Goyal}]{gaurav2020machine}
\bibinfo{author}{D.~Gaurav}, \bibinfo{author}{S.~Shandilya}, \bibinfo{author}{S.~Tiwari}, \bibinfo{author}{A.~Goyal},
\newblock \bibinfo{title}{A machine learning method for recognizing invasive content in memes},
\newblock in: \bibinfo{booktitle}{Communications in Computer and Information Science}, \bibinfo{year}{2020}, pp. \bibinfo{pages}{195--213}. \DOIprefix\doi{10.1007/978-3-030-65384-2\_15}.
\bibitem[{Gundapu and Mamidi(2022)}]{Gundapu2022detection}
\bibinfo{author}{S.~Gundapu}, \bibinfo{author}{R.~Mamidi},
\newblock \bibinfo{title}{Detection of propaganda techniques in visuo-lingual metaphor in memes},
\newblock \bibinfo{journal}{arXiv}  (\bibinfo{year}{2022}).
\bibitem[{Abujaber et~al.(2021)Abujaber, Qarqaz, and Abdullah}]{abujaber2021lecun}
\bibinfo{author}{D.~Abujaber}, \bibinfo{author}{A.~Qarqaz}, \bibinfo{author}{M.~Abdullah},
\newblock \bibinfo{title}{Lecun at semeval-2021 task 6: Detecting persuasion techniques in text using ensembled pretrained transformers and data augmentation},
\newblock in: \bibinfo{booktitle}{Proceedings of SemEval 2021 - 15th Intl. Workshop on Semantic Evaluation}, \bibinfo{year}{2021}, pp. \bibinfo{pages}{1068--1074}.
\bibitem[{Alam et~al.(2022)Alam, Mubarak, Zaghouani, Da~San~Martino, and Nakov}]{alam2022overview}
\bibinfo{author}{F.~Alam}, \bibinfo{author}{H.~Mubarak}, \bibinfo{author}{W.~Zaghouani}, \bibinfo{author}{G.~Da~San~Martino}, \bibinfo{author}{P.~Nakov},
\newblock \bibinfo{title}{Overview of the wanlp 2022 shared task on propaganda detection in arabic},
\newblock \bibinfo{journal}{WANLP 2022 - 7th Arabic Natural Language Processing - Proceedings of the Workshop}  (\bibinfo{year}{2022}) \bibinfo{pages}{108--118}.
\bibitem[{Cui et~al.(2023)Cui, Li, Zhang, and Yuan}]{cui2023multimodal}
\bibinfo{author}{J.~Cui}, \bibinfo{author}{L.~Li}, \bibinfo{author}{X.~Zhang}, \bibinfo{author}{J.~Yuan},
\newblock \bibinfo{title}{Multimodal propaganda detection via anti-persuasion prompt enhanced contrastive learning},
\newblock in: \bibinfo{booktitle}{ICASSP, IEEE International Conference on Acoustics, Speech and Signal Processing - Proceedings}, \bibinfo{year}{2023}, p.~\bibinfo{pages}{0}. \DOIprefix\doi{10.1109/icassp49357.2023.10096771}.
\bibitem[{Hossain et~al.(2021)Hossain, Naim, Tasneem, Tasnia, and Chy}]{hossain2021csecudsg}
\bibinfo{author}{T.~Hossain}, \bibinfo{author}{J.~Naim}, \bibinfo{author}{F.~Tasneem}, \bibinfo{author}{R.~Tasnia}, \bibinfo{author}{A.~Chy},
\newblock \bibinfo{title}{Csecu-dsg at semeval-2021 task 6: Orchestrating multimodal neural architectures for identifying persuasion techniques in texts and images},
\newblock in: \bibinfo{booktitle}{Proceedings of SemEval 2021 - 15th International Workshop on Semantic Evaluation}, \bibinfo{year}{2021}, pp. \bibinfo{pages}{1088--1095}.
\bibitem[{Li et~al.(2021)Li, Li, and Sun}]{li20211213li}
\bibinfo{author}{P.~Li}, \bibinfo{author}{X.~Li}, \bibinfo{author}{X.~Sun},
\newblock \bibinfo{title}{1213li at semeval-2021 task 6: Detection of propaganda with multi-modal attention and pre-trained models},
\newblock in: \bibinfo{booktitle}{SemEval 2021 - 15th International Workshop on Semantic Evaluation, Proceedings of the Workshop}, \bibinfo{year}{2021}, pp. \bibinfo{pages}{1032--1036}.
\bibitem[{Liu et~al.(2022)Liu, Geigle, Krebs, and Gurevych}]{liu2022figmemes}
\bibinfo{author}{C.~Liu}, \bibinfo{author}{G.~Geigle}, \bibinfo{author}{R.~Krebs}, \bibinfo{author}{I.~Gurevych},
\newblock \bibinfo{title}{Figmemes: A dataset for figurative language identification in politically-opinionated memes},
\newblock in: \bibinfo{booktitle}{Proceedings of the 2022 Conference on Empirical Methods in Natural Language Processing}, \bibinfo{year}{2022}, pp. \bibinfo{pages}{7069--7086}.
\bibitem[{Wu et~al.(2022)Wu, Li, Li, and Wang}]{wu2022propaganda}
\bibinfo{author}{H.~Wu}, \bibinfo{author}{X.~Li}, \bibinfo{author}{L.~Li}, \bibinfo{author}{Q.~Wang},
\newblock \bibinfo{title}{Propaganda techniques detection in low-resource memes with multi-modal prompt tuning},
\newblock in: \bibinfo{booktitle}{Proceedings - IEEE International Conference on Multimedia and Expo}, \bibinfo{year}{2022}, p.~\bibinfo{pages}{0}. \DOIprefix\doi{10.1109/icme52920.2022.9859642}.
\bibitem[{Zhu et~al.(2021)Zhu, Wang, and Zhang}]{zhu2021ynuhpcc}
\bibinfo{author}{X.~Zhu}, \bibinfo{author}{J.~Wang}, \bibinfo{author}{X.~Zhang},
\newblock \bibinfo{title}{Ynu-hpcc at semeval-2021 task 6: Combining albert and text-cnn for persuasion detection in texts and images},
\newblock in: \bibinfo{booktitle}{SemEval 2021 - 15th International Workshop on Semantic Evaluation, Proceedings of the Workshop}, \bibinfo{year}{2021}, pp. \bibinfo{pages}{1045--1050}.
\bibitem[{Chen et~al.(2023)Chen, Zhao, Piao, Ding, and Cui}]{chen2023multimodal}
\bibinfo{author}{P.~Chen}, \bibinfo{author}{L.~Zhao}, \bibinfo{author}{Y.~Piao}, \bibinfo{author}{H.~Ding}, \bibinfo{author}{X.~Cui},
\newblock \bibinfo{title}{Multimodal visual-textual object graph attention network for propaganda detection in memes},
\newblock \bibinfo{journal}{Multimedia Tools and Applications} \bibinfo{volume}{1} (\bibinfo{year}{2023}) \bibinfo{pages}{1--10}.
\bibitem[{Cui et~al.(2023)Cui, Li, and Tao}]{cui2023beornot}
\bibinfo{author}{J.~Cui}, \bibinfo{author}{L.~Li}, \bibinfo{author}{X.~Tao},
\newblock \bibinfo{title}{Be-or-not prompt enhanced hard negatives generating for memes category detection},
\newblock in: \bibinfo{booktitle}{Proceedings - IEEE International Conference on Multimedia and Expo}, \bibinfo{year}{2023}.
\bibitem[{Das et~al.(2022)Das, Banerjee, and Mukherjee}]{das2022hate}
\bibinfo{author}{M.~Das}, \bibinfo{author}{S.~Banerjee}, \bibinfo{author}{A.~Mukherjee},
\newblock \bibinfo{title}{hate-alert\@dravidianlangtech-acl2022: Ensembling multi-modalities for tamil trollmeme classification},
\newblock \bibinfo{journal}{arXiv}  (\bibinfo{year}{2022}).
\bibitem[{Hegde et~al.(2021)Hegde, Hande, Priyadharshini, Bharathi, and Chakravarthi}]{Hedge2021do}
\bibinfo{author}{S.~U. Hegde}, \bibinfo{author}{A.~Hande}, \bibinfo{author}{R.~Priyadharshini}, \bibinfo{author}{B.~Bharathi}, \bibinfo{author}{B.~R. Chakravarthi},
\newblock \bibinfo{title}{Do images really do the talking? analysing the significance of images in tamil troll meme classification},
\newblock \bibinfo{journal}{arXiv}  (\bibinfo{year}{2021}).
\bibitem[{Nandi et~al.(2022)Nandi, Alam, and Nakov}]{nandi2022team}
\bibinfo{author}{R.~N. Nandi}, \bibinfo{author}{F.~Alam}, \bibinfo{author}{P.~Nakov},
\newblock \bibinfo{title}{Teamx\@dravidianlangtech-acl2022: A comparative analysis for troll-based meme classification},
\newblock \bibinfo{journal}{arXiv}  (\bibinfo{year}{2022}).
\bibitem[{Zhang et~al.(2022)Zhang, Xu, Wang, Zhou, Zhao, and Teng}]{zhang2022mintrec}
\bibinfo{author}{H.~Zhang}, \bibinfo{author}{H.~Xu}, \bibinfo{author}{X.~Wang}, \bibinfo{author}{Q.~Zhou}, \bibinfo{author}{S.~Zhao}, \bibinfo{author}{J.~Teng},
\newblock \bibinfo{title}{Mintrec: A new dataset for multimodal intent recognition},
\newblock in: \bibinfo{booktitle}{Proceedings of the 30th ACM International Conference on Multimedia}, \bibinfo{year}{2022}, pp. \bibinfo{pages}{1688--1697}.
\bibitem[{Mirzoeff(2023)}]{mirzoeff2023white}
\bibinfo{author}{N.~Mirzoeff}, \bibinfo{title}{White Sight: Visual Politics and Practices of Whiteness}, \bibinfo{publisher}{MIT Press}, \bibinfo{year}{2023}.
\bibitem[{Sekimoto and Brown(2023)}]{sekimoto2023race}
\bibinfo{author}{S.~Sekimoto}, \bibinfo{author}{C.~Brown}, \bibinfo{title}{Race and multimodality: An introduction to the special issue}, \bibinfo{year}{2023}.
\bibitem[{Yus(2019)}]{yus2019multimodality}
\bibinfo{author}{F.~Yus},
\newblock \bibinfo{title}{Multimodality in memes: A cyberpragmatic approach},
\newblock \bibinfo{journal}{Analyzing digital disc.: New insights and future dir.}  (\bibinfo{year}{2019}) \bibinfo{pages}{105--131}.
\bibitem[{Lemke(1998)}]{lemke1998metamedia}
\bibinfo{author}{J.~L. Lemke},
\newblock \bibinfo{title}{Metamedia literacy: Transforming meanings and media},
\newblock \bibinfo{journal}{Handbook of literacy and technology: Transformations in a post-typographic world} \bibinfo{volume}{283301} (\bibinfo{year}{1998}).
\bibitem[{Radford et~al.(2021)Radford, Kim, Hallacy, Ramesh, Goh, Agarwal, Sastry, Askell, Mishkin, Clark, Krueger, and Sutskever}]{radford2021learning}
\bibinfo{author}{A.~Radford}, \bibinfo{author}{J.~W. Kim}, \bibinfo{author}{C.~Hallacy}, \bibinfo{author}{A.~Ramesh}, \bibinfo{author}{G.~Goh}, \bibinfo{author}{S.~Agarwal}, \bibinfo{author}{G.~Sastry}, \bibinfo{author}{A.~Askell}, \bibinfo{author}{P.~Mishkin}, \bibinfo{author}{J.~Clark}, \bibinfo{author}{G.~Krueger}, \bibinfo{author}{I.~Sutskever},
\newblock \bibinfo{title}{Learning transferable visual models from natural language supervision},
\newblock in: \bibinfo{booktitle}{International conference on machine learning}, \bibinfo{organization}{PMLR}, \bibinfo{year}{2021}, pp. \bibinfo{pages}{8748--8763}.
\bibitem[{Wu et~al.(2024)Wu, Gao, Pan, Li, Ma, Liu, and Liu}]{wu2024fuser}
\bibinfo{author}{F.~Wu}, \bibinfo{author}{B.~Gao}, \bibinfo{author}{X.~Pan}, \bibinfo{author}{L.~Li}, \bibinfo{author}{Y.~Ma}, \bibinfo{author}{S.~Liu}, \bibinfo{author}{Z.~Liu},
\newblock \bibinfo{title}{Fuser: An enhanced multimodal fusion framework with congruent reinforced perceptron for hateful memes detection},
\newblock \bibinfo{journal}{Information Processing \& Management} \bibinfo{volume}{61} (\bibinfo{year}{2024}) \bibinfo{pages}{103772}.
\bibitem[{Wu et~al.(2012)Wu, Li, Wang, and Zhu}]{wu2012probase}
\bibinfo{author}{W.~Wu}, \bibinfo{author}{H.~Li}, \bibinfo{author}{H.~Wang}, \bibinfo{author}{K.~Q. Zhu},
\newblock \bibinfo{title}{Probase: A probabilistic taxonomy for text understanding},
\newblock in: \bibinfo{booktitle}{Proceedings of the 2012 ACM SIGMOD international conference on management of data}, \bibinfo{year}{2012}, pp. \bibinfo{pages}{481--492}.
\bibitem[{Tommasini et~al.(2023)Tommasini, Ilievski, and Wijesiriwardene}]{tommasini2023imkg}
\bibinfo{author}{R.~Tommasini}, \bibinfo{author}{F.~Ilievski}, \bibinfo{author}{T.~Wijesiriwardene},
\newblock \bibinfo{title}{Imkg: The internet meme knowledge graph},
\newblock in: \bibinfo{booktitle}{European Semantic Web Conference}, \bibinfo{organization}{Springer}, \bibinfo{year}{2023}, pp. \bibinfo{pages}{354--371}.
\bibitem[{Vrande{\v{c}}i{\'c} and Kr{\"o}tzsch(2014)}]{vrandevcic2014wikidata}
\bibinfo{author}{D.~Vrande{\v{c}}i{\'c}}, \bibinfo{author}{M.~Kr{\"o}tzsch},
\newblock \bibinfo{title}{Wikidata: a free collaborative knowledgebase},
\newblock \bibinfo{journal}{Communications of the ACM} \bibinfo{volume}{57} (\bibinfo{year}{2014}) \bibinfo{pages}{78--85}.
\bibitem[{Auer et~al.(2007)Auer, Bizer, Kobilarov, Lehmann, Cyganiak, and Ives}]{auer2007dbpedia}
\bibinfo{author}{S.~Auer}, \bibinfo{author}{C.~Bizer}, \bibinfo{author}{G.~Kobilarov}, \bibinfo{author}{J.~Lehmann}, \bibinfo{author}{R.~Cyganiak}, \bibinfo{author}{Z.~Ives},
\newblock \bibinfo{title}{Dbpedia: A nucleus for a web of open data},
\newblock in: \bibinfo{booktitle}{international semantic web conference}, \bibinfo{organization}{Springer}, \bibinfo{year}{2007}, pp. \bibinfo{pages}{722--735}.
\bibitem[{Bollacker et~al.(2008)Bollacker, Evans, Paritosh, Sturge, and Taylor}]{bollacker2008freebase}
\bibinfo{author}{K.~Bollacker}, \bibinfo{author}{C.~Evans}, \bibinfo{author}{P.~Paritosh}, \bibinfo{author}{T.~Sturge}, \bibinfo{author}{J.~Taylor},
\newblock \bibinfo{title}{Freebase: a collaboratively created graph database for structuring human knowledge},
\newblock in: \bibinfo{booktitle}{Proceedings of the 2008 ACM SIGMOD international conference on Management of data}, \bibinfo{year}{2008}, pp. \bibinfo{pages}{1247--1250}.
\bibitem[{Bates et~al.(2023)Bates, Christensen, Nakov, and Gurevych}]{bates2023template}
\bibinfo{author}{L.~Bates}, \bibinfo{author}{P.~E. Christensen}, \bibinfo{author}{P.~Nakov}, \bibinfo{author}{I.~Gurevych},
\newblock \bibinfo{title}{A template is all you meme},
\newblock \bibinfo{journal}{arXiv preprint arXiv:2311.06649}  (\bibinfo{year}{2023}).
\bibitem[{Scheuerman et~al.(2021)Scheuerman, Jiang, Fiesler, and Brubaker}]{scheuerman2021framework}
\bibinfo{author}{M.~K. Scheuerman}, \bibinfo{author}{J.~A. Jiang}, \bibinfo{author}{C.~Fiesler}, \bibinfo{author}{J.~R. Brubaker},
\newblock \bibinfo{title}{A framework of severity for harmful content online},
\newblock \bibinfo{journal}{Proceedings of the ACM on Human-Computer Interaction} \bibinfo{volume}{5} (\bibinfo{year}{2021}) \bibinfo{pages}{1--33}.
\bibitem[{Jahanbakhsh et~al.(2021)Jahanbakhsh, Zhang, Berinsky, Pennycook, Rand, and Karger}]{jahanbakhsh2021exploring}
\bibinfo{author}{F.~Jahanbakhsh}, \bibinfo{author}{A.~X. Zhang}, \bibinfo{author}{A.~J. Berinsky}, \bibinfo{author}{G.~Pennycook}, \bibinfo{author}{D.~G. Rand}, \bibinfo{author}{D.~R. Karger},
\newblock \bibinfo{title}{Exploring lightweight interventions at posting time to reduce the sharing of misinformation on social media},
\newblock \bibinfo{journal}{Proceedings of the ACM on Human-Computer Interaction} \bibinfo{volume}{5} (\bibinfo{year}{2021}) \bibinfo{pages}{1--42}.
\bibitem[{Kumarage et~al.(2024)Kumarage, Bhattacharjee, and Garland}]{kumarage2024harnessing}
\bibinfo{author}{T.~Kumarage}, \bibinfo{author}{A.~Bhattacharjee}, \bibinfo{author}{J.~Garland},
\newblock \bibinfo{title}{Harnessing artificial intelligence to combat online hate: Exploring the challenges and opportunities of large language models in hate speech detection},
\newblock \bibinfo{journal}{arXiv preprint arXiv:2403.08035}  (\bibinfo{year}{2024}).
\bibitem[{Beskow et~al.(2020)Beskow, Kumar, and Carley}]{beskow2020evolution}
\bibinfo{author}{D.~M. Beskow}, \bibinfo{author}{S.~Kumar}, \bibinfo{author}{K.~M. Carley},
\newblock \bibinfo{title}{The evolution of political memes: Detecting and characterizing internet memes with multi-modal deep learning},
\newblock \bibinfo{journal}{Information Processing \& Management} \bibinfo{volume}{57} (\bibinfo{year}{2020}) \bibinfo{pages}{102170}.

\end{thebibliography}
